\documentclass{ieeetj}
\usepackage{cite}
\usepackage{amsmath,amssymb,amsfonts}
\usepackage{algorithmic}
\usepackage{graphicx,color}
\usepackage{textcomp}
\usepackage{xcolor}
\usepackage{tabularx} %
\usepackage{hyperref}
\hypersetup{hidelinks=true}
\usepackage{algorithm,algorithmic}
\def\BibTeX{{\rm B\kern-.05em{\sc i\kern-.025em b}\kern-.08em
    T\kern-.1667em\lower.7ex\hbox{E}\kern-.125emX}}
\AtBeginDocument{\definecolor{tmlcncolor}{cmyk}{0.93,0.59,0.15,0.02}\definecolor{NavyBlue}{RGB}{0,86,125}}

\def\OJlogo{\vspace{-4pt}$<$Society logo(s) and publication title will appear here.$>$}
\def\seclogo{\vspace{10pt}$<$Society logo(s) and publication title will appear here.$>$}

\def\authorrefmark#1{\ensuremath{^{\textbf{#1}}}}

\DeclareMathOperator*{\argmin}{arg\,min}

\begin{document}
\receiveddate{XX Month, XXXX}
\reviseddate{XX Month, XXXX}
\accepteddate{XX Month, XXXX}
\publisheddate{XX Month, XXXX}
\currentdate{XX Month, XXXX}
\doiinfo{XXXX.2022.1234567}

\markboth{Open-Set 3D Scene Graphs for Field Robotics: An Outdoor Case Study}{Samuelson {et al.}}

\title{Open-Set 3D Scene Graphs for Field Robotics: An Outdoor Case Study}

\author{Chad R. Samuelson~\authorrefmark{1}, Gabriel R. Slade~\authorrefmark{1}, and Joshua G. Mangelson~\authorrefmark{1}}
\affil{Brigham Young University}
\corresp{Corresponding author: Chad Samuelson (email: chadrs2@byu.edu).}
\authornote{This work was partially funded under Office of Naval Research award numbers N00014-24-1-2301 and N00014-24-1-2503, as well as by the Center for Autonomous Air Mobility and Sensing (CAAMS), a National Science Foundation Industry-University Cooperative Research Center (IUCRC) under NSF award number 2139551, along with significant contributions from CAAMS industry members.}

\begin{abstract}
Three-dimensional scene graphs (3DSGs) have emerged as a promising approach for building geometrically grounded, semantically informed, hierarchical general-purpose maps to support high-level robotic reasoning.
However, the behavior of 3DSGs in real-world outdoor deployments remains poorly understood, particularly when combined with open-set vision-language models (VLMs). 
In this field report, we analyze the components common to most 3DSG representations across five outdoor robotic datasets to characterize challenges that arise in complex outdoor environments.
Using the recently proposed Terra 3DSG as a case study, we investigate semantic point embeddings, place-node graph navigation, region-level understanding, and memory size across the five diverse datasets. 
We additionally introduce novel consistency metrics to evaluate whether semantic and structural graph properties remain stable across repeated traversals of the same environment. 
Our analysis reveals that outliers and multiple modes are common in VLM point embeddings across all tested datasets with outlier ratios above $0.1$ for around $30\%$ of points.
We demonstrate the feasibility of outdoor 3DSGs for navigation-based object retrieval, achieving success rates near $70\%$, though performance is limited by traversability failures and inefficient routing, with trajectories averaging approximately $66\%$ suboptimal path efficiency. 
Region-level understanding remains challenging in complex natural environments, with low average F1 scores around $0.359$. 
Overall, our results show that outdoor 3DSGs can maintain compact (less than $600$MB for multi-kilometer trajectories) and relatively consistent large-scale environment representations, while highlighting open challenges in handling multiple semantic modes, incorporating traversability into graph structures, and improving higher-level region understanding.
\end{abstract}


\IEEEspecialpapernotice{(Invited Paper)}

\maketitle

\section{INTRODUCTION}
\label{sec:intro}

\IEEEPARstart{A}{utonomous} robots operating in large-scale outdoor environments require maps that support not only geometric navigation but also high-level semantic reasoning and scene understanding. 
3D scene graphs (3DSGs) have recently emerged as a promising representation for integrating geometry, semantics, and hierarchical structure within a single map, enabling task-level reasoning and decision making.
These 3DSG approaches have recently incorporated open-set vision-language models (VLMs) like CLIP~\cite{clip_radfordLearningTransferable2021} to encode semantics into the map.
While these representations have shown strong results mainly in the indoor setting, 3DSG's behavior in real-world complex outdoor settings remains predominantly unexplored. 

A growing body of work has begun to extend indoor 3DSG techniques to outdoor environments \cite{greve2024curb, steinkeCollaborativeDynamic2025b, zhangParkingSGOpenVocabulary2025, rayTaskMotion2024, straderIndoorOutdoor2024, shanGraph2Nav3D2025, ongATLASNavigator2025, samuelsonTerra2025}. 
These efforts demonstrate the promise of scene graph representations beyond indoor settings, but their evaluation is often limited in scope, focusing on isolated tasks, simpler outdoor environments, or limited performance metrics such as object detection accuracy, memory usage, or successful object retrieval. 
Outdoor environments further complicate evaluation due to increased scale, unstructured geometry, and ambiguous distinctions between terrain, objects, and regions. 
This is particularly true outside of highly structured urban settings. 
These factors make systematic field evaluation difficult, but also essential for understanding the practical utility of outdoor 3DSGs.

The goal of this paper is to study the behavior and failure modes of open-set, VLM-based 3DSG systems under realistic outdoor robotic operating conditions.
We conduct a field study across five outdoor robotic datasets collected in complex and varied environments. 
To ground our analysis, we include a case study of a representative terrain-aware outdoor 3DSG system, which combines LiDAR-based mapping with CLIP-based semantic features \cite{samuelsonTerra2025}, focusing on its behavior under real-world conditions.
We additionally analyze failure modes of VLM-based semantic grounding, advantages and disadvantages in navigation using a 3DSG, and introduce metrics to quantify structural and semantic consistency in 3DSGs across environments and repeated traversals.
Our focus is on understanding when and why VLM-based 3DSG systems succeed or fail in realistic outdoor deployment settings, and what this implies for their use in robotic mapping and reasoning.
The key contributions of this field report are:
\begin{enumerate}
    \item An in-depth analysis of VLM point embeddings in outdoor scenes which, by both quantitative and qualitative evidence, uncovers the presence of semantic outlier embeddings and multiple embedding modes,
    \item Real-world navigation experiments across large outdoor scenes using only a 3DSG graph with an A* planner, demonstrating both the utility and limitations of the evaluated open-set 3DSG formulation, 
    \item Quantitative analysis of region-level semantic reasoning across greatly varied scenes with an analysis exposing needs for better region semantic reasoning,
    \item An empirical study of the scalability and memory footprint of the evaluated 3DSG representation,
    \item Novel consistency metrics defined and used in a multiple-run 3DSG consistency evaluation in outdoor deployment, and 
    \item Identification and discussion of key success factors and failure modes relevant to 3DSG deployment for real-world field robotics. 
\end{enumerate}

The remainder of this paper is organized as follows. 
Section~\ref{sec:related_work} reviews the related work of outdoor 3DSGs and related field studies. 
Section~\ref{sec:terra_overview} provides a brief overview of the Terra method \cite{samuelsonTerra2025} used in some experiments.
Section~\ref{sec:datasets_robot_platform} gives an in depth discussion of the five datasets collected and the robot platform and sensors used for dataset collection.
The analysis sections progress hierarchically through the 3DSG representation, beginning with low-level VLM point embeddings in Section~\ref{sec:point_embs_analysis}, followed by navigation over the place-node graph in Section~\ref{sec:place_node_nav_analysis}.
Section~\ref{sec:region_analysis} presents a quantitative analysis of 3DSG capabilities in region-level semantic understanding using VLM reasoning.
Section~\ref{sec:full_3dsg_analysis} focuses on the broader 3DSG utility in field robotics with a quantitative analysis of graph and memory size across the datasets as well as in-depth consistency experiments.
Section~\ref{sec:lessons_learned} provides a brief discussion of the takeaways and lessons learned of VLM-based 3DSG methods in outdoor environments.
Finally, Section~\ref{sec:conclusion} concludes and summarizes the findings in this field report.

\section{RELATED WORK} 
\label{sec:related_work}

\subsection{SEMANTIC MAPPING AND 3DSG REPRESENTATIONS}
\label{subsec:sem_mapping_sect}

Scene graphs were originally introduced as a structured representation for modeling objects and their relationships within 2D images \cite{krishna2DSG2017, scenegraph2dsurvey_li}. 
Building on this foundation, Armeni et al.~\cite{armeni3DSemantic2016, armeni3DScene2019b} extended scene graph representations to three-dimensional indoor environments by integrating semantic labels with reconstructed 3D geometry, resulting in hierarchical 3DSGs that organize objects, rooms, and floors.

Since these developments, 3DSGs have continued to evolve, with a common emphasis on organized map representations that jointly encode geometry, semantics, hierarchy, and other physical or semantic relationships \cite{rosinol3DDynamic2020, hughes2022hydra, bavle2022sgraphs+, werby23hovsg, guConceptGraphsOpenVocabulary2024, kassab2024barenecessities, maggio2024Clio, bavle2025sgraphs20hierarchicalsemantic}. 
These representations aim to bridge low-level perception with high-level reasoning by providing a unified map structure suitable for long-term autonomy and decision making.

A key motivation for 3DSGs is their ability to support task-level reasoning.
Recent approaches have incorporated open-set vision-language models (VLMs) to enable flexible semantic queries for object retrieval and scene understanding.
CLIP~\cite{clip_radfordLearningTransferable2021} is a commonly used VLM in open-set semantic systems due to its lightweight nature, and ability to extract semantic embeddings prior to any specified textual or image prompt.
In parallel, many systems include layers for places or navigation that capture traversable free space, allowing robots to plan and execute navigation tasks within the same semantic map \cite{werby23hovsg, guConceptGraphsOpenVocabulary2024, kassab2024barenecessities, maggio2024Clio, bavle2025sgraphs20hierarchicalsemantic}.
By organizing information hierarchically and encoding relationships between entities, 3DSGs further enable context-aware reasoning for more complex robotic tasks \cite{werby23hovsg, maggio2024Clio, bavle2025sgraphs20hierarchicalsemantic}.

Evaluation of indoor 3DSG systems has primarily focused on object and room detection accuracy, memory footprint, and task-level performance such as navigation success. While these metrics demonstrate the effectiveness of 3DSGs in structured indoor environments, they provide limited insight into system behavior under repeated deployment, long-term operation, or significant environmental variation.

\subsection{OUTDOOR SEMANTIC MAPPING AND 3DSGS}

While 3DSGs have demonstrated strong performance in structured indoor environments, extending semantic mapping and scene graph representations to outdoor settings introduces fundamentally new challenges. 
Outdoor environments are typically larger in scale, less explicitly structured, and more visually and geometrically diverse, with ambiguous boundaries between objects, terrain, and regions.
These characteristics complicate perception, semantic abstraction, and long-term map consistency, motivating the development of outdoor-specific semantic mapping and 3DSG approaches.

Due to the wide range of outdoor scenes, a subclass of outdoor 3DSGs are focused on urban-specific 3DSGs. 
These methods focus primarily on mapping and tracking dynamic and static objects along with navigable street nodes \cite{greve2024curb, steinkeCollaborativeDynamic2025b, zhangParkingSGOpenVocabulary2025}. 
While thoroughly evaluated and tested within their specific areas of interest, they are primarily targeted toward urban and autonomous driving applications.

Other approaches have sought to adapt indoor 3DSG techniques, which demonstrate broad capability across diverse environments, to more general outdoor settings.
Given the complexity of outdoor perception and navigation, several methods focus on specific aspects of this challenge. 
Ray et al.~\cite{rayTaskMotion2024} focus on task definitions and navigation constraints, while Strader et al.~\cite{straderIndoorOutdoor2024} address region definition in semantically ambiguous outdoor scenes.
Strader et al.~\cite{strader2025language} further demonstrate language-grounded hierarchical planning and execution for outdoor autonomous navigation using multi-robot 3DSGs, supporting object- and region-centric navigation based on natural-language queries. 
Their evaluation demonstrates end-to-end task execution in a single large-scale outdoor operational environment, whereas our field study focuses on characterizing representation behavior and failure modes across five environments with substantially different terrain, geometry, and perceptual conditions.
Shan et al.~\cite{shanGraph2Nav3D2025} propose a more general outdoor 3DSG framework that emphasizes object relationships and region definitions derived from urban road boundaries. 

Ong et al.~\cite{ongATLASNavigator2025} and Samuelson et al.~\cite{samuelsonTerra2025} represent two recent directions in VLM-based outdoor 3DSG-like systems: ATLAS builds hierarchical maps using semantic Gaussian splatting, while Terra constructs a hierarchical 3DSG using a voxel-based representation with CLIP-based semantics for lightweight and scalable deployment. 

While both systems demonstrate impressive task-level capabilities, our field report directly addresses two critical gaps left open by these works. 
First, while ATLAS and Terra evaluate high-level system performance (e.g., navigation success or memory footprint), they treat the underlying open-set semantic space as a black box. 
In contrast, our first contribution explicitly decompresses this space to expose and analyze the multi-modal and outlier characteristics of VLM point embeddings in the wild. 
Second, while both methods validate their representations on isolated single-session trajectories, neither evaluates representation stability over time. 
We bridge this gap by using Terra as a representative case study to rigorously evaluate multi-session map stability, introducing novel consistency metrics to quantify how both the geometry and semantics of these 3DSG structures degrade or hold up across repeated deployments.

\subsection{EVALUATION PRACTICES FOR OUTDOOR SEMANTIC MAPPING SYSTEMS}

Outdoor semantic mapping and 3DSG systems are commonly evaluated using metrics such as semantic object detection accuracy, correctness of object relationships, successful object retrieval or navigation performance, memory footprint, and region identification accuracy, typically across one or two outdoor datasets \cite{rayTaskMotion2024, straderIndoorOutdoor2024, shanGraph2Nav3D2025, ongATLASNavigator2025, steinkeCollaborativeDynamic2025b, zhangParkingSGOpenVocabulary2025, samuelsonTerra2025}.
While these evaluations demonstrate feasibility and task-level capability, they provide limited insight into system robustness, scalability, or semantic consistency under repeated deployment and environmental variation.

This evaluation adopts metrics commonly used in prior outdoor 3DSG studies while extending them to a broader field setting, evaluating performance across five diverse natural outdoor environments. 
This evaluation additionally analyzes low-level semantic details in the embeddings themselves detecting outliers and multiple semantic modes.
We also define a new set of consistency metrics that examine the stability of the 3DSG structure across multiple runs in the same scene, an aspect that has received limited attention in existing indoor or outdoor 3DSG evaluations.

\section{TERRA 3DSG OVERVIEW}
\label{sec:terra_overview}

\begin{figure*}[t]
\centering
\begin{tabular}{cc}
\includegraphics[width=0.45\textwidth]{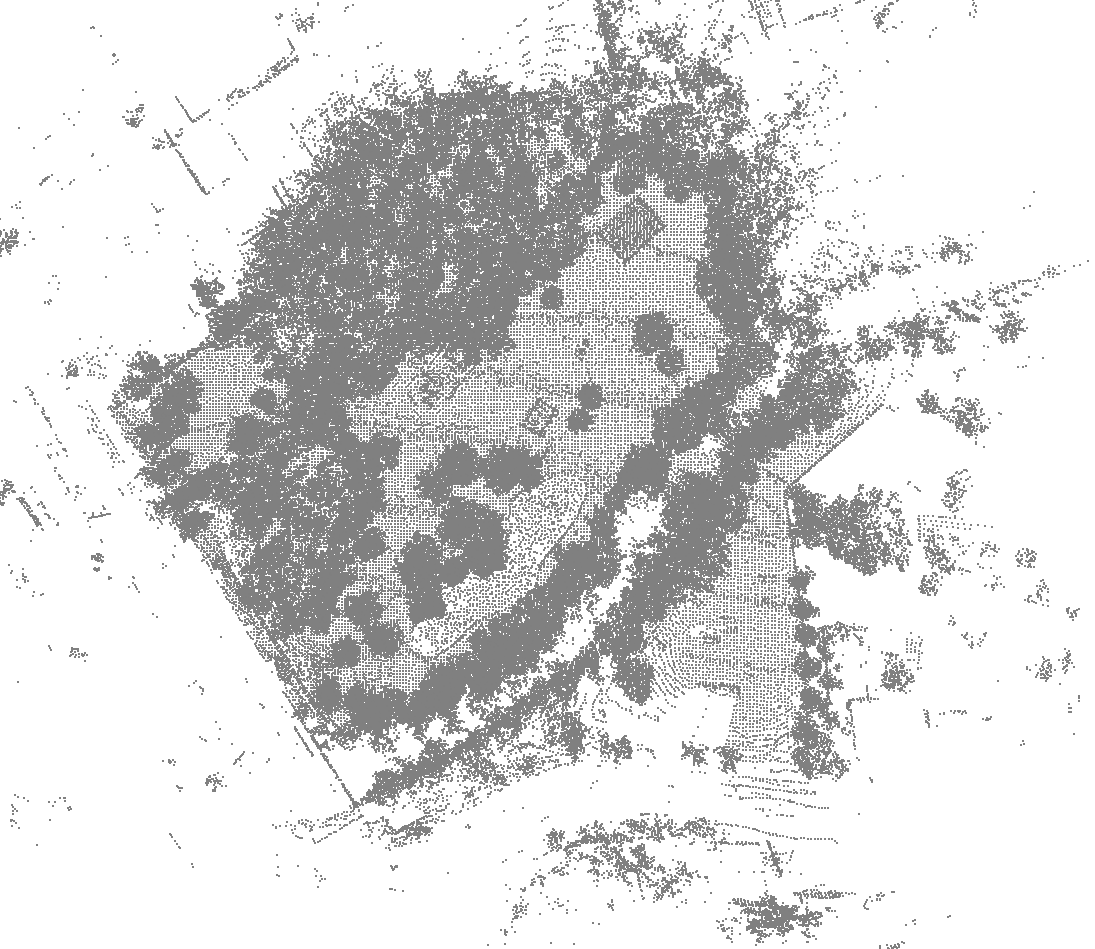} &
\includegraphics[width=0.45\textwidth]{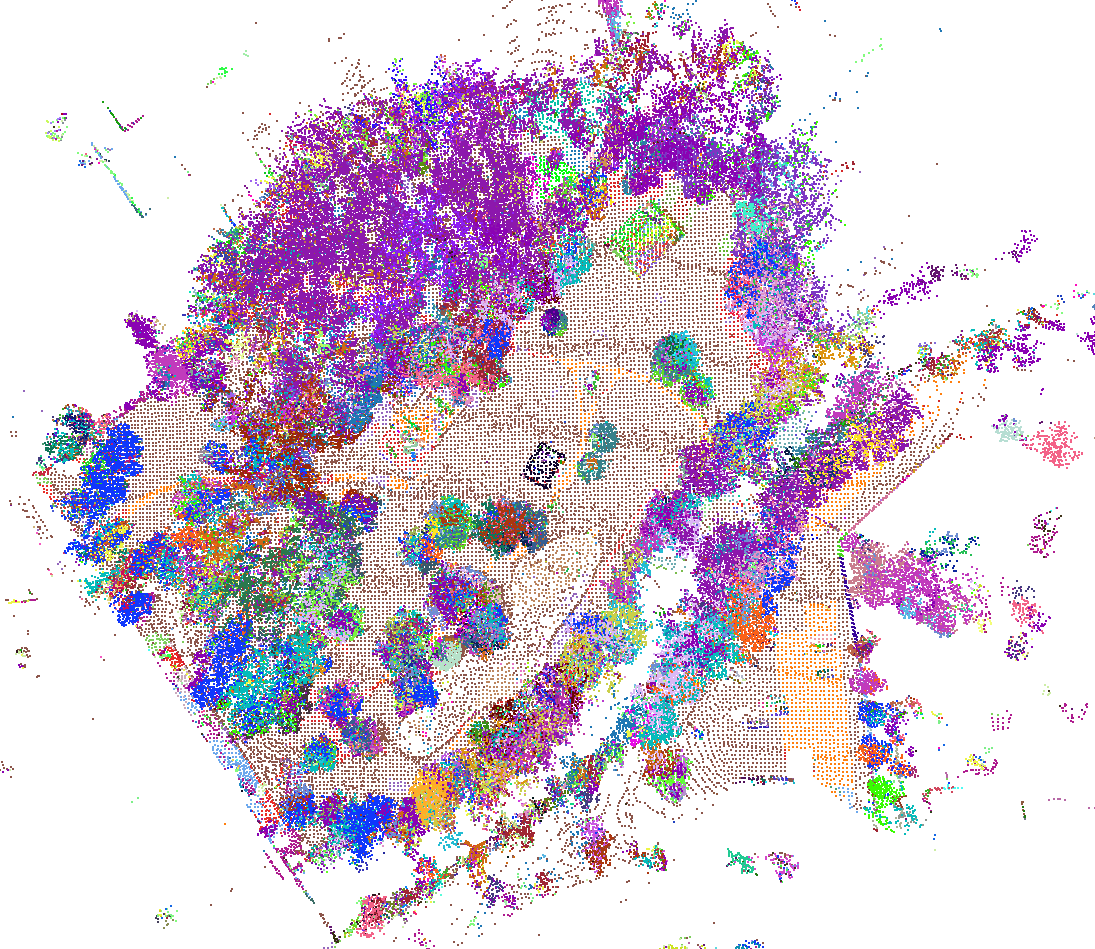} \\

{\fontsize{8}{9.5}\selectfont\rmfamily(a) LIO-SAM Point Cloud} &
{\fontsize{8}{9.5}\selectfont\rmfamily(b) Metric-Semantic Map colored by semantic embeddings} \\

\includegraphics[width=0.45\textwidth]{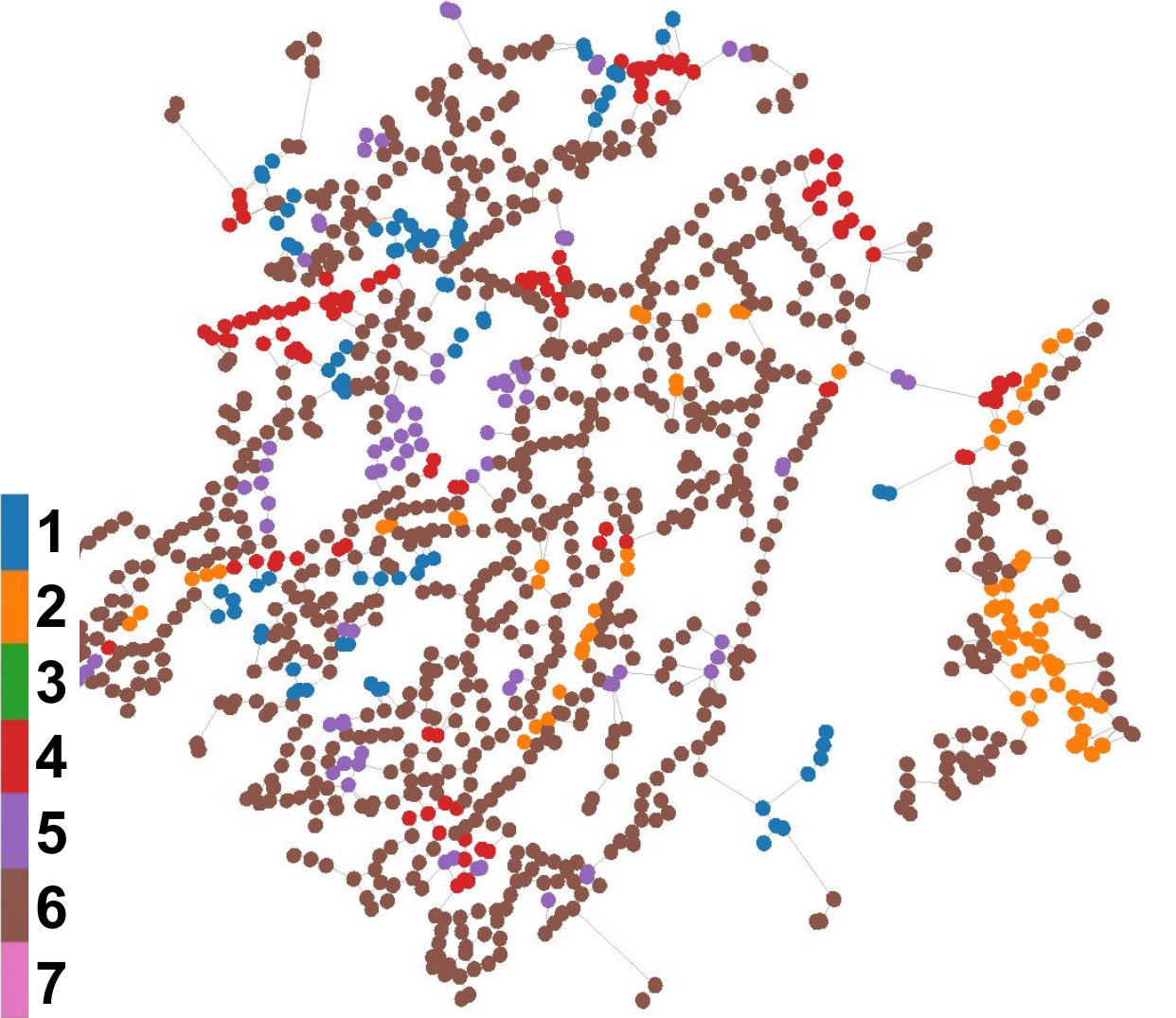} &
\includegraphics[width=0.45\textwidth]{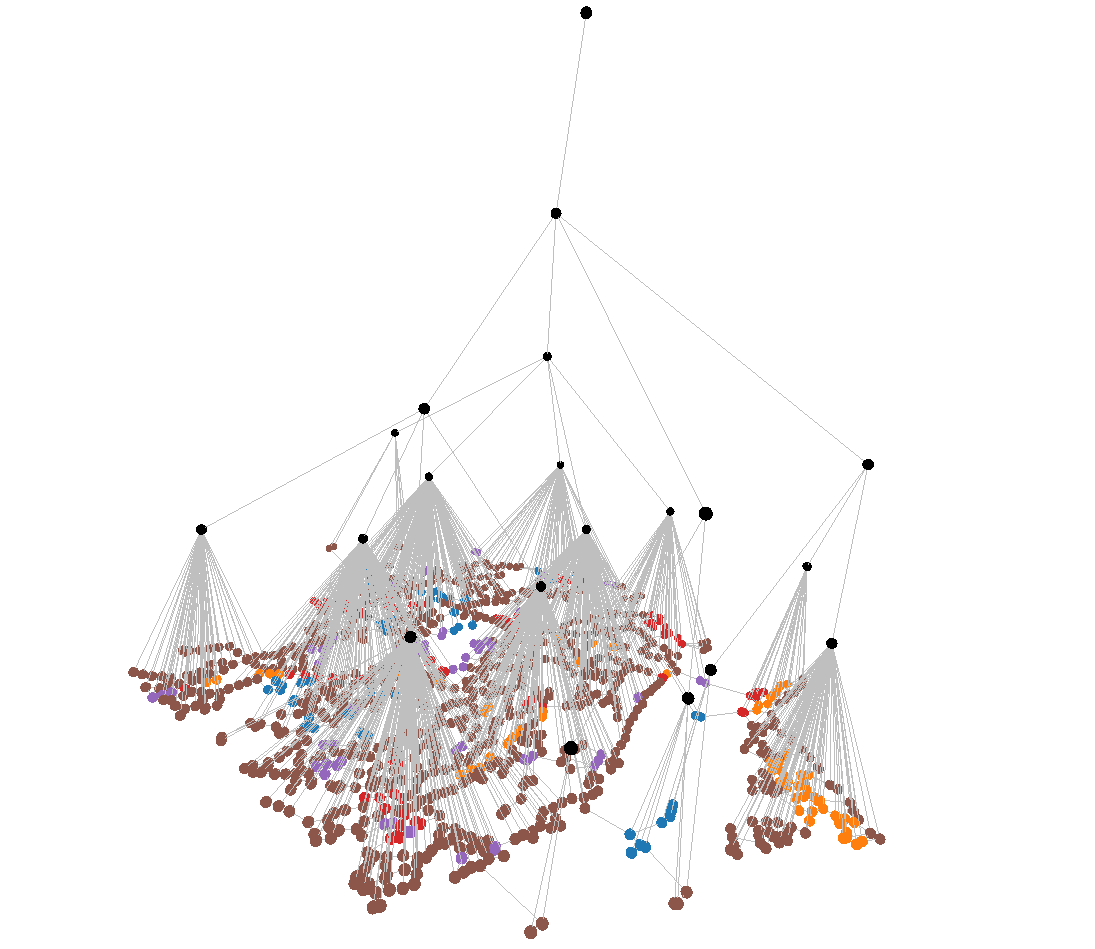} \\

{\fontsize{8}{9.5}\selectfont\rmfamily(c) Place node graph colored by terrain type} &
{\fontsize{8}{9.5}\selectfont\rmfamily(c) Hierarchical region and place node layers} \\

\end{tabular}
\caption{
Example qualitative Terra results showing the sequential phases of Terra in building the full 3DSG on the River Park dataset. (c) Place nodes are colored by terrain type with the following terrains: 1) leaves, 2) road, 3) sidewalk, 4) gravel, 5) dirt, 6) grass, and 7) water.
}
\label{fig:terra_visual_results}
\end{figure*}

Below we provide a brief overview of the Terra 3DSG method to aid in interpretation of the results in this case study. 
For further details and information, we refer the reader to \cite{samuelsonTerra2025}. 
The Terra 3DSG technique is split into three phases: 1) metric-semantic mapping, 2) task-agnostic 3DSG generation, and 3) task-driven 3DSG querying and navigation.
A visual example of this process is shown in Figure~\ref{fig:terra_visual_results}.

\subsection{METRIC-SEMANTIC MAP}
\label{subsec:terra_msmap}

The Terra metric-semantic map (MS-Map) constructs a global voxelized point cloud using LIO-SAM~\cite{slam_liosam2020shan} and enriches it with open-set semantics from images (see Figure~\ref{fig:terra_visual_results}(a)-(b)). 
Terrain labels are predicted with a trained terrain segmentation model (YOLO~\cite{yolov8_ultralytics}) and the class text names are embedded with CLIP, while class-agnostic object masks are extracted using FastSAM~\cite{zhao2023fastsam} and the image masks are then embedded with CLIP. 
LiDAR points are projected into each frame and associated with overlapping CLIP embeddings, which are aggregated across frames merging those embeddings above a cosine similarity threshold, $\tau_{match}$, or adding novel embeddings--while each global point maintains counts of its associated embeddings over time.

\subsection{TASK-AGNOSTIC 3DSG}
\label{subsec:terra_3dsg}

Given the MS-Map, Terra constructs task-agnostic place and region node layers by extracting place nodes and connectivity from terrain-specific generalized Voronoi diagram (GVD)~\cite{lauEfficientgridbased2013} edges (see Figure~\ref{fig:terra_visual_results}(c)). 
The semantic embedding of each place node is obtained by averaging the CLIP embeddings of all images in which the node appears within the camera's field of view.
Region nodes are formed by hierarchically clustering place nodes using either agglomerative or spectral clustering techniques based on semantic and geometric consistency, as described in \cite{samuelsonTerra2025} (see Figure~\ref{fig:terra_visual_results}(d)). 
Each region node's semantic embedding is computed as the average of its children node embeddings.
Together, the MS-Map and the resulting place and region node layers form the core of Terra's task-agnostic 3DSG design.

\subsection{TASK-DRIVEN 3DSG QUERYING AND NAVIGATION}
\label{subsec:terra_query_nav}

The task-agnostic Terra 3DSG supports a wide range of outdoor autonomous tasks such as object retrieval, region querying, and terrain-aware navigation using natural language queries embedded with CLIP.
Task-relevance is determined using cosine similarity between the query and the 3DSG above a similarity threshold, $\alpha$.
For object retrieval, Terra either operates directly on the MS-Map by extracting and clustering~\cite{dbscan_ester1996density} task-relevant points into bounding boxes, or leveraging the 3DSG by selecting relevant region and place nodes before extracting and clustering nearby points~\cite{bentley1975kdtrees}.
In both MS-Map and 3DSG detection methods, two aggregation strategies are used: average, which averages associated point embeddings, and max, which selects the single embedding with the highest occurrence count for that point. 
For region querying, Terra identifies the top-$k$ relevant region nodes. 
Within each selected region, only the query-relevant child place nodes are retained, defining the region to be monitored.
For navigation, paths are planned using A* to target place nodes, where targets are obtained through object retrieval and traversal costs between place nodes may be weighted by terrain preferences and constraints.

\subsection{TERRA PARAMETERS}
\label{subsec:terra_params}

Unless explicitly stated, the parameters used for Terra 3DSG construction are listed in detail in the Appendix.

\section{DATASETS AND ROBOT PLATFORM}
\label{sec:datasets_robot_platform}

\subsection{DATASETS}
\label{subsec:datasets}

\begin{table*}[t]
\centering
\caption{ Dataset Summaries }
\begin{tabular}{p{75pt}p{40pt}p{70pt}p{35pt}p{80pt}p{115pt}}
\hline
\textbf{Dataset} & \textbf{Trajectory Length} $[m]$ & \textbf{Environment Type} & \textbf{Vegetation} & \textbf{Terrain Types} & \textbf{Perceptual Challenges} \\
\hline

\textbf{River Park} & 
951.56 & 
Recreational park & 
Dense & 
Leaves, road, sidewalk, dirt, grass, water & 
Strong lighting contrast, canopy occlusions, large and small objects, trajectory on a variety of terrains \\
\hline

\textbf{Nunns Park} & 
1238.27  & 
Riverside recreational area & 
Sparse & 
Leaves, road, sidewalk, gravel, dirt, grass, water &
Leaf-covered surfaces, underpass shadowing \\
\hline

\textbf{Marina Part 1} & 
3004.28 & 
Marina / RV park & 
Sparse & 
Leaves, road, sidewalk, gravel, dirt, grass, water & 
Open geometry, dynamic vehicles, large range of object sizes \\
\hline

\textbf{Marina Part 2} & 
1269.63 & 
Marina &
Sparse & 
Leaves, road, gravel, dirt, grass, water &
Road-gravel transition, limited vertical structure \\
\hline

\textbf{Rock Canyon Campground} & 
1151.43 & 
Forested trail &
Dense & 
Leaves, gravel, dirt, grass & 
Repetitive natural geometry, mixed-surface trails, ambiguous objects \\
\hline
\multicolumn{6}{p{490pt}}{
Feature comparisons of the five datasets collected for evaluating Terra. 
The datasets cover a multitude of diverse environments, terrain complexities, lengths, and lighting conditions to provide a reliable evaluation test bed for Terra. 
There were a total of seven terrain types, and the overall combined trajectory length was $7615.17~m$.
}
\end{tabular}
\label{tab:dataset_comparison}
\end{table*}

\begin{figure*}[p]
\centering
\setlength{\tabcolsep}{6pt}
\begin{tabular}{cc}
\begin{minipage}[t]{0.48\textwidth}
\vspace{0pt}
\begin{tabular}{@{}c c@{}}
\includegraphics[width=0.57\textwidth]{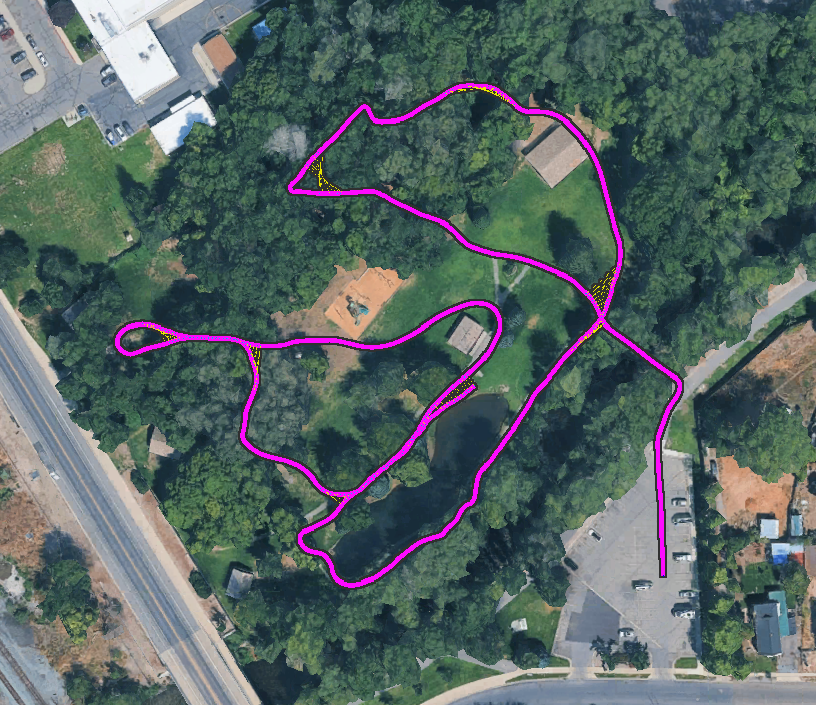} &
\end{tabular}
\begin{tabular}{@{}c@{}}
\includegraphics[width=0.33\textwidth]{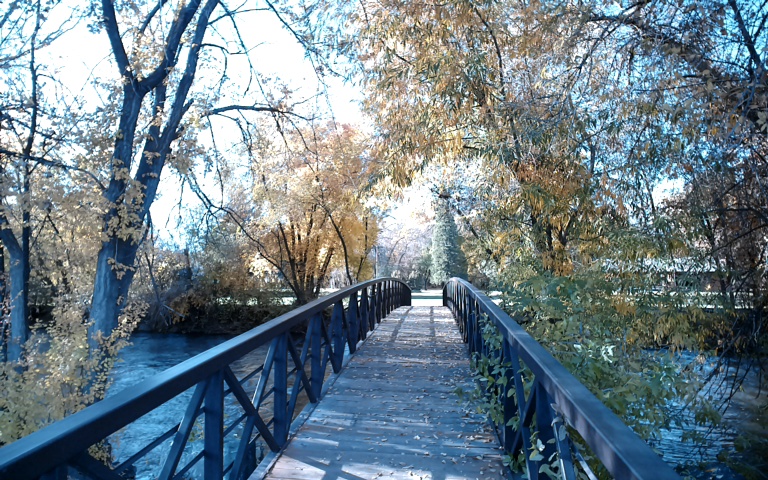}\\
\includegraphics[width=0.33\textwidth]{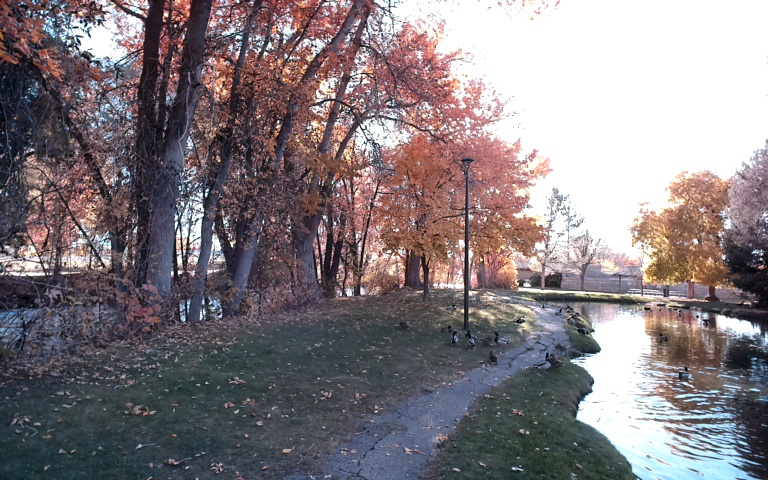}\\
\includegraphics[width=0.33\textwidth]{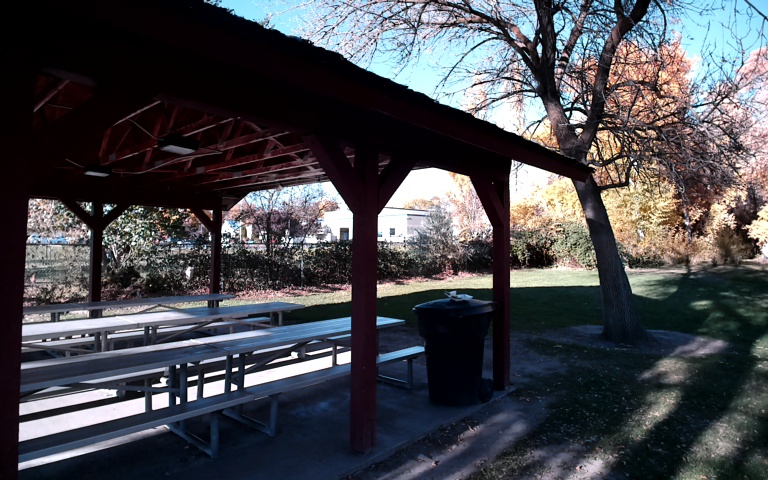}
\end{tabular}
\centering
{\fontsize{8}{9.5}\selectfont\rmfamily (a) River Park}\label{fig:provo_river_images}
\end{minipage}
&
\begin{minipage}[t]{0.48\textwidth}
\vspace{0pt}
\begin{tabular}{@{}c c@{}}
\includegraphics[width=0.57\textwidth]{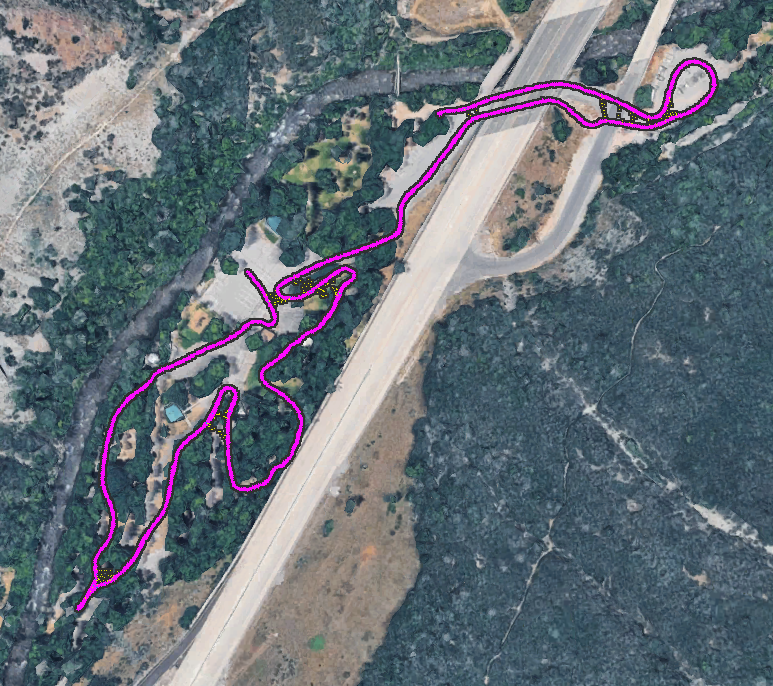} &
\end{tabular}
\begin{tabular}{@{}c@{}}
\includegraphics[width=0.33\textwidth]{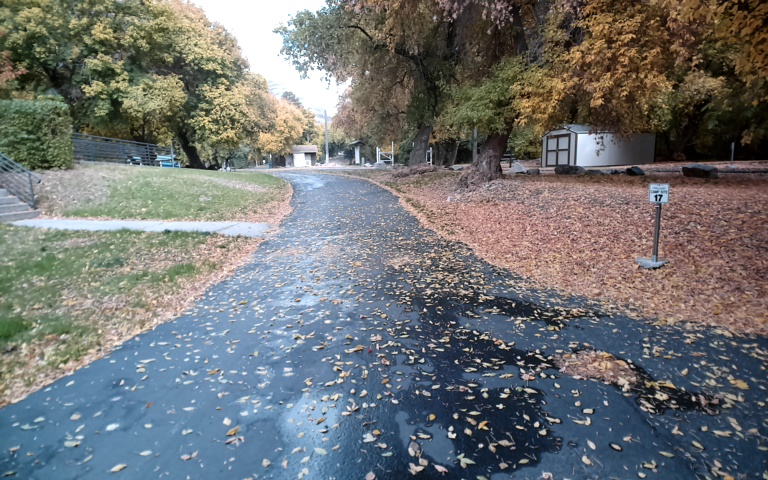}\\
\includegraphics[width=0.33\textwidth]{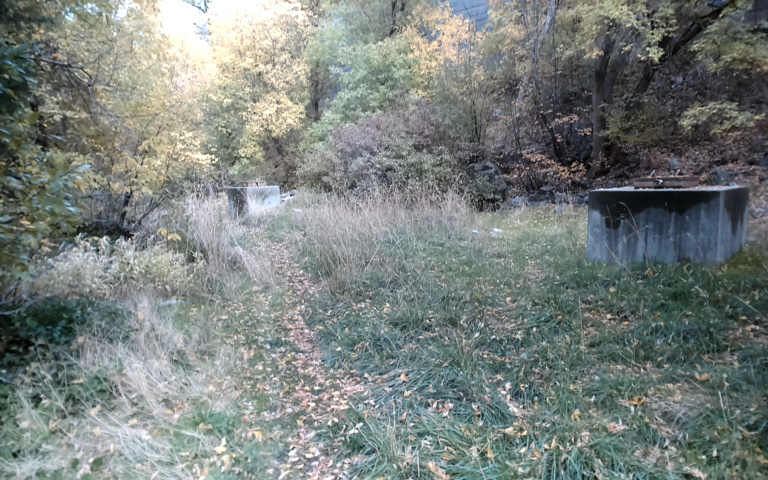}\\
\includegraphics[width=0.33\textwidth]{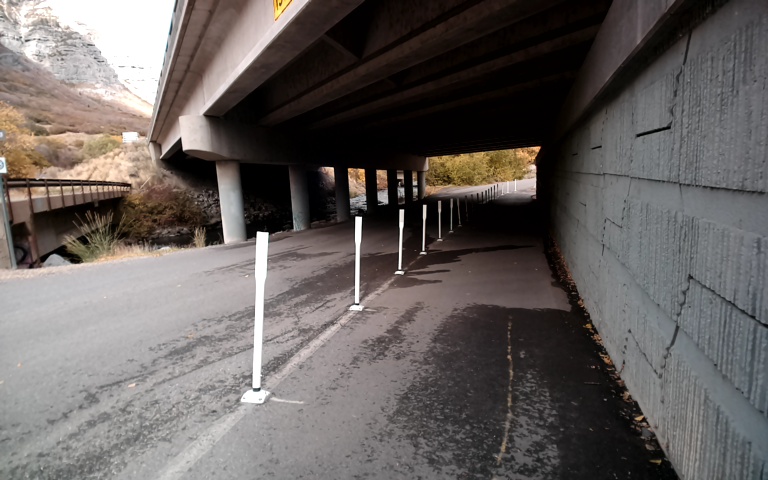}
\end{tabular}
\centering
{\fontsize{8}{9.5}\selectfont\rmfamily(b) Nunns Park}\label{fig:nunns_park_images}
\end{minipage}
\\[6pt]

\begin{minipage}[t]{0.48\textwidth}
\vspace{0pt}
\begin{tabular}{@{}c c@{}}
\includegraphics[width=0.57\textwidth]{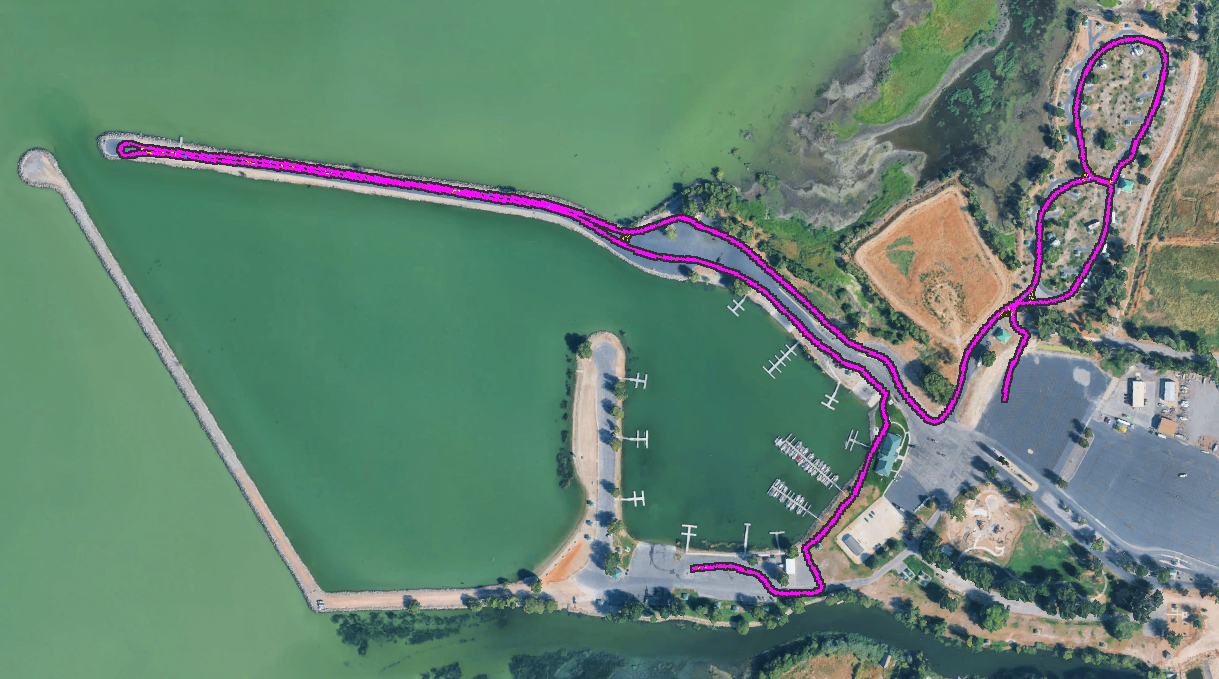} &
\end{tabular}
\begin{tabular}{@{}c@{}}
\includegraphics[width=0.33\textwidth]{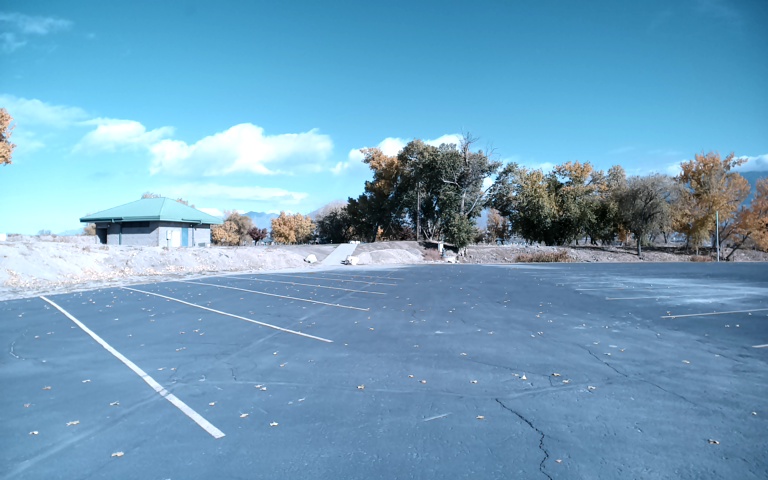}\\
\includegraphics[width=0.33\textwidth]{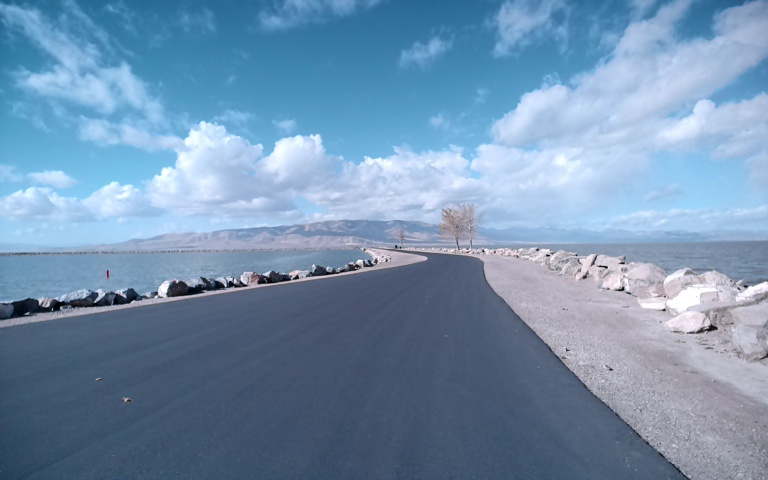}\\
\includegraphics[width=0.33\textwidth]{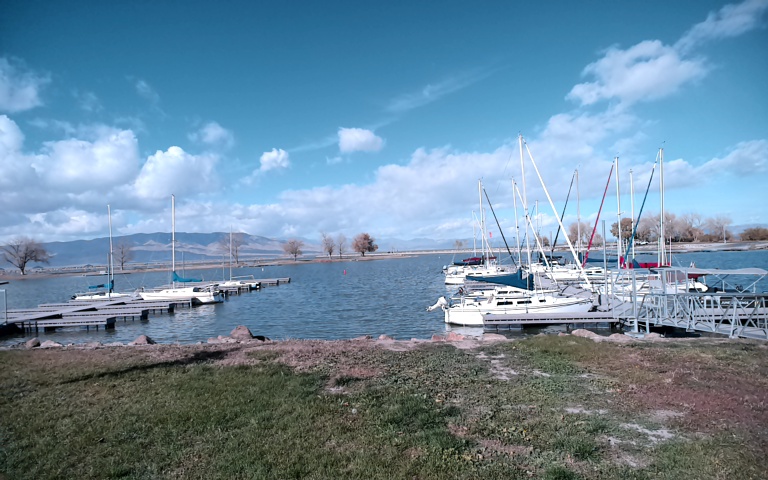}
\end{tabular}
\centering
{\fontsize{8}{9.5}\selectfont\rmfamily(c) Marina Part 1}\label{fig:marina_p1_images}
\end{minipage}
&
\begin{minipage}[t]{0.48\textwidth}
\vspace{0pt}
\begin{tabular}{@{}c c@{}}
\includegraphics[width=0.57\textwidth]{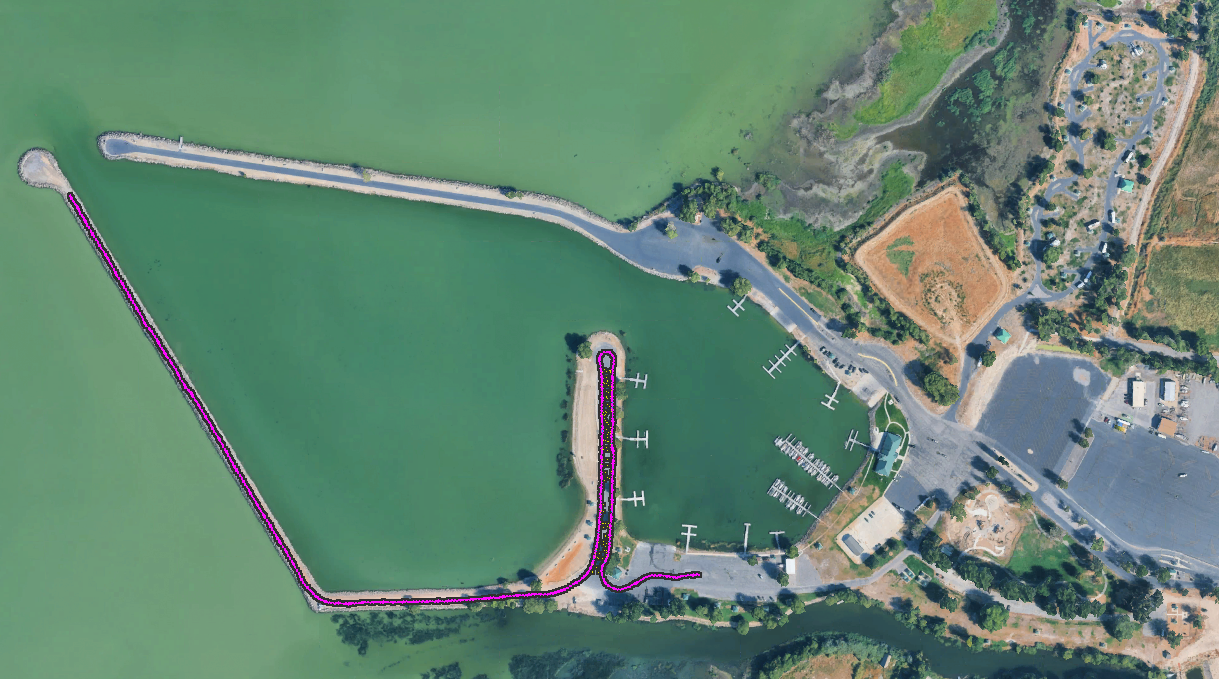} &
\end{tabular}
\begin{tabular}{@{}c@{}}
\includegraphics[width=0.33\textwidth]{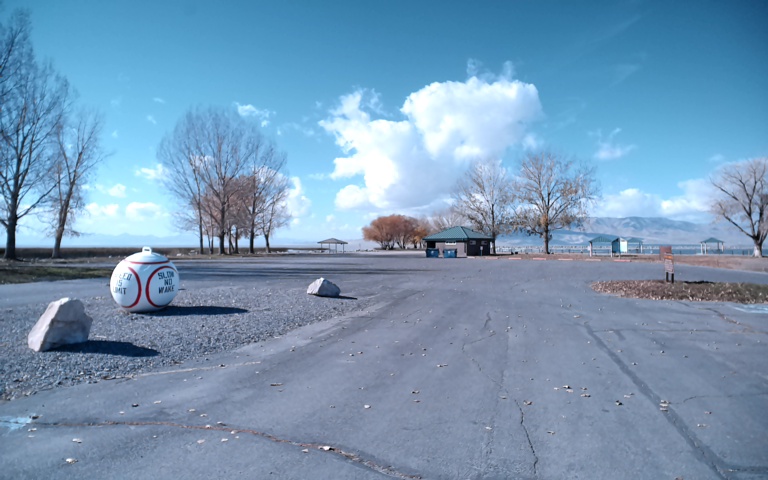}\\
\includegraphics[width=0.33\textwidth]{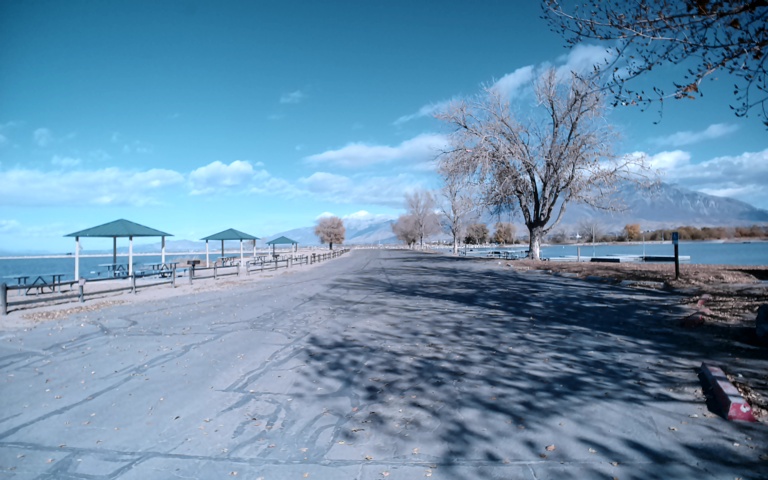}\\
\includegraphics[width=0.33\textwidth]{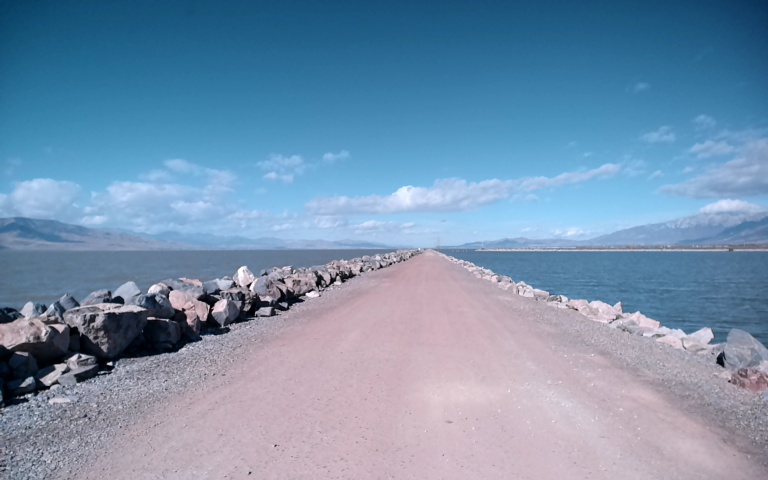}
\end{tabular}
\centering
{\fontsize{8}{9.5}\selectfont\rmfamily(d) Marina Part 2}\label{fig:marina_p2_images}
\end{minipage}
\\[6pt]

\multicolumn{2}{c}{
\begin{minipage}[t]{0.48\textwidth}
\vspace{0pt}
\begin{tabular}{@{}c c@{}}
\includegraphics[width=0.57\textwidth]{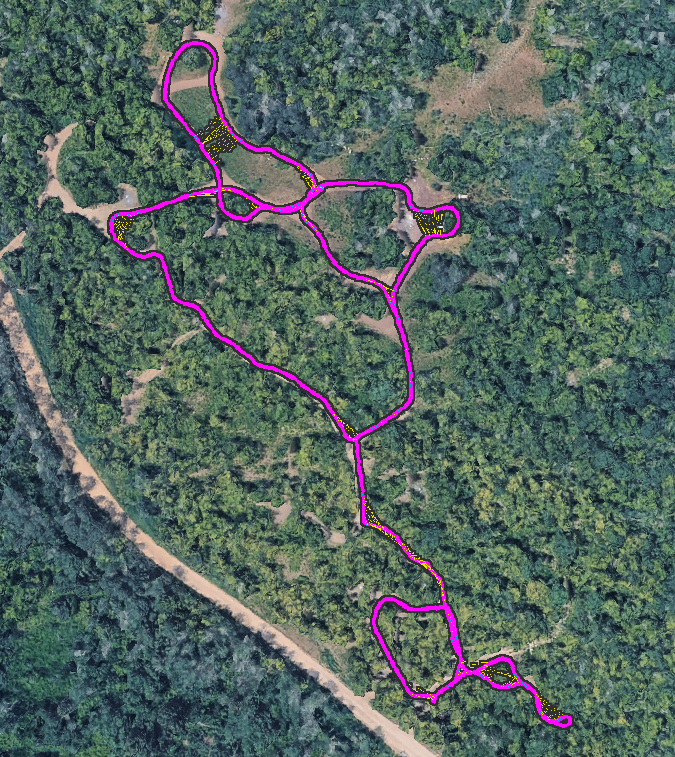} &
\end{tabular}
\begin{tabular}{@{}c@{}}
\includegraphics[width=0.33\textwidth]{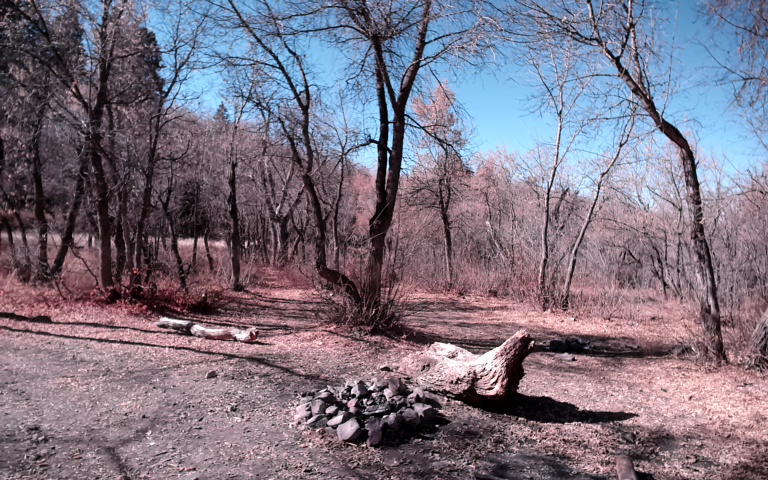}\\
\includegraphics[width=0.33\textwidth]{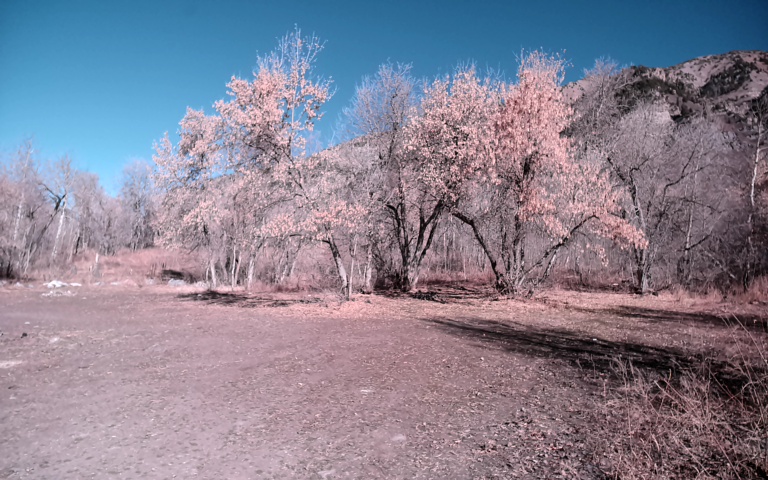}\\
\includegraphics[width=0.33\textwidth]{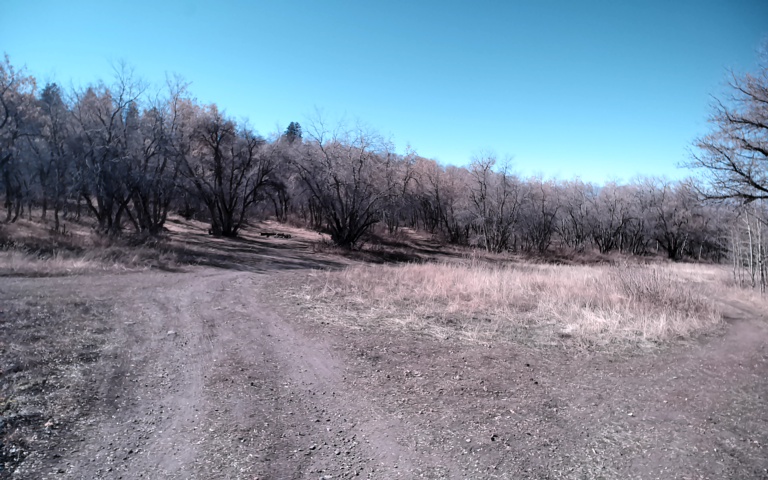}
\end{tabular}
\centering
{\fontsize{8}{9.5}\selectfont\rmfamily(e) Rock Canyon Campground}\label{fig:rock_canyoun_campground_images}
\end{minipage}
}
\end{tabular}
\caption{Overview of the five field-test datasets. 
For each dataset, the satellite image (left) shows the spatial extent of the trajectory, while the three camera images (right) illustrate representative visual conditions encountered during data collection.
} \label{fig:dataset_visual_overview}
\end{figure*}

To rigorously evaluate 3DSGs and their components under realistic deployment conditions, we collected five large-scale outdoor datasets spanning a diverse range of natural environments, terrain types, and perceptual challenges. 
The datasets were designed to stress different aspects of outdoor 3DSG construction, including terrain ambiguity, traversability complexities, robustness to lighting and vegetation variability, scalability with trajectory length, and consistency under repeated exposure to similar environments. 
Together, these datasets provide a systematic testbed for analyzing when and why VLM-based 3DSGs succeed or fail in unstructured outdoor settings.
A 3DSG using Terra was built for each of the datasets, with an example shown in Figure~\ref{fig:terra_visual_results} for the River Park dataset.
It is of note that in contrast to \cite{samuelsonTerra2025}, which used a single camera, data was collected with three cameras to increase semantic field of view.

Table~\ref{tab:dataset_comparison} summarizes the key characteristics of each dataset, including trajectory length, environment type, vegetation density, terrain diversity, and perceptual challenges.
Figure~\ref{fig:dataset_visual_overview} provides a visual overview of each dataset, showing satellite imagery with the robot trajectory estimated from LIO-SAM overlaid. 
To the right of each satellite image, three representative images are shown to illustrate the visual diversity and environmental complexity present in each scene.

The River Park dataset encompasses a parking lot, recreational park space, pond, and a bridge crossing over a river (see Figure~\ref{fig:dataset_visual_overview}(a)). 
The environment exhibits strong lighting variation due to dense tree canopy and sun glare, as well as terrain ambiguity caused by narrow paved paths and leaf-covered ground surfaces. 
The environment also contains a river with a bridge as well as a stream in the back woods introducing challenges for navigation task traversability. 
Moderate pedestrian activity further introduces dynamic occlusions. 
The presence of distinct functional areas within a compact space makes this dataset well suited for evaluating hierarchical region segmentation. 
This dataset primarily evaluates 3DSG's robustness to lighting variation, terrain ambiguity, and region construction under partial observability.

The Nunns Park dataset was collected in a recreational camping area adjacent to a river (see Figure~\ref{fig:dataset_visual_overview}(b)). 
Similar to River Park, this environment presents terrain ambiguity due to leaf-covered road surfaces and significant lighting contrast introduced by a highway underpass. 
The highway underpass additionally causes GPS dropout, enabling future evaluation of 3DSG localization and navigation in GPS-denied conditions. 
This scene contains a large number and wide variety of objects, enabling more extensive object-level evaluation. 
This dataset primarily evaluates 3DSG's ability to robustly detect object representations while accurately tracking terrain under occlusion and variable lighting conditions.

The Marina Part 1 and 2 datasets were collected from two adjacent regions of the same marina environment. 
Marina Part 1 includes an RV park, an extended parking lot, boat docks, and a long breakwater segment (see Figure~\ref{fig:dataset_visual_overview}(c)), while Marina Part 2 consists primarily of boat docks and a gravel breakwater (see Figure~\ref{fig:dataset_visual_overview}(d)). 
These environments are characterized by wide-open spaces, long-range visibility, and limited natural region boundaries. 
Together, the marina datasets evaluate 3DSG's ability to reason over large open areas, distinguish subtle terrain transitions, and construct meaningful regions in the absence of strong geometric or semantic dividers. 
In addition, the RV park of the Marina Part 1 dataset is used to evaluate scene graph consistency over several runs.

The Rock Canyon Campground dataset is a remote, trail-dominated environment, that covers a large area with increased terrain diversity, scattered open grass regions, and a small number of man-made structures such as fire pits (see Figure~\ref{fig:dataset_visual_overview}(e)). 
The scene contains few distinctive objects, ambiguous region structure and traversable terrain largely confined to the primary trail.
This dataset evaluates 3DSG's ability to maintain semantic and region
consistency over long trajectories, interpret ambiguous regions at scale, and track traversable terrain across narrow and intermittently branching paths.

Terrain semantics for all datasets were inferred using a YOLO terrain classification model trained on the GOOSE~\cite{mortimer2024goose} and GOOSE-Ex~\cite{hagmanns2024gooseEx} datasets. 
We chose only terrain-relevant classes to distinguish between common outdoor surface types: leaves, road, sidewalk, gravel, dirt, grass, and water.
We note that normally leaves are not a terrain class, but these datasets were collected in the fall and winter seasons when the leaves were mostly on the ground.

\subsection{ROBOT PLATFORM}
\label{robot_platform}

\begin{figure}
\centering

\begin{minipage}{\columnwidth}
\centering
\includegraphics[width=0.85\columnwidth]{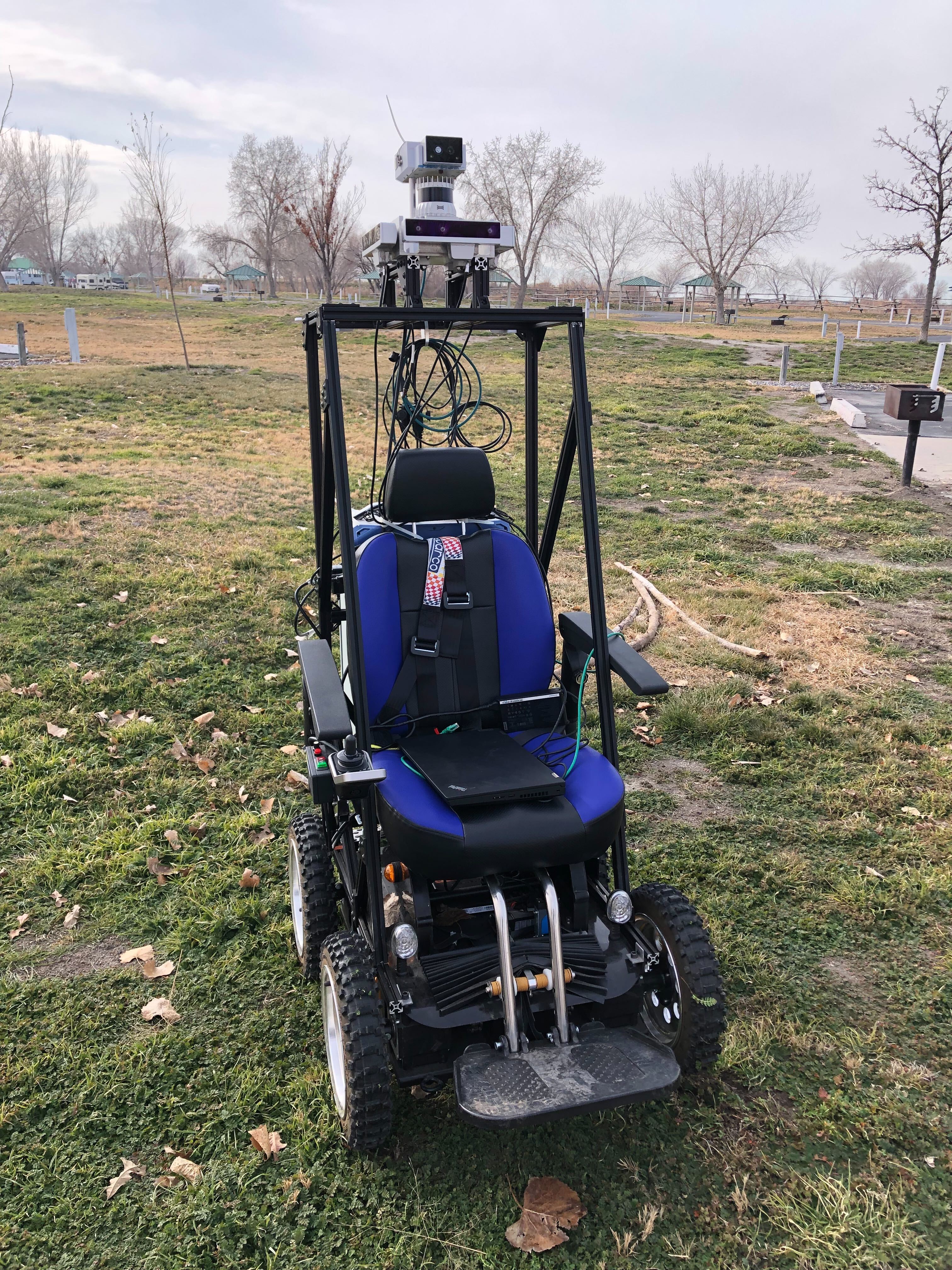}
\par\smallskip
{\fontsize{8}{9.5}\selectfont\rmfamily(a) Robotic Wheelchair Platform}
\end{minipage}

\medskip

\begin{minipage}{\columnwidth}
\centering
\includegraphics[width=0.85\columnwidth]{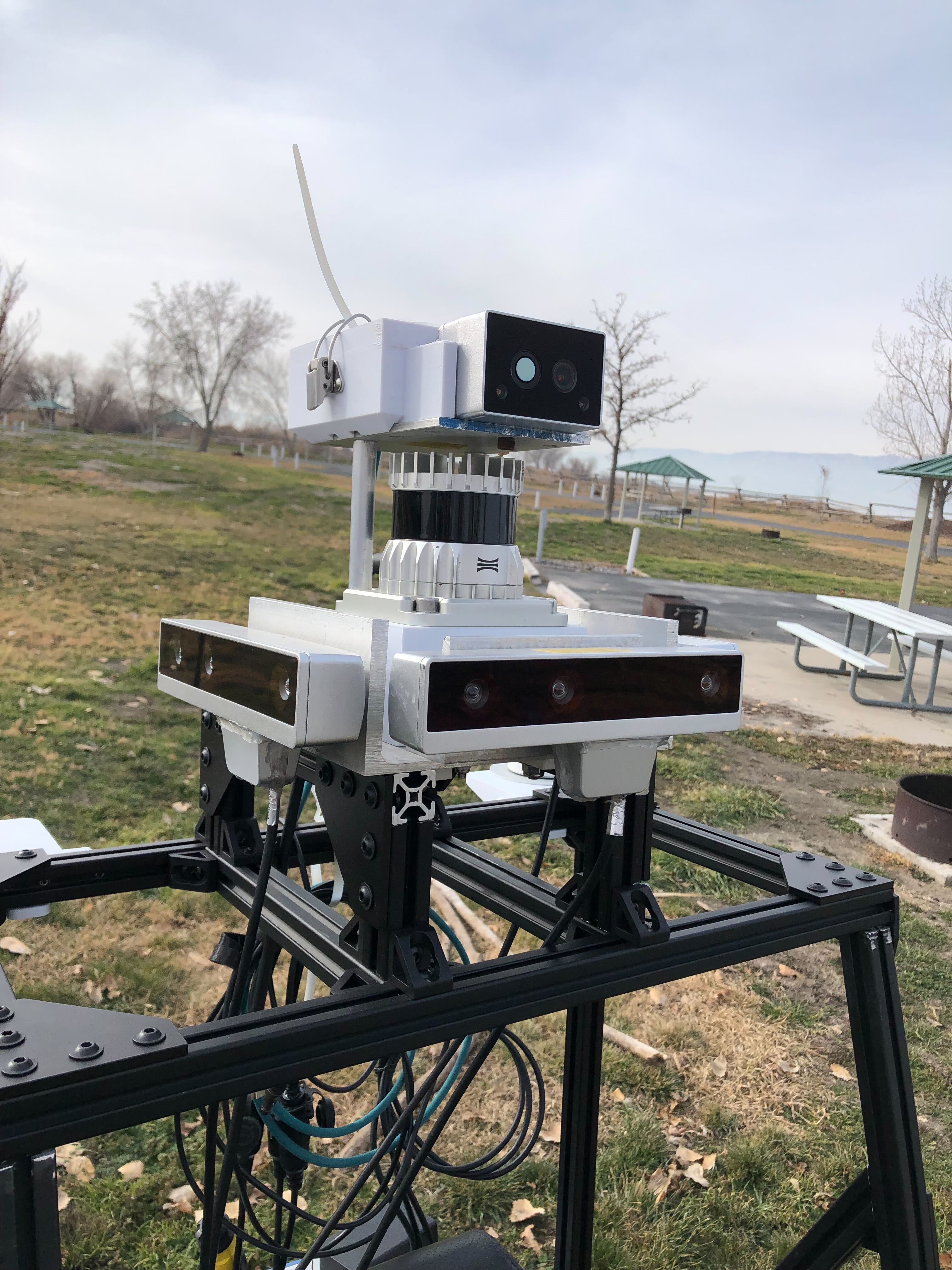}
\par\smallskip
{\fontsize{8}{9.5}\selectfont\rmfamily(b) Sensor Platform}
\end{minipage}

\caption{(a) Image of our robotic wheelchair platform. (b) Depicts the rigid sensor platform used for data collection. Only the 3 OAK-D LR cameras positioned at 90 degree angles from each other, the Ouster OS1 128-beam LiDAR, and 6-axis IMU were used.}
\label{fig:wheelchair_overall}
\end{figure}

All datasets were gathered with our custom robotic wheelchair platform (Figure~\ref{fig:wheelchair_overall}(a)). 
This platform consists of a rigid sensor platform mounted upon a motorized wheelchair (Figure~\ref{fig:wheelchair_overall}(b)).
The motorized wheelchair was manually driven to collect these datasets, with future plans to develop autonomous navigation capabilities.
The rigid sensor platform has three OAK-D LR cameras positioned at 90 degree angles from each other, and an Ouster OS1 128-beam LiDAR with an integrated 6-axis IMU. 
While this sensor platform has many other sensors on board, we only used the three OAK-D LR cameras and the Ouster LiDAR. 
It is noted that the stereo capability of the OAK-D LR cameras is not utilized; only a single monocular image is used from each camera, and all depth information is obtained exclusively from the LiDAR.
The laptop used to collect the data from the sensor platform was an Intel Core i7-10850H CPU, NVIDIA Quadro RTX 3000 Mobile GPU with 32 GB of RAM.
The OAK-D LR cameras recorded at $10$Hz, the LiDAR at $10$Hz, and the IMU at $100$Hz.

The LiDAR was time synced to the laptop via PTP. 
The OAK-D cameras' onboard DepthAI driver automatically syncs to the laptop's clock below $200 \mu s$ of accuracy. 
Camera intrinsics were computed using both the OAK-D provided intrinsics and the Kalibr calibration library \cite{Furgale2013UnifiedCalibration, Rehder2016Kalibr}. 
Extrinsics between cameras and LiDAR were manually computed via visual inspection by overlapping LiDAR scans on images and repeated for each dataset to ensure accuracy.

\subsection{FIELD TESTING OUTLINE}

The following sections present an in-depth evaluation of the core layers commonly found in modern 3DSG frameworks.
Although the experiments are conducted using Terra-generated 3DSGs, the diversity and complexity of the evaluated outdoor datasets, together with the general hierarchical structure of Terra, enable broader insights into the behavior of 3DSGs and VLM-based semantic reasoning in complex outdoor environments.
The analysis progresses hierarchically through the representation, beginning with low-level VLM point embeddings, followed by navigation over the place-node graph toward semantically detected objects, then region-level semantic understanding using VLM reasoning, and finally concluding with an evaluation of the overall utility of 3DSGs for real-world field robotic applications.

\section{ANALYSIS OF VLM POINT EMBEDDINGS IN OUTDOOR ENVIRONMENTS}
\label{sec:point_embs_analysis}

Most evaluations of VLMs in robotics focus either on indoor environments or on downstream robotic tasks.
However, before assessing downstream task performance, it is important to first understand the structure and behavior of the VLM embeddings themselves in outdoor environments.

A common pipeline for incorporating VLM semantics into robotic maps begins by oversegmenting an image using a vision foundation model such as FastSAM.
Ideally, each generated mask corresponds to a distinct entity in the scene, such as an individual car or tree.
Each masked region is then encoded by a VLM, such as CLIP, producing a semantic embedding for that entity.
The resulting embedding is associated with all 3D map points that project within the corresponding image mask.
As the robot traverses the environment and observes the same scene from multiple viewpoints, each map point accumulates a set of associated semantic embeddings.
Under ideal conditions, geometrically consistent map points should exhibit relatively consistent semantic embeddings over time.

In practice, however, segmentation outputs are often inconsistent due to factors such as viewing distance, boundary ambiguity, occlusion, lighting variation, and the stochastic behavior of foundation models.
As a result, the semantic embeddings associated with a single map point may exhibit substantial variation across observations.
The following analyses investigate this phenomenon across the five outdoor datasets collected in this study.

\subsection{PRESENCE OF OUTLIER EMBEDDINGS}
\label{subsec:outliers}

We first investigate the extent to which embedding outliers exist by inspecting the specific embeddings mapped to each 3D map point in the LiDAR map.

\subsubsection{Metrics}
\label{subsubsec:outlier_metrics}

Outliers are commonly detected using the modified z-score approach designed for unimodal data that does not follow a Gaussian distribution due to skew or other factors~\cite{modzscore_iglewicz1993outliers}.
The modified z-score is defined only for one-dimensional data.

Let $X$ be the set of associated semantic embedding vectors with $x_i \in X$ representing the $i$-th embedding vector in the set $X$.
To get a one-dimensional distribution of the embedding associations for a single point, we first compute the mean of the associated embedding vectors, $\mu$.
We then compute the cosine distance between each embedding and this mean defined as
\begin{equation}
    d_i = 1 - \text{cos\_sim}(x_i, \mu),
\end{equation}
where $\text{cos\_sim}$ refers to the cosine similarity between $x_i$ and $\mu$.
This set of distances corresponding to the set of associated embedding vectors represents a unimodal distribution centered near zero (i.e., embeddings near the mean vector) with distances ranging up to $1$.

Given this distribution of embedding distances, we follow the traditional modified z-score equations in~\cite{modzscore_iglewicz1993outliers} defined as
\begin{equation}
    z_{mod,i} = 0.6745 \frac{d_i -\mu_{med}}{\text{MAD}},
\end{equation}
where $\mu_{med}$ is the median distance of the 1D distribution and the median absolute deviation (MAD) is defined as
\begin{equation}
    \text{MAD} = \text{median}( | d_0 - \mu_{med} |, | d_1 - \mu_{med} |, \cdots ).
\end{equation}
The $i$-th embedding is then said to be an outlier if $| z_{mod,i} | > 3.5$, where $3.5$ is the commonly used outlier threshold as discussed in~\cite{modzscore_iglewicz1993outliers}.

We now define the outlier ratio for a single map point given its set of semantic embeddings as the total number of outliers divided by the size of the embedding set $X$ (i.e. $|\text{outliers}| / |X|$).

\subsubsection{Results}
\label{subsubsec:outlier_results}

\begin{figure}[t]
    \centering
    \includegraphics[width=0.99\columnwidth]{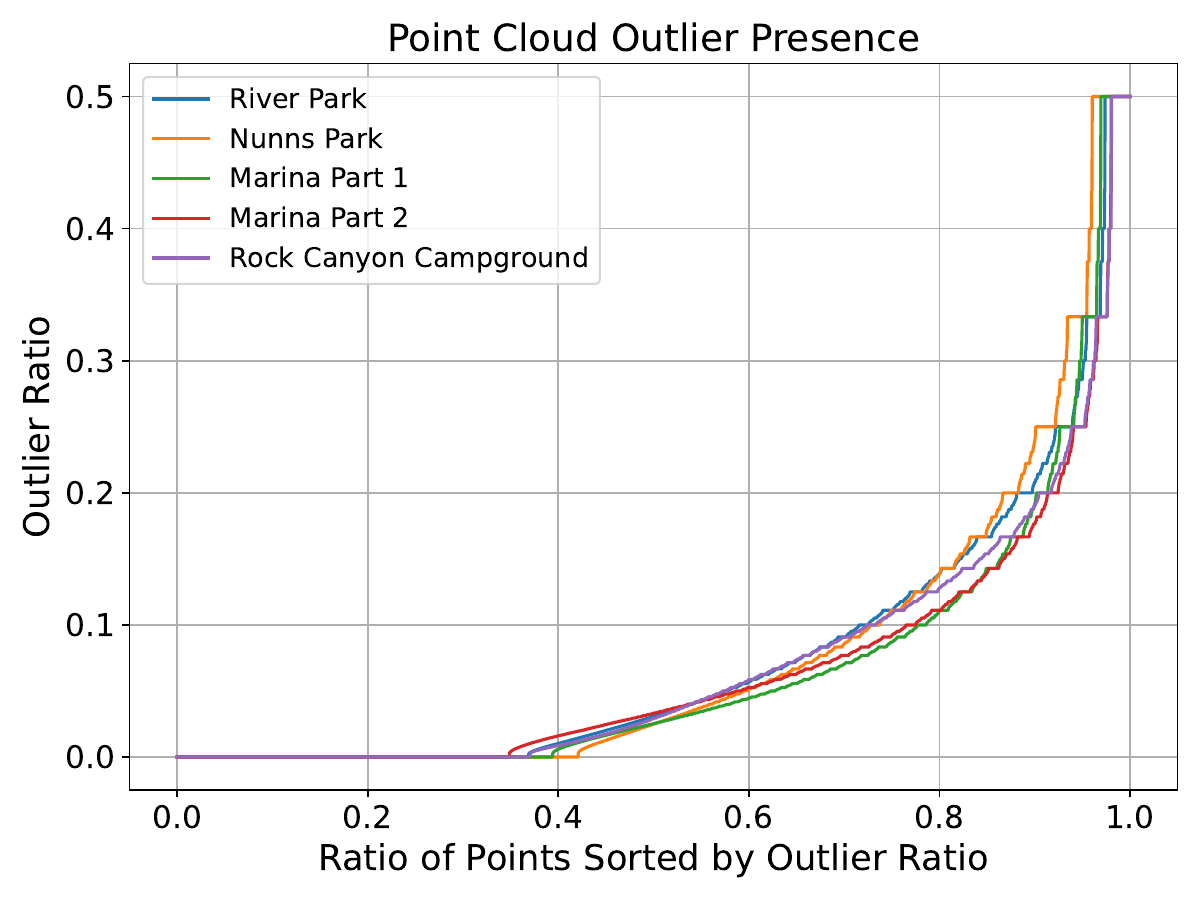}
    \vspace{-0.5cm}
    \caption{Outlier ratio computed for every point across all five datasets.}
    \label{fig:outlier_ratio_exp}
\end{figure}

 Figure~\ref{fig:outlier_ratio_exp} shows the outlier ratio computed for every semantic point in our dataset point clouds.
 Note that the number of associated embeddings for a single point range from one to over one thousand embeddings and the number of points in these point clouds are mostly around 200,000.
 While the trend across all datasets shows a majority of the points having relatively few outliers compared to the total number of associated embeddings, there are still around $30\%$ of point cloud points with outlier ratios above $0.1$.
 This demonstrates that either outliers must be explicitly considered or the embeddings associated with individual points may not satisfy the unimodal assumption.

\subsection{MULTIPLE EMBEDDING MODES}
\label{subsec:multi_modes}

The previous analysis evaluated embedding consistency under the assumption that the associated embeddings for a single map point form a unimodal distribution.
However, in outdoor environments this assumption may not always hold.
Due to inconsistent segmentation boundaries, partial observations, occlusions, lighting changes, and viewpoint-dependent appearance variation, embeddings associated with a single geometric map point may instead form multiple distinct semantic modes.
For example, a point near the boundary between a tree and the sky may sometimes inherit embeddings corresponding to vegetation and other times inherit embeddings corresponding to the sky.
Similarly, distant objects or partially occluded regions may produce substantially different segmentation masks across observations, resulting in multiple semantically distinct embedding groups being associated with the same map point.

To investigate whether these multiple mode embedding distributions occur in practice, we cluster the set of embeddings associated with each map point using HDBSCAN~\cite{hdbscan_campello13, hdbscan_campello15, hdbscan_code17}.
Under this formulation, multiple dense embedding clusters associated with a single point indicate the presence of multiple semantic modes.

\subsubsection{Metrics}
\label{subsubsec:multi_modes_metrics}

To evaluate the quality of the generated clusters, we use the silhouette score, a standard metric for unsupervised clustering evaluation~\cite{rousseeuw1987silhouettes}.
The silhouette score measures how well a data point matches its assigned cluster relative to neighboring clusters.

For each map point, a silhouette score is computed using the clustering assignments of its associated embeddings.
These per-point scores are then averaged across all points in a dataset to obtain a global clustering quality metric.
The silhouette score ranges from $-1$ to $1$, where values near $1$ indicate well-separated clusters, values near $0$ indicate overlapping cluster boundaries, and negative values indicate poor clustering assignments.

\subsubsection{Results}
\label{subsubsec:multi_modes_results}

We initially compute the silhouette score of the generated HDBSCAN clusters per point and average the scores across all points in the dataset.
The averaged dataset silhouette scores are as follows: $0.493$ (River Park), $0.500$ (Nunns Park), $0.460$ (Marina Part 1), $0.479$ (Marina Part 2), $0.496$ (Rock Canyon Campground).

Across all datasets, the average silhouette scores remain consistently near $0.5$, suggesting that the discovered embedding clusters are reasonably well separated.
These results indicate that the embedding associations for many map points are not purely noisy variations around a single semantic representation, but instead frequently exhibit multiple distinct semantic modes.
This behavior suggests multiple modes in semantic associations are a characteristic of outdoor, open-set VLM-based mapping pipelines.

\begin{figure}
\centering

\begin{minipage}{\columnwidth}
\centering
\includegraphics[width=0.95\columnwidth]{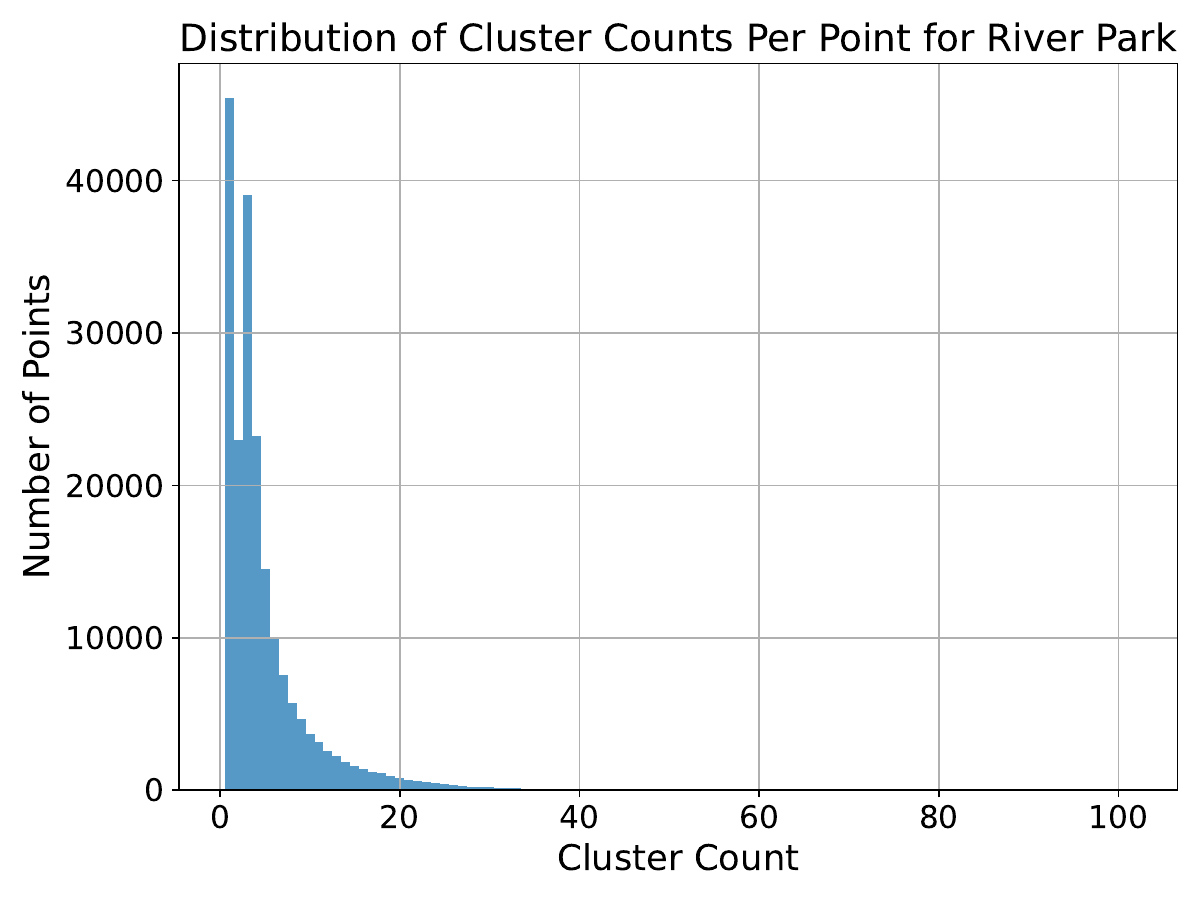}
\par\smallskip
{\fontsize{8}{9.5}\selectfont\rmfamily(a)}
\end{minipage}

\medskip

\begin{minipage}{\columnwidth}
\centering
\includegraphics[width=0.95\columnwidth]{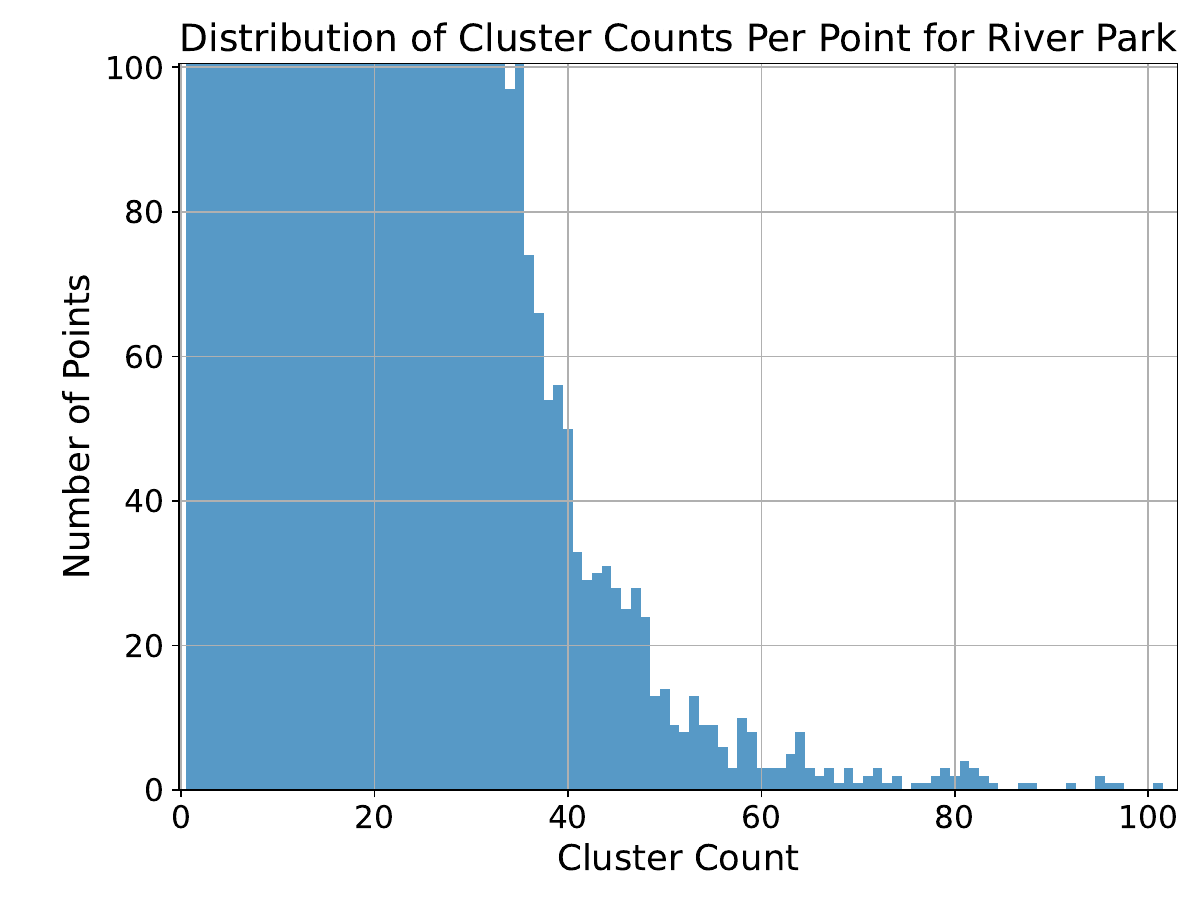}
\par\smallskip
{\fontsize{8}{9.5}\selectfont\rmfamily(b)}
\end{minipage}

\caption{
Histogram of the number of embedding clusters associated with map points in the River Park dataset.
Figure (b) shows a zoomed-in view of (a).
While most points contain fewer than 10 clusters, some points exhibit up to 100 clusters, demonstrating the significant presence of multiple semantic modes within the dataset.
}
\label{fig:cluster_distribution}
\end{figure}

\begin{figure}[t]
    \centering
    \includegraphics[width=0.99\columnwidth]{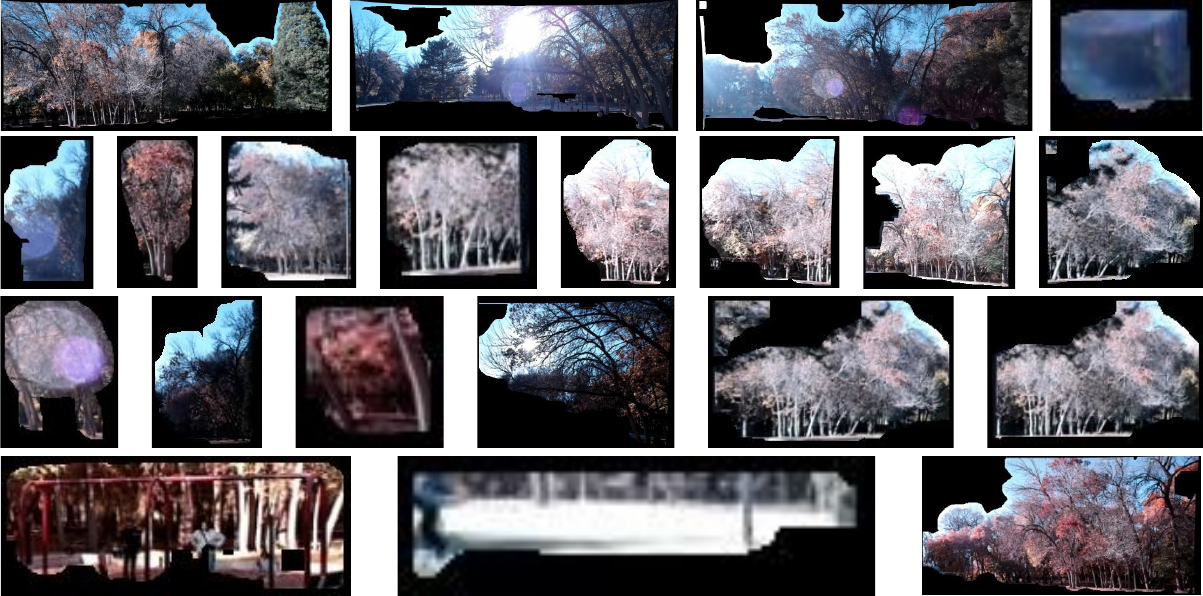}
    \vspace{-0.5cm}
    \caption{
    Representative masked image from each of 21 embedding clusters associated with a single point in the River Park dataset.
    This example highlights the diversity of visual observations for the same map point.
    }
    \label{fig:point_img_clusters}
\end{figure}

\begin{figure}
\centering

\begin{minipage}{\columnwidth}
\centering
\includegraphics[width=0.95\columnwidth]{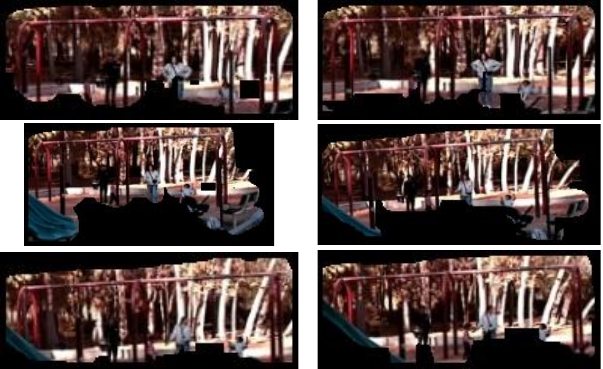}
\par\smallskip
{\fontsize{8}{9.5}\selectfont\rmfamily(a) Cluster of swing set}
\end{minipage}

\medskip

\begin{minipage}{\columnwidth}
\centering
\includegraphics[width=0.95\columnwidth]{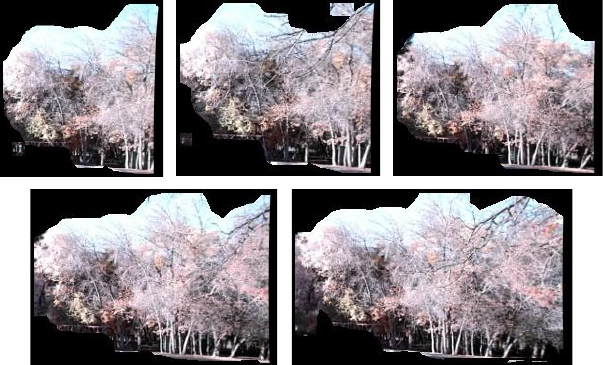}
\par\smallskip
{\fontsize{8}{9.5}\selectfont\rmfamily(b) Cluster of some dead trees}
\end{minipage}

\caption{
Masked images associated with two embedding clusters shown in Figure~\ref{fig:point_img_clusters}.
These examples qualitatively demonstrate the visual and semantic consistency present within individual clusters.
}
\label{fig:clustered_imgs}
\end{figure}

Figure~\ref{fig:cluster_distribution} illustrates the distribution of cluster counts across points in the River Park dataset.
Most points contain fewer than 10 embedding clusters (or semantic modes), although some points exhibit as many as 100 clusters (see Figure~\ref{fig:cluster_distribution}(b)).
While only the River Park distribution is shown, all datasets exhibit similar trends.

Figure~\ref{fig:point_img_clusters} provides qualitative evidence of the diversity of image regions associated with a single global map point.
The figure shows one representative masked image from each of the 21 clusters associated with a single point cloud point.
Embeddings labeled as noise by HDBSCAN are omitted from this visualization, as these embeddings may represent either true outliers or additional semantic modes not captured by the clustering.
Figures~\ref{fig:clustered_imgs}(a)-(b) further visualize all image masks belonging to two of these clusters.
These examples qualitatively demonstrate strong visual and semantic consistency within individual clusters despite substantial variation across clusters.  

\subsection{GENERAL TAKEAWAYS}
\label{subsec:vlm_emb_takeaways}

VLM embeddings associated with image masks in outdoor environments do not always follow a single, unimodal distribution for a given point.
Although most points exhibit relatively few embeddings identified as outliers, under the unimodal assumption, a meaningful portion of the data contains significant outlier behavior that may instead reflect multiple semantic modes.

Variations in viewpoint, observation distance, segmentation boundaries, occlusions, and other environmental factors may lead to multiple semantic embedding modes being associated with a single 3D map point.
These factors make multimodal embedding distributions an important consideration for outdoor 3DSGs that associate image-derived VLM embeddings with persistent 3D locations, particularly when SAM-like class-agnostic segmentation produces masks with varying semantic content across frames.
The number of modes observed for a point can range from only a few to over 100 depending on the scene and available imagery.
Our results show that HDBSCAN effectively identifies these distinct semantic modes for individual map points.

Future work should investigate how these multiple mode embedding distributions can be leveraged to improve semantic reasoning about environments and scene entities, rather than relying on naive approaches that average all embeddings into a single point representation.

\section{ANALYSIS OF 3DSG PLACE NODE GRAPHS FOR OUTDOOR NAVIGATION}
\label{sec:place_node_nav_analysis}

3DSGs have demonstrated the capability to detect open-set objects given natural language queries~\cite{samuelsonTerra2025}. 
Terra and related 3DSGs maintain a places layer of the graph enabling efficient path planning to navigate to detected objects.

Previous open-set 3DSG works have demonstrated navigation and planning primarily in indoor environments~\cite{werby23hovsg, guConceptGraphsOpenVocabulary2024, maggio2024Clio}, while outdoor open-set 3DSG research has largely focused on semantic mapping and object detection~\cite{samuelsonTerra2025, steinkeCollaborativeDynamic2025b}.
Strader et al.~\cite{strader2025language} more recently demonstrated language-grounded planning and execution for outdoor autonomous navigation using multirobot 3DSGs and open-set object maps.
In contrast, this analysis evaluates the embedding-based semantic representation and terrain-derived place-node graph used by Terra for contextualized object retrieval and navigation across diverse, often hard-to-traverse, outdoor environments.

We also evaluate the benefit of providing contextual information to specify which object in the environment we want to navigate to. 
Since this navigation experiment was run months after building the initial 3DSGs, we chose to navigate to static objects and thus the Rock Canyon dataset was omitted from this evaluation due to the lack of distinct static objects. 

\subsection{CONTEXTUALIZED OBJECT QUERYING WITH SMALL VLMS}
\label{subsec:nav_context_object_method}

While some 3DSGs are built on a large VLM (LVLM) or large language model (LLM), light-weight 3DSGs, like Terra, rely solely on a small VLM, like CLIP.
Though efficient, this brings with it a lack of higher-level context understandings.
We propose a modified approach that breaks down the query first into object detection and then contextualized relevant rankings for object selection.

Our contextualized object prompting approach first prompts the 3DSG to identify all bounding boxes of the given object name (e.g. ``car") using the MS-Max method proposed in~\cite{samuelsonTerra2025}.
Each bounding box is associated with its spatially closest place node in the 3DSG.
Second, we then sort the bounding box associated place nodes based on semantic relevance to a provided context description (e.g., ``by the garbage can").
Explicitly, this sorting uses a combined score of object bounding box semantic similarity weighted by $0.7$ and place node context semantic similarity weighted by $0.3$.
An ablation on the choice of these weights is given in the Appendix in Table~\ref{tab:context_object_ablation}.
The top place node and the bounding box associated with it represent our contextualized object prediction.
Our approach is referred to as \textit{Object+Context} in the tables below.

We compare this approach to 1) passing in the object name and selecting the most relevant bounding box (referred to as \textit{Object Only}), and 2) passing in the full sentence description (e.g., ``go to the car by the garbage can") and selecting the most relevant bounding box (referred to as \textit{Full Sentence}).

\subsection{PLACE NODE NAVIGATION ANALYSIS}
\label{subsec:place_node_nav_exp}

\subsubsection{Experimental Setup}
\label{subsubsec:nav_exp_setup}

The goal of this experiment is to evaluate how well current 3DSG techniques can be used for navigation to natural-language described locations in a large-scale outdoor scene. 
To isolate the evaluation of the 3DSG representation from the open problem of localization, we use GPS to localize the robot within the 3DSG. 
Since GPS was not collected in the original dataset, we identify corresponding features in the point cloud and satellite imagery to estimate a transformation from the 3DSG coordinate frame to a local GPS frame, which is then applied to all place nodes. 
GPS is used only to associate the robot's position with the 3DSG and not to specify queried object locations. 
We leave robust scene-graph-based localization for future work.

We next define $17$ object prompts across the datasets that refer to objects that are either unique or uniquely identified by contextual descriptions (see Table~\ref{tab:object_prompts} in the Appendix). 
For each dataset, we also select three spatially distributed starting place nodes to evaluate navigation from different regions of the map across varying terrain and obstacles.

For each natural-language query, we use the method described in the previous subsection to identify the top-ranked goal place node, defined by cosine-similarity, and plan an A* path through the place node graph from each starting location. 
We save the GPS coordinates of the place nodes along each planned path as navigation waypoints. 
If the top-1 place node does not correspond to the queried object, we subsequently plan paths from the top-1 to top-2 and from the top-2 to the top-3 ranked place nodes.

For navigation, we manually drive the robotic wheelchair from the starting GPS coordinate to each waypoint sequentially, using a bird's-eye satellite view of the planned path and robot position, along with the distance and bearing to the next waypoint.
The planner advances to the next waypoint within $3$-m and stops within $2$-m of the goal. 
Although navigation is manually executed, this setup isolates the ability of the 3DSG to support natural-language-based navigation; fully autonomous closed-loop navigation is left for future work.

\subsubsection{Metrics}
\label{subsubsec:nav_metrics}

The metrics used for navigation are strict success, relaxed success, and path efficiency.

A strict success (SS) is defined as having the desired object both in view of a camera from the robot and the closest point on the object within $10$-m of the robot's position.
A relaxed success (RS) is defined as either having the object in view of a camera or within $10$-m of the robot.
We use SS-1 or RS-1 representing the strict or relaxed success rate to the top-1 place node.
Similarly, SS-3 and RS-3 represent the strict or relaxed success rates when considering whether any of the top-3 place nodes satisfy the success criteria.

Path efficiency measures how closely the executed robot trajectory matches an approximate reference path to the detected object and is therefore only defined for successful navigation trials (either SS or RS).
It is computed as the ratio between the approximate reference path length and the actual robot trajectory length.
The reference path length is approximated using the Euclidean distance from the starting GPS position to the ground-truth object GPS position when the direct path is unobstructed by water.
If water obstructs the direct path, the reference path is approximated by following the nearest traversable boundary around the water toward the goal location.
This approximation accounts for water obstacles but does not explicitly model other terrain constraints that may make the direct path infeasible.

\subsubsection{Results}
\label{subsubsec:nav_results}

We first isolate the 3DSG's capability to identify the correct objects of interest provided the natural language object and context prompts.
We report only the relaxed success rate by identifying in the point cloud if the place node is near the object of interest and manually verifying whether the object of interest was visible in the images associated with that chosen place node.
We don't report strict success since we don't have a GPS distance measure purely at the graph level.
These results are shown in Table~\ref{tab:obj_det_exp}.

We see that sequentially combining the object and then context descriptions (\textit{Object+Context}) produces the best success rates.
\textit{Object Only} prompts perform next best likely because most of the object prompts used, query unique objects in the datasets (e.g., there is only one ``stop sign" in the Marina Part 1 dataset).
Providing a \textit{Full Sentence} prompt still performs well, but with the lowest success rate. 
This is likely caused when the context contains other object names in it making a small VLM, like CLIP, struggle to disambiguate which object is the focus of the prompt.
We thus use only the \textit{Object+Context} approach for our real-world navigation experiments.

\begin{table}[t]
\centering
\caption{3DSG Object Detection Results}
\setlength{\tabcolsep}{3pt}
\begin{tabular}{p{80pt}p{60pt}p{60pt}}
\hline
Method & RS-1 $\uparrow$ & RS-3 $\uparrow$ \\
\hline
Object Only   & 0.647 & 0.765 \\
Full Sentence & 0.529 & 0.706 \\
Object+Context & \textbf{0.765} & \textbf{0.824} \\
\hline
\multicolumn{3}{p{200pt}}{
Relaxed success rates (RS) for object detection on the 3DSG given $17$ contextualized object prompts.
We see that combining separate object and then context information for unique contextualized object based prompts obtains the highest success rate. 
}
\end{tabular}
\vspace{0.25cm}
\label{tab:obj_det_exp}
\end{table}

For real-world navigation, we plan paths to $14$ out of the $17$ total object prompts since those are all of the objects that the \textit{Object+Context} approach correctly detected (see RS-3 in Table~\ref{tab:obj_det_exp}). 
Given these $14$ object prompts and three starting locations for each dataset, we report a total of $42$ unique navigation trials in Table~\ref{tab:nav_exp}.

\begin{figure*}[t]
\centering
\begin{tabular}{ccc}

\centering
\includegraphics[width=0.335\textwidth]{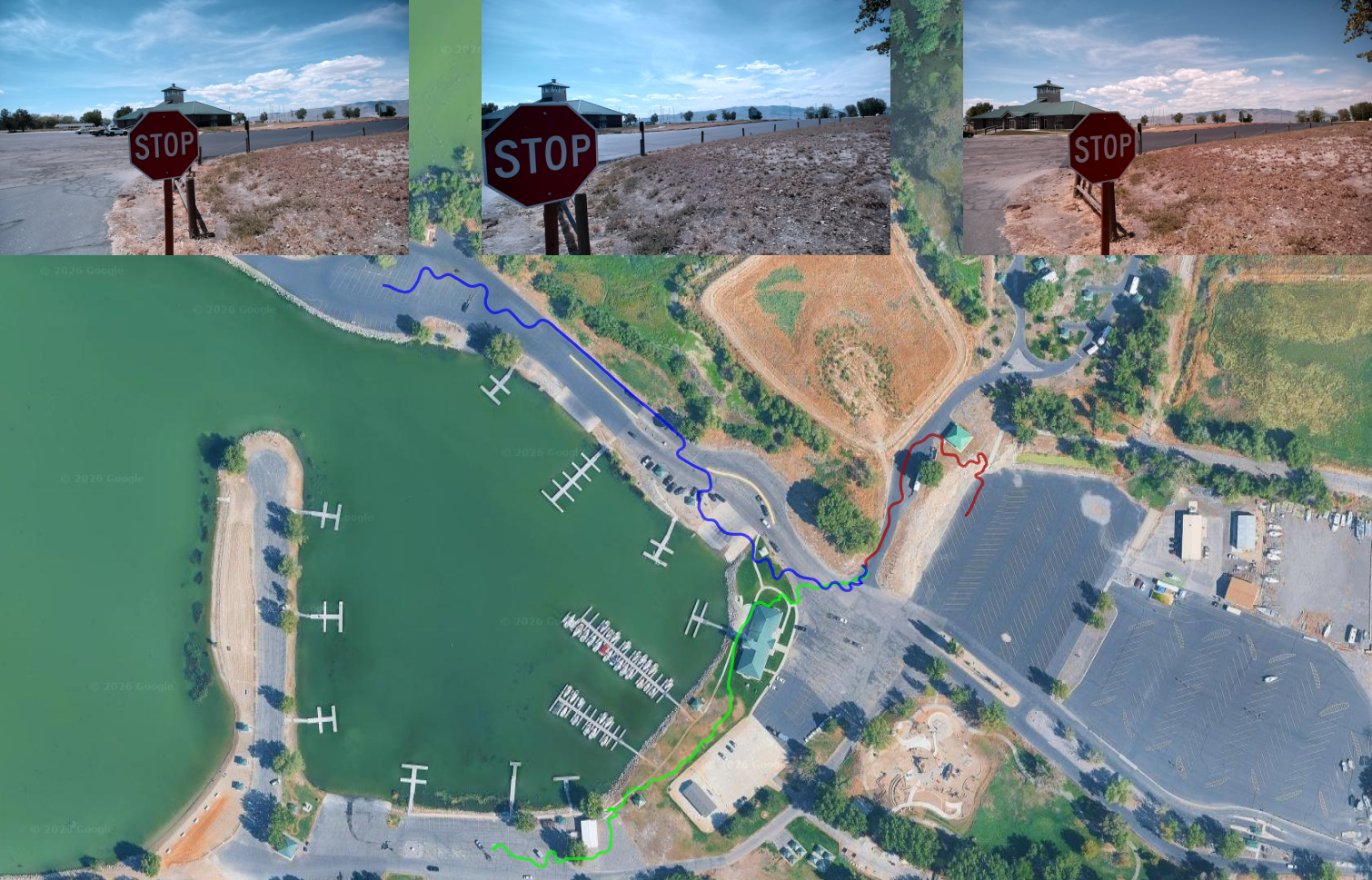} &
\includegraphics[width=0.30\textwidth]{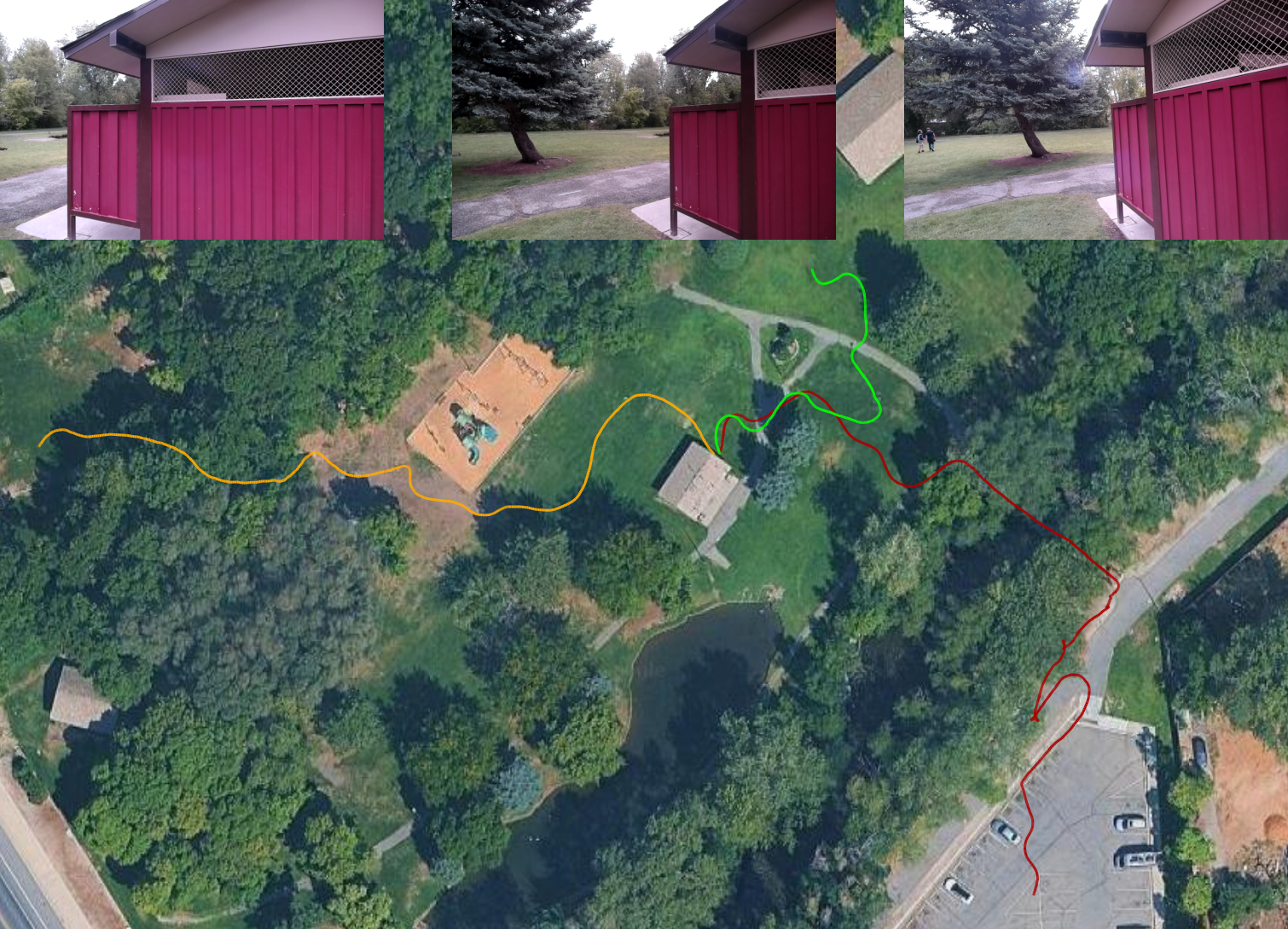} &
\includegraphics[width=0.30\textwidth]{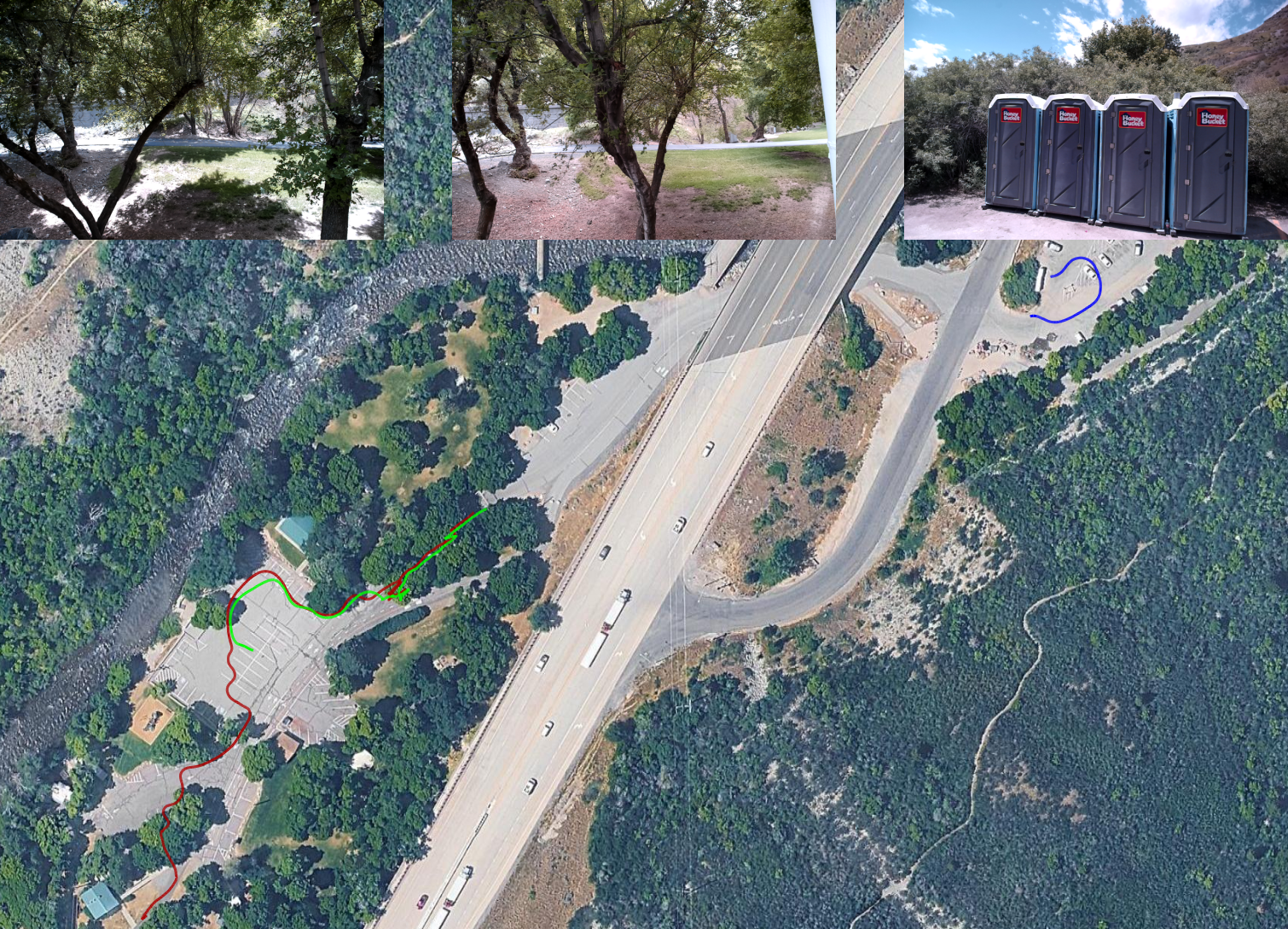} \\

{\fontsize{8}{9.5}\selectfont\rmfamily(a) ``stop sign by road" (Marina Part 1)}\label{fig:stop_sign_nav} &
{\fontsize{8}{9.5}\selectfont\rmfamily(b) ``garbage by the red building" (River Park)}\label{fig:garbage_bldg_nav} &
{\fontsize{8}{9.5}\selectfont\rmfamily(c) ``blue honey buckets" (Nunns Park)}\label{fig:honey_buckets_nav} \\

\end{tabular}
\caption{
Qualitative results on three demonstrative object navigation tasks.
Three camera images in each figure shows the robot's camera view at the trajectory end from the three different starting locations.
The trajectories taken by the robot are colored in red, green, and either blue/orange (switched for visual clarity).
(a) Demonstrates a successful navigation to the stop sign from three diverse starting locations.
(b) Demonstrates a partial success (i.e., RS) where the garbage can is not in the camera view, but it is within 10 meters of the robot's GPS coordinates.
(c) Demonstrates a success for the blue trajectory starting location, but a navigation failure for the other two starting locations.
This navigation failure is caused by planned path GPS waypoints veering off the side of the road down a steep drop-off.
} \label{fig:visual_nav_exp}
\end{figure*}

We report the success rate and path efficiency for each dataset to provide insight into the navigation complexities of certain outdoor environments.
We note that due to either success or navigation failure, we only needed to navigate to the top-1 place node for each trial and thus we only report the RS-1 and SS-1 results.
The reported path efficiency is the mean and standard deviation path efficiency across all trials in a given dataset with one standard deviation shown in the parentheses.

\begin{table}[t]
\centering
\caption{Real-World 3DSG-Based Object Navigation Results}
\setlength{\tabcolsep}{3pt}
\begin{tabular}{p{50pt}p{40pt}p{35pt}p{35pt}p{50pt}}
\hline
Dataset & \# of Trials & SS-1 $\uparrow$ & RS-1 $\uparrow$ & Path Efficiency \\
\hline
River Park   & 15 & 0.400 & 0.800 & 0.68 ($\pm$ 0.09) \\
Nunns Park   & 12 & 0.417 & 0.417 & 0.64 ($\pm$ 0.26) \\
Marina Part 1 & 6 & 1.000 & 1.000 & 0.70 ($\pm$ 0.16) \\
Marina Part 2 & 9 & 0.667 & 0.667 & 0.60 ($\pm$ 0.17) \\
\hline
\textbf{Summary} & 42 & 0.548 & 0.691 & 0.66 ($\pm$ 0.17) \\
\hline
\multicolumn{5}{p{240pt}}{
Relaxed success rates (RS) for object detection on the 3DSG.
We see that combining separate object and then context information for unique contextualized object based prompts obtains the highest success rate. 
}
\end{tabular}
\vspace{0.25cm}
\label{tab:nav_exp}
\end{table}

Obtaining a majority of successful navigation rates, demonstrate the usability and potential of general outdoor 3DSGs for downstream robotic navigation tasks (see Table~\ref{tab:nav_exp}).

An unideal navigation success rate between $54.8-69.1\%$ can largely be attributed to two factors.
The first factor has to do with the choice of navigating to the closest-place node of the detected object.
Doing so does not guarantee visual coverage of the object or that there is a place node within $10$ meters of the object both of which define a strict successful navigation. 
This factor explains the discrepancy between SS-1 and RS-1 for the River Park dataset.
A visual demonstrative example is shown in Figure~\ref{fig:visual_nav_exp}(b) where we navigate to the correct garbage bin object by the red building, but the object is just around the corner and out of view.
Importantly, the endpoint criteria (in-view and within $10$-m) did not result in any navigation failures; rather, they only account for the difference between relaxed and strict success rates.

The second factor is navigation failure caused by untraversable terrain.
A navigation failure occurs when the planned path directs the robot to go to a location that is not-traversable for our ground robot to navigate to.
Across these datasets, navigation failure was caused by GPS waypoints either directing the robot to go off a ledge or to cross a deeper water stream.
Navigation failures were the sole cause of RS-1 scores below $1.0$ across all datasets: 3 out of the 15 trials in River Park, 7 out of the 12 trials in Nunns Park, and 3 out of the 9 trials in the Marina Part 2 dataset.
A visual example is shown in Figure~\ref{fig:visual_nav_exp}(c) where due to the first two starting locations, the robot's path required it at one point to drive off a multiple meter deep ledge (partially visible in the first two image frames).

Roughly a $66\%$ path efficiency shows that although navigable, the place node graph structure is less than ideal in determining the optimal route to the queried object.
This efficiency could be improved by applying a smoothing method to the planned path, such as B-Splines~\cite{bsplines_deBoor1978}, or improving the graph connectivity.
A better place node graph connectivity could involve fully connecting place nodes within a given radius, though this could potentially lead to more navigation failures.

\subsection{GENERAL TAKEAWAYS}
\label{subsec:nav_takeaways}

3DSGs do enable robotic navigation to contextualized object queries.
The obvious initial limitation for navigation is that the 3DSG can only ever perform as well as the VLM that it's built upon.
While we present an approach to better handle context-based queries for light VLMs that struggle interpreting complex natural language queries, they still won't have high-level understandings of the query at hand.
While the extent of this limitation depends on the capabilities of the underlying VLM, it is particularly relevant to 3DSG methods, such as Terra, that utilize small and lightweight VLMs to generate semantic embeddings.

The place node layer of outdoor 3DSGs does enable successful navigation to detected objects; however, traversability needs to be embedded into the graph structure to guarantee successful navigation. 
This limitation arises from Terra's use of a terrain-based place node graph structure that does not explicitly encode traversability for navigation and may therefore not apply to outdoor 3DSGs that incorporate traversability directly into their graph structures.

Additionally, while GVD-based graphs normally guarantee a safe structure for path planning, they can result in sub-optimal navigation paths.
Better graph algorithms should be utilized in outdoor 3DSG development to enable more efficient planned paths for complex outdoor open scenes.

\section{ANALYSIS OF VLM-BASED REGION UNDERSTANDING IN OUTDOOR ENVIRONMENTS}
\label{sec:region_analysis}

\subsection{REGION QUERYING}
\label{subsec:region_querying}

We next analyze the ability of VLM-based 3DSGs to identify relevant regions given natural language queries.
Having an understanding of semantically and geometrically meaningful regions is important for outdoor robots, as it allows them to perform tasks such as monitoring specific areas of an environment or provide context for deeper task reasoning.
We use the methods described in Terra~\cite{samuelsonTerra2025} to retrieve relevant place nodes pertaining to each description based region.
All the region prompts that were used are listed out in Table~\ref{tab:region_prompts} in the Appendix.

\begin{table*}[t]
\centering
\caption{Region Querying Results}
\setlength{\tabcolsep}{6pt}
\begin{tabular*}{\textwidth}{p{72pt}p{58pt}p{58pt}p{58pt}p{58pt}p{58pt}p{58pt}}
\hline
Method 
& \multicolumn{3}{c}{Micro $\uparrow$}
& \multicolumn{3}{c}{Macro $\uparrow$} \\
\cline{2-4} \cline{5-7}
& Avg Prec. 
& Avg Rec. 
& Avg F1 
& Avg Prec. 
& Avg Rec. 
& Avg F1 \\
\hline

\multicolumn{7}{l}{\textbf{River Park} - 4 regions} \\ \hline
Spectral
& 0.242 $\pm$ 0.059 & 0.843 $\pm$ 0.132 & 0.364 $\pm$ 0.032 & 0.261 $\pm$ 0.040 & 0.722 $\pm$ 0.155 & 0.359 $\pm$ 0.025 \\
Agglomerative
& \textbf{0.284} $\pm$ 0.136 & \textbf{0.848} $\pm$ 0.182 & \textbf{0.384} $\pm$ 0.095 & \textbf{0.278} $\pm$ 0.059 & \textbf{0.751} $\pm$ 0.167 & \textbf{0.371} $\pm$ 0.055 \\

\hline
\multicolumn{7}{p{490pt}}{\textbf{Nunns Park} - 4 regions} \\ \hline
Spectral
& 0.210 $\pm$ 0.042 & \textbf{0.657} $\pm$ 0.290 & \textbf{0.305} $\pm$ 0.086 & \textbf{0.209} $\pm$ 0.073 & \textbf{0.634} $\pm$ 0.164 & 0.284 $\pm$ 0.047 \\
Agglomerative
& \textbf{0.211} $\pm$ 0.043 & 0.542 $\pm$ 0.326 & 0.281 $\pm$ 0.101 & 0.206 $\pm$ 0.025 & 0.581 $\pm$ 0.201 & \textbf{0.288} $\pm$ 0.0506 \\

\hline
\multicolumn{7}{p{490pt}}{\textbf{Marina Part 1} - 4 regions} \\ \hline
Spectral
& 0.548 $\pm$ 0.119 & \textbf{0.383} $\pm$ 0.152 & \textbf{0.415} $\pm$ 0.098 & 0.616 $\pm$ 0.071 & \textbf{0.378} $\pm$ 0.137 & \textbf{0.365} $\pm$ 0.070 \\
Agglomerative 
& \textbf{0.623} $\pm$ 0.127 & 0.353 $\pm$ 0.189 & 0.396 $\pm$ 0.150 & \textbf{0.629} $\pm$ 0.103 & 0.337 $\pm$ 0.156 & 0.344 $\pm$ 0.106 \\

\hline
\multicolumn{7}{p{490pt}}{\textbf{Marina Part 2} - 1 regions} \\ \hline
Spectral
& 0.903 $\pm$ 0.137 & \textbf{0.642} $\pm$ 0.243 & \textbf{0.702} $\pm$ 0.165 & 0.903 $\pm$ 0.137 & \textbf{0.642} $\pm$ 0.243 & \textbf{0.702} $\pm$ 0.165 \\
Agglomerative
& \textbf{0.999} $\pm$ 0.003 & 0.469 $\pm$ 0.248 & 0.598 $\pm$ 0.237 & \textbf{0.999} $\pm$ 0.003 & 0.469 $\pm$ 0.248 & 0.598 $\pm$ 0.237 \\

\hline
\multicolumn{7}{p{490pt}}{\textbf{Rock Canyon Campground} - 2 regions} \\ \hline
Spectral
& 0.211 $\pm$ 0.062 & \textbf{0.313} $\pm$ 0.060 & \textbf{0.242} $\pm$ 0.049 & 0.117 $\pm$ 0.035 & \textbf{0.202} $\pm$ 0.039 & \textbf{0.142} $\pm$ 0.029 \\
Agglomerative
& \textbf{0.227} $\pm$ 0.076 & 0.299 $\pm$ 0.077 & 0.240 $\pm$ 0.054 & \textbf{0.126} $\pm$ 0.044 & 0.193 $\pm$ 0.0496 & 0.140 $\pm$ 0.032 \\

\hline
\multicolumn{7}{p{490pt}}{The averaged precision, recall, and F1 scores over all $k$s for each dataset. Results of two region clustering methods are analyzed: agglomerative clustering and spectral clustering.} \\
\end{tabular*}
\label{tab:region_querying}
\end{table*}

\begin{figure}
    \centering
    \includegraphics[width=0.95\columnwidth]{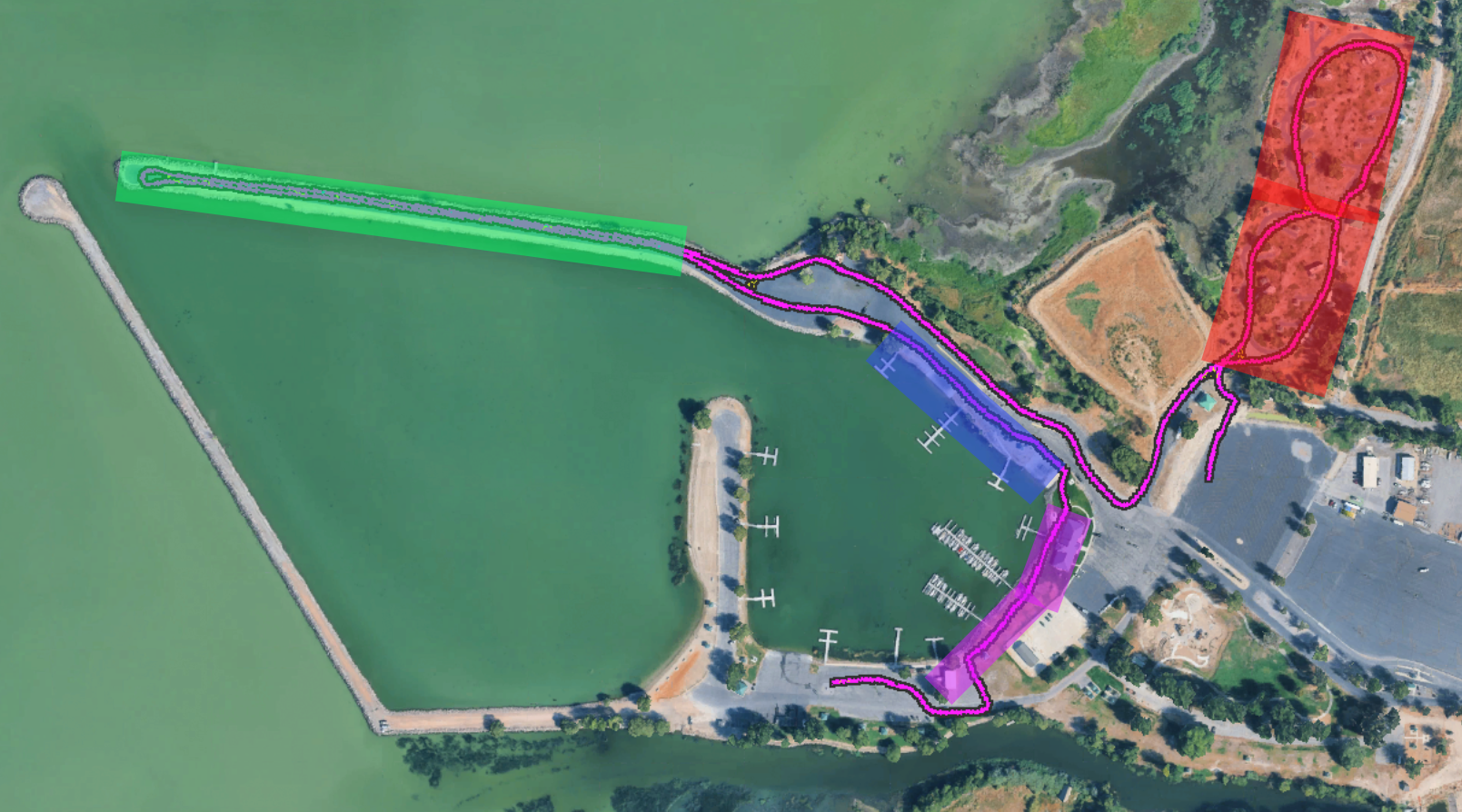}
    \caption{Labeled regions for the Marina Part 1 dataset as used in the region querying experiments. The given descriptions of the regions by color are as follows: Red: ``Pavilions and picnic tables with parking stalls"; Green: ``Long road with water on both sides"; Blue: ``Boat ramps and docks with parking lot"; and Purple: ``Buildings and sidewalk path." }
    \label{fig:marina_p1_labeled_regions}
\end{figure}

\subsubsection{Experimental Setup} 
The region querying experiments were performed by showing a student who was familiar with the areas in the datasets the pictures that were taken with the sensor platform. 
After seeing the pictures, they labeled and described unique regions on all of the satellite pictures shown in Figure~\ref{fig:dataset_visual_overview}.
An example of the Marina Part 1 dataset with regions labeled is shown in Figure~\ref{fig:marina_p1_labeled_regions}.
These descriptions were used for region querying prompts given to Terra to predict which place nodes were associated with that prompt. 
Two different methods of clustering region nodes were evaluated: agglomerative clustering and spectral clustering. 
The parameter $k$ for Terra is the number of region nodes the region predictor will select per query (in order of highest cosine similarity to the region prompt). 
Terra returns the relevant place nodes from the children of these region nodes. 
For each dataset, region querying was performed for a range of $k$ values from 1 to 15.

\subsubsection{Metrics} 
Micro metrics were calculated per method and per dataset by separately summing the true positives, false positives, and false negatives for each region description and calculating the recall, precision, and F1 scores conventionally.
These metrics are partially biased because of datasets having varying region sizes. 
Larger regions have more nodes than regions that are smaller, and thus impact the metrics more.
Because of this, the macro- precision, recall, and F1 are also reported, and are calculated by averaging the precision, recall, and F1 scores respectively for each region in a dataset.
It is noted that the macro F1 is also biased giving small regions equal weight to large identified regions.

\begin{figure*}
    \centering
    \begin{minipage}{0.49\textwidth}
        \centering
        \includegraphics[width=\linewidth]{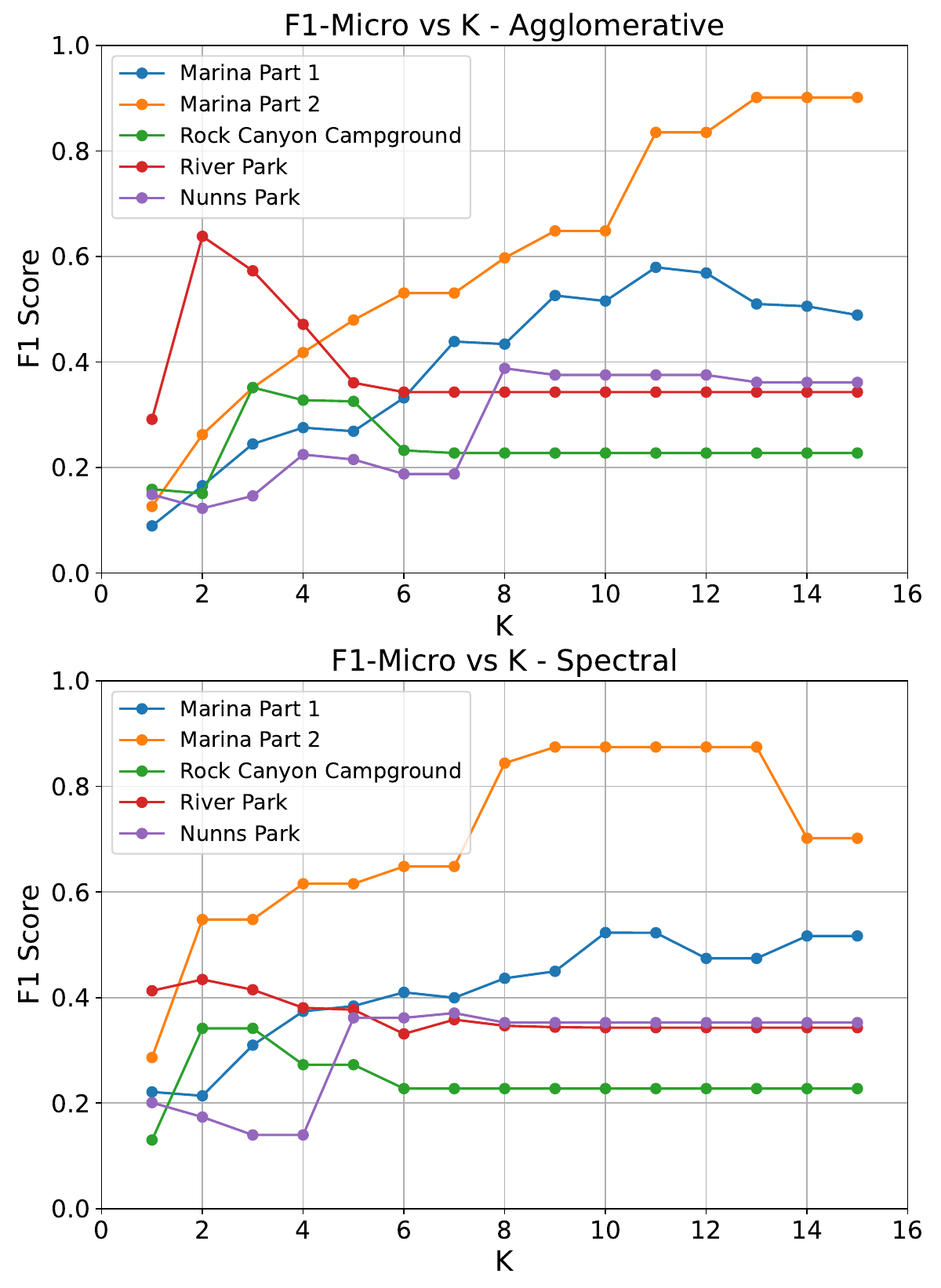}
    \end{minipage}
    \hfill
    \begin{minipage}{0.49\textwidth}
        \centering
        \includegraphics[width=\linewidth]{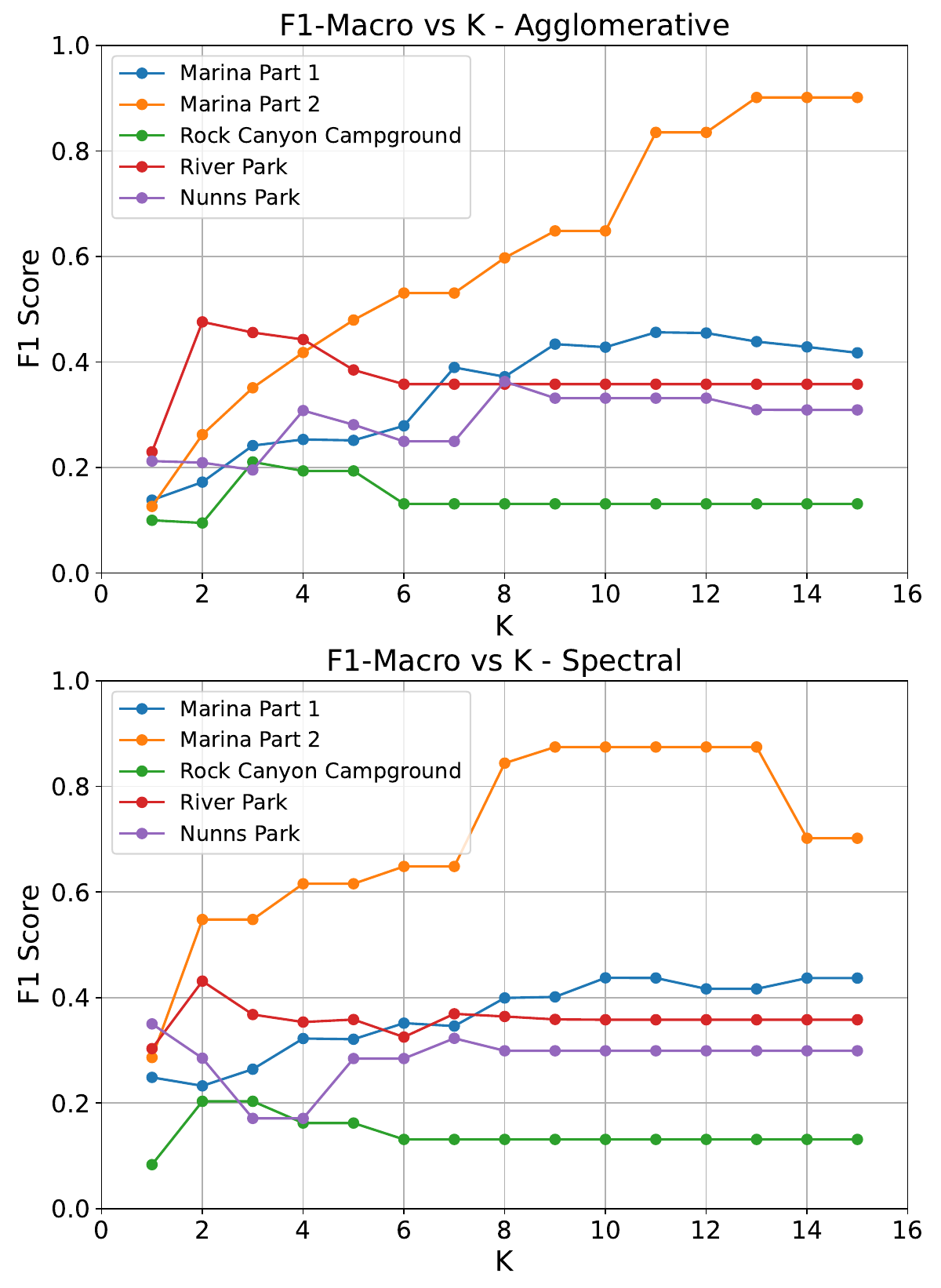}
    \end{minipage}
    
    \caption{Region querying results. Micro-F1 scores (left) and Macro-F1 scores (right) reported across a range of $k$ values for each dataset.}
    \label{fig:reg_querying_k_plots}
\end{figure*}

\begin{figure}
    \centering
    \includegraphics[width=0.95\columnwidth]{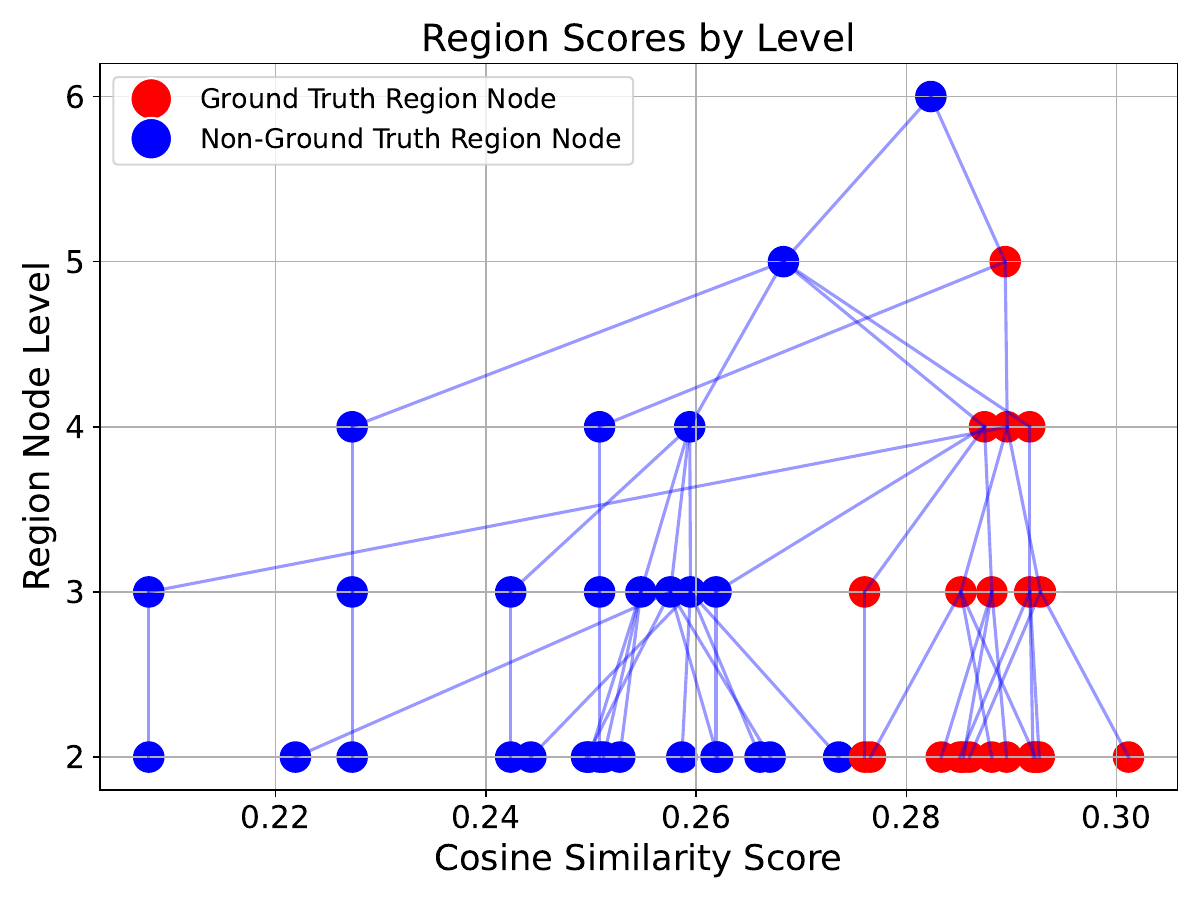}
    \caption{Region nodes and their cosine similarity scores by level for query ``Long dirt road with water and dirt rocks" from the Marina Part 2 dataset.}
    \label{fig:reg_scores_by_level}
\end{figure}

\subsubsection{Results}
The micro and macro F1 scores for each clustering method are reported for each value of $k$ in Figure~\ref{fig:reg_querying_k_plots}.
We compute the mean and standard deviation across the values of $k$ for precision, recall, and F1 scores for each method.
These results are reported in Table~\ref{tab:region_querying}. 

Both the agglomerative and spectral clustering method perform on par with each other, with no statistically significant difference between the two.
It is noted that the micro and macro scores for the Marina Part 2 are the same as there is only one region labeled for that dataset. 

Of note are the lower scores of the Rock Canyon Campground dataset, compared to the Marina datasets.
We suspect these differences can partially be attributed to the degree of homogeneity and artificiality in each of the datasets.
The Marina datasets are more urban-like with more man-made constructions such as buildings, boat ramps, and pavement.
These features allow more distinct and clear descriptions and boundaries of regions.
The Rock Canyon dataset, on the other hand, has scenery that is a lot more organic with more unstructured and ambiguous regions, like foliage around trails, and clearings.
Nunns Park and River Park lie in the middle with both a mix of man-made structures and natural features.
This points out the need for precise prompts as well as general ambiguity in describing unstructured outdoor environments.
In fact, it is noted that the metrics reported may have been affected by annotation and region subjectivity across all the datasets.
This is a major challenge with region querying.
The issues of prompt tuning and region ambiguity are addressed further in Section~\ref{sec:region_analysis}.\ref{subsec:region_prompt_tuning_exp}.

As seen in Figure~\ref{fig:reg_querying_k_plots}, the choice of the parameter $k$ can heavily affect the results.  
The best $k$ for each dataset is drastically different.
For example, the River Park dataset (the smallest dataset by trajectory length) performed best, in terms of F1 score, with a $k$ of 2, while the Marina Part 1 dataset (the largest dataset) performed the best with a $k$ between 10 and 11.
Ideally, if region clustering is performed well, a lower $k$ would be preferred, indicating that regions are segregated well, both semantically and geometrically.
Further research should investigate better ways to cluster regions to better balance semantic and geometric clustering.

Upon further investigation, another reason why the optimal $k$ prefers to be high in many cases is because lower region nodes (in the hierarchical region graph) are often selected before higher region nodes.
For example, Figure~\ref{fig:reg_scores_by_level} shows the Marina Part 2's region nodes displayed by their cosine similarity to the prompt and their level.
For our purposes, region nodes are determined to be``ground truth" if the majority of their place node descendants are part of the ground truth set (these nodes are colored red in Figure~\ref{fig:reg_scores_by_level}).
It can be seen that some red ground truth level 2 and 3 region nodes will be selected before the red ground truth level 4 or 5 region nodes due to their higher similarity scores, even though these upper-level region node may better summarize the region geometrically.

Ideally, 3DSG region nodes provide higher-level general semantic understanding of their children place nodes.
This would lend itself to ground truth region nodes having a higher similarity to the region description than that of its children, as it should emphasize the shared features of its children nodes, and get rid of erroneous image-specific noise.
However, the opposite is found, with only 2 region nodes, out of the 68 total ground truth region nodes across all queries and datasets having a higher similarity than all of its children.
The majority of the time, the similarity of the parent nodes lie between the similarity of its children, as can be seen in Figure~\ref{fig:reg_scores_by_level}.

\subsection{PROMPT TUNING}
\label{subsec:region_prompt_tuning_exp}

\begin{table}[t]
\caption{Region Querying with Prompt Tuning Results}
\setlength{\tabcolsep}{3pt}
\begin{tabular}{p{105pt}p{45pt}p{45pt}p{20pt}}
\hline
Region Prompt & Precision $\uparrow$ & Recall $\uparrow$ & F1 $\uparrow$ \\
\hline
\multicolumn{4}{l}{\textbf{Marina Part 1} - Only 1 of the 4 Regions} \\ \hline
Original      & 0.713 & 0.521 & 0.602 \\
Clarification & 0.840 & \textbf{0.610} & \textbf{0.707} \\
Negation      & 0.152 & 0.152 & 0.152 \\
Scenery      & 0.688 & 0.345 & 0.460 \\
Combination   & \textbf{0.970} & 0.313 & 0.474 \\
\hline
\multicolumn{4}{p{240pt}}{
Region metrics for five different prompts of the same region from the Marina Part 1 dataset. The prompts are as follows: Original: ``Pavilions and picnic tables with parking stalls''; Clarification: ``Green pavilions and picnic tables with parking stalls''; Negation: ``Empty RV park with no RVs''; Scenery: ``Pavement road with open grassy areas and sparse trees''; Combination: ``Green pavilions with paved roads and grass.''}
\end{tabular}
\vspace{0.25cm}
\label{tab:region_prompt_tuning}
\end{table}

Region querying inherits the sensitivity of different prompts from CLIP. 
Multiple distinct descriptions can accurately characterize the same region, but produce substantially different results when used as CLIP queries. 
Exacerbating this is the inherently subjective nature of regions in outdoor environments. 
The boundaries of a region are often ambiguous, and selecting the best description of a region is even more so. 
To illustrate this, we analyze the effects of varying the description of a particular region.

In the Marina Part 1 dataset, one region was initially described as ``Pavilions and picnic tables with parking stalls" (see red labeled region on Figure~\ref{fig:marina_p1_labeled_regions}). 
Four other descriptions are evaluated with the same ground truth place nodes. 
For all the prompts, agglomerative clustering and a $k$ of 11 were used.
The associated metrics are shown in Table~\ref{tab:region_prompt_tuning}.

All five prompts resulted in different metrics, ranging from $0.152$ to $0.707$ F1 scores. 
The clarification prompt, ``Green pavilions and picnic tables with parking stalls," adds the adjective ``green" to the initial prompt, and yields improved precision, recall, and F1 scores. 
Of particular note are the results of the negation prompt ``Empty RV park with no RVs."
Although this is a valid description of the region, this prompt achieved the lowest F1 score ($F1=0.152$).
This behavior aligns with known limitations of CLIP which tend to struggle with negation \cite{brody2023cliplogical, ma2022crepe}.
The result is that negated phrases tend to align more closely with images or text associated with ``RVs" rather than with the negated phrase ``no RVs."  
While prompt sensitivity is a known limitation of foundation models, especially smaller models such as CLIP, the ambiguity and subjective interpretation of natural outdoor regions further amplify this challenge.  

\subsection{GENERAL TAKEAWAYS}
\label{subsec:region_takeaways}

Distinguishing and querying regions is a major challenge for outdoor scene graphs.
Regions in unstructured outdoor environments can be ambiguous in both description and boundaries.
As discovered in Section~\ref{sec:region_analysis}.\ref{subsec:region_querying}, the more homogeneous the scenery, the poorer a VLM-based 3DSG performs at clustering regions.
Section~\ref{sec:region_analysis}.\ref{subsec:region_prompt_tuning_exp} further explored that tuning the region prompt description can have drastic effects on querying results.
This is not a 3DSGs-only problem, however, as even throughout the labeling process, the student that labeled regions, had to relabel, or clarify descriptions upon realizing that descriptions were either not distinct, or were too ambiguous.
Future work could consider incorporating multi-annotator region labeling to assist the evaluation of this problem.
The inherent ambiguity and homogeneity of regions in complex outdoor scenes are likely to present a significant challenge for region detection and querying in outdoor 3DSGs.

Compounding the challenge of unstructured regions is the discovery that averaging embeddings often results in a decreased similarity to region description prompts than some of its constituents.
This may indicate that averaging embeddings to summarize regions may not be the best way to agglomerate information from a VLM for broader region understanding.
Averaging may be diluting information that could be critical in distinguishing regions.
Although this observation arises from the embedding aggregation approach used in Terra, averaging embeddings is also a common practice in VLM-based applications. 
Further research is therefore needed to investigate superior region querying algorithms and ways of retaining important information and building more general semantic understanding.

\section{ANALYSIS OF FULL 3DSG UTILITY FOR OUTDOOR FIELD ROBOTICS}
\label{sec:full_3dsg_analysis}

The previous sections evaluated individual components commonly used in modern 3DSG systems, including VLM semantic embeddings, place-node navigation structures, and region-level semantic reasoning.
While these experiments provide insight into the behavior and limitations of specific layers within a 3DSG's hierarchy, the practical value of a 3DSG for outdoor field robotics ultimately depends on the performance and stability of the complete integrated representation.
In this section, we therefore evaluate the overall utility of the Terra 3DSG under realistic deployment conditions.
We focus on two properties that are particularly important for long-term outdoor autonomy: the scalability and memory efficiency of the representation, and the structural, semantic, and geometric consistency of the graph across repeated traversals of the same environment.

\subsection{MEMORY SIZE}
\label{subsec:memory_size_analysis}

\begin{figure}
    \centering
    \includegraphics[width=0.95\columnwidth]{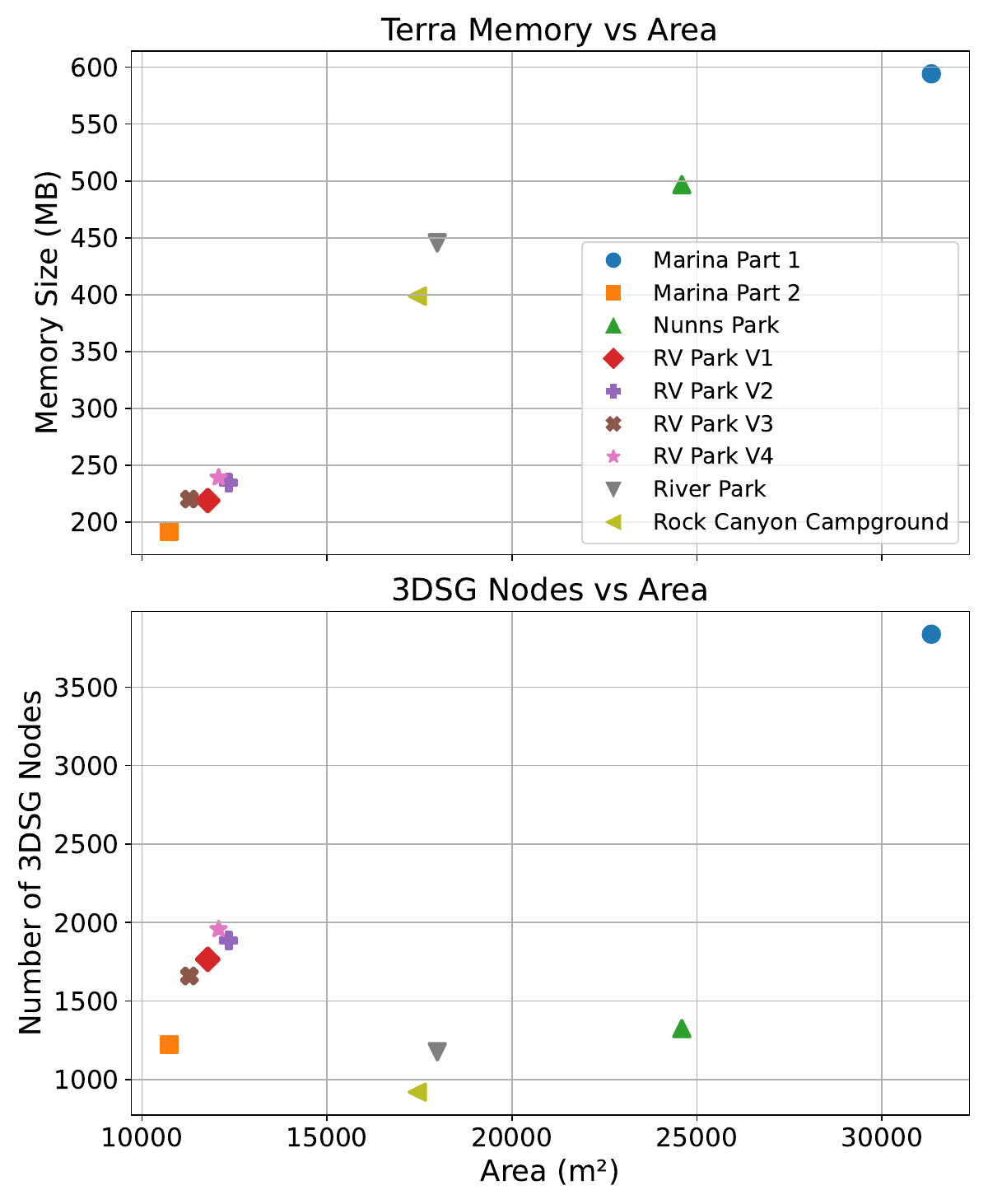}
    \caption{Memory experiment comparing mapped area size to Terra memory size and number of nodes.
     The largest 3DSG remains under 600 MB supporting the claim that 3DSGs can be memory efficient in large-scale outdoor scenes.
    }
    \label{fig:mem_exp}
\end{figure}

This experiment looks at visualizing trends in the Terra 3DSG graph and memory size across the main five datasets evaluated and discussed previously as well as the four consistency datasets of the RV Park.
We included the consistency datasets to help see trends and to evaluate the change in Terra 3DSG graph and memory sizes when the geometrical area and scene changes very little. 

\subsubsection{Metrics} 
Terra 3DSG class objects are stored as \texttt{Terra.pkl} files.
The primary attributes that the Terra 3DSG class object contains are the full metric map (i.e., the sparsified LIO-SAM global point cloud), averaged semantic embeddings for every global point, unique CLIP embeddings from the MS-Map (before averaging), and the entire place and region node 3DSG.
Memory size is defined as the size of the Terra class loaded in \texttt{python} from the saved \texttt{.pkl} file, reported in megabytes (MB).
The number of 3DSG nodes is defined as the total number of all place and region nodes.
To compute the area of a dataset, we first extract the 2D xy-coordinates of all points in the sparse point cloud.
Given these xy-coordinates and the LIO-SAM resolution (e.g., $0.4~m$), we can build an occupancy grid and count the number of occupied cells.
Area is then calculated as the number of occupied cells times the square resolution.
We found that this area definition was more informative than trajectory length (due to the fact that many trajectories revisit the same area multiple times) or a simple square area calculation (due to the fact that many trajectories are far from rectangular in shape (e.g., Marina Part 1)). 
Note that we only present the memory size of the agglomerative clustered regions as we found insignificant differences in memory and number of nodes between agglomerative and spectral region methods.

\subsubsection{Results}
The results of our memory experiment are shown in Figure~\ref{fig:mem_exp}. 
We see a relatively clear linear trend in the area covered by the robot and the memory size of the Terra 3DSG.
It is impressive that the memory size of the Terra 3DSG for a large $30000~m^2$ area ($3~km$ trajectory length) is only 600 MB which is less than other state of the art methods compared to in the Terra paper~\cite{samuelsonTerra2025} evaluated on smaller datasets.
We also note that the places and regions 3DSG portion of Terra is under 10 MB across all datasets. 

\begin{figure*}[t]
\centering

\begin{tabular}{cccc}
\includegraphics[width=0.225\textwidth]{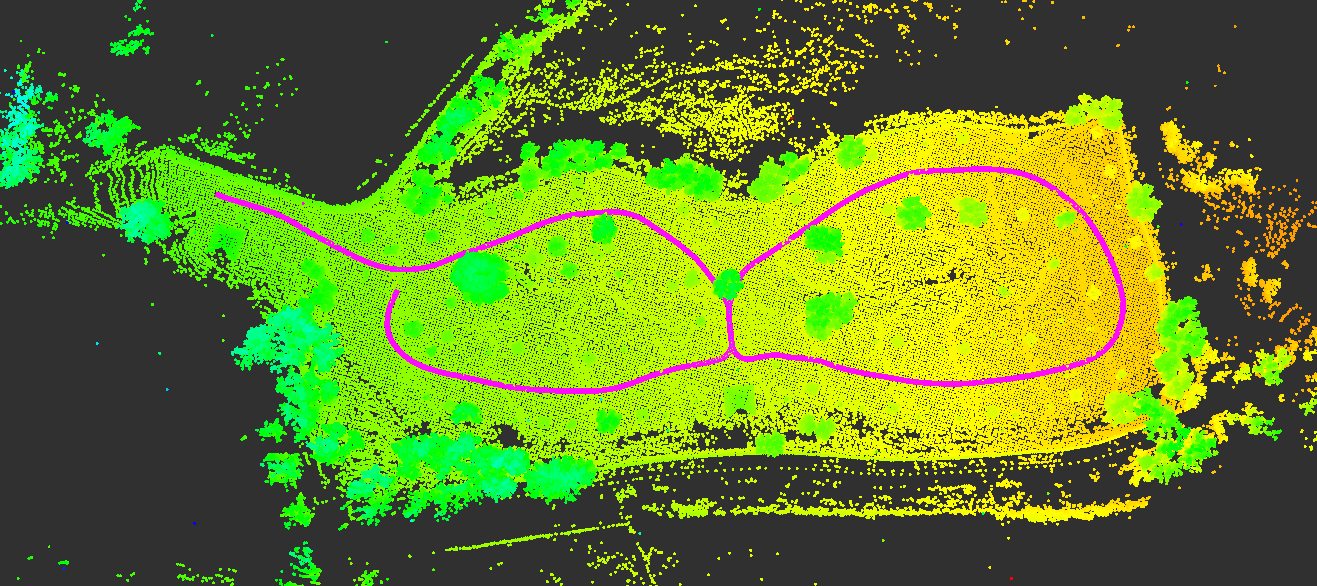} &
\includegraphics[width=0.225\textwidth]{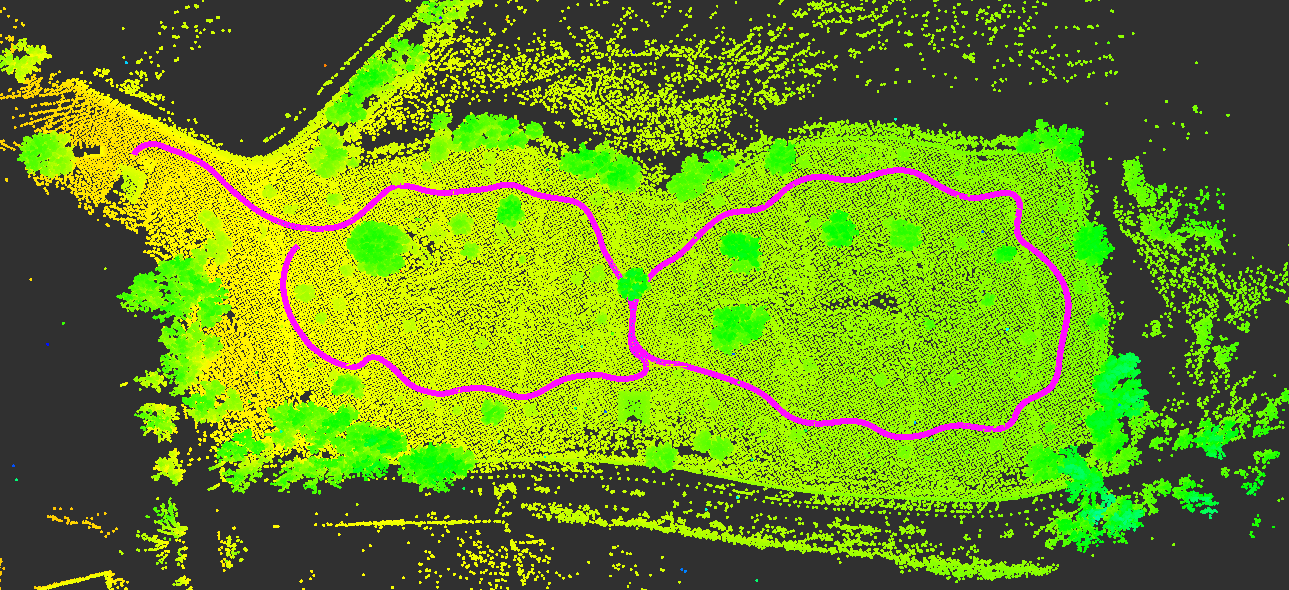} &
\includegraphics[width=0.225\textwidth]{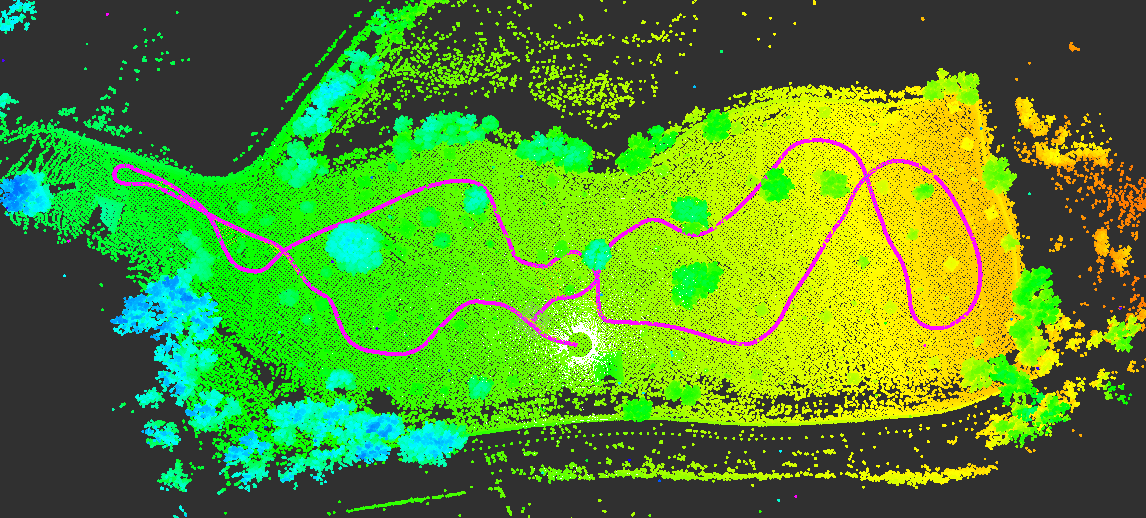} &
\includegraphics[width=0.225\textwidth]{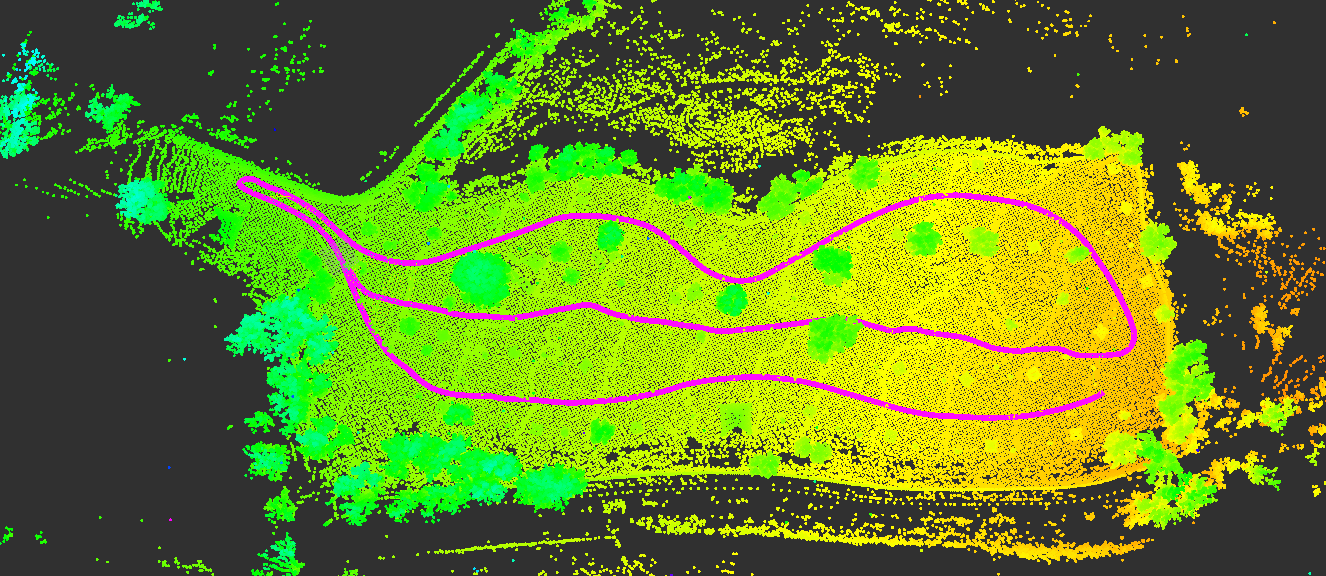} \\

\rotatebox[origin=c]{180}{\includegraphics[width=0.225\textwidth]{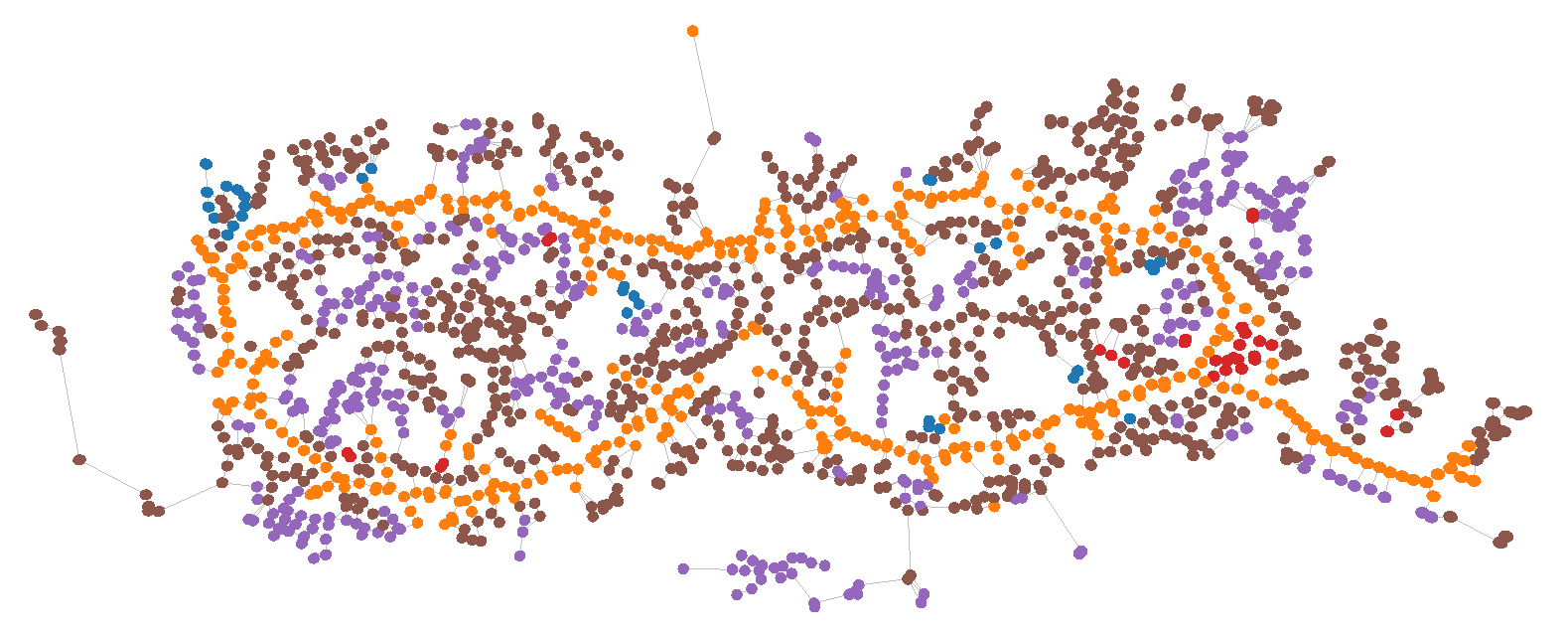}} &
\rotatebox[origin=c]{180}{\includegraphics[width=0.225\textwidth]{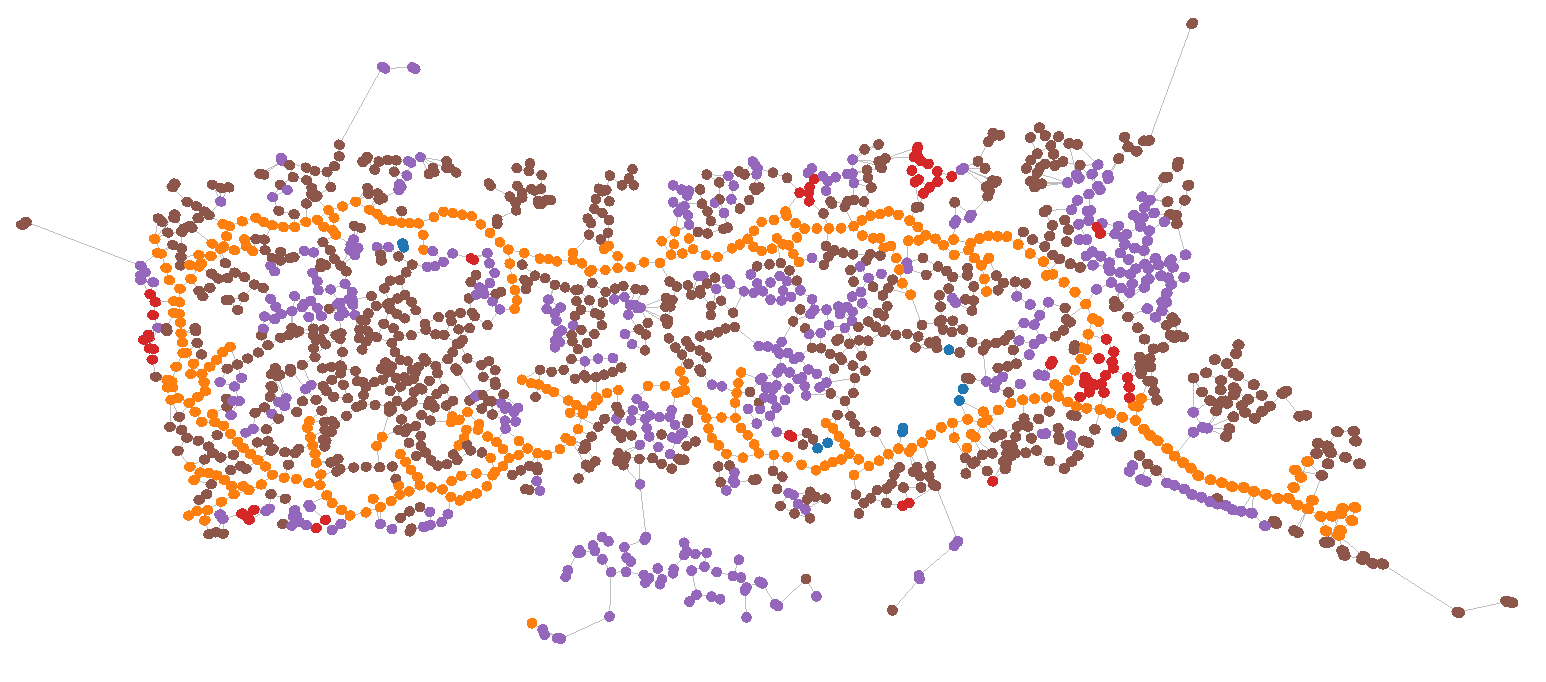}} &
\rotatebox[origin=c]{180}{\includegraphics[width=0.225\textwidth]{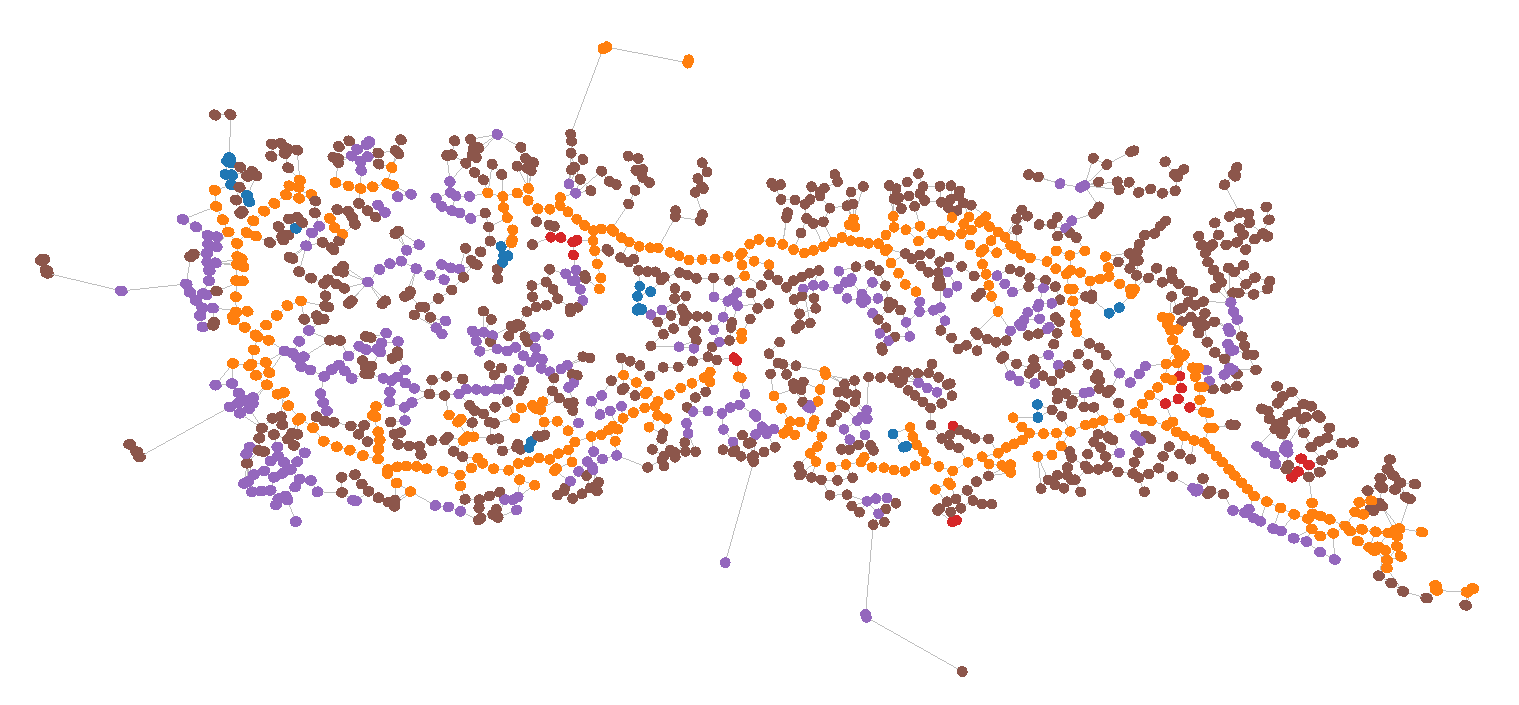}} &
\rotatebox[origin=c]{180}{\includegraphics[width=0.225\textwidth]{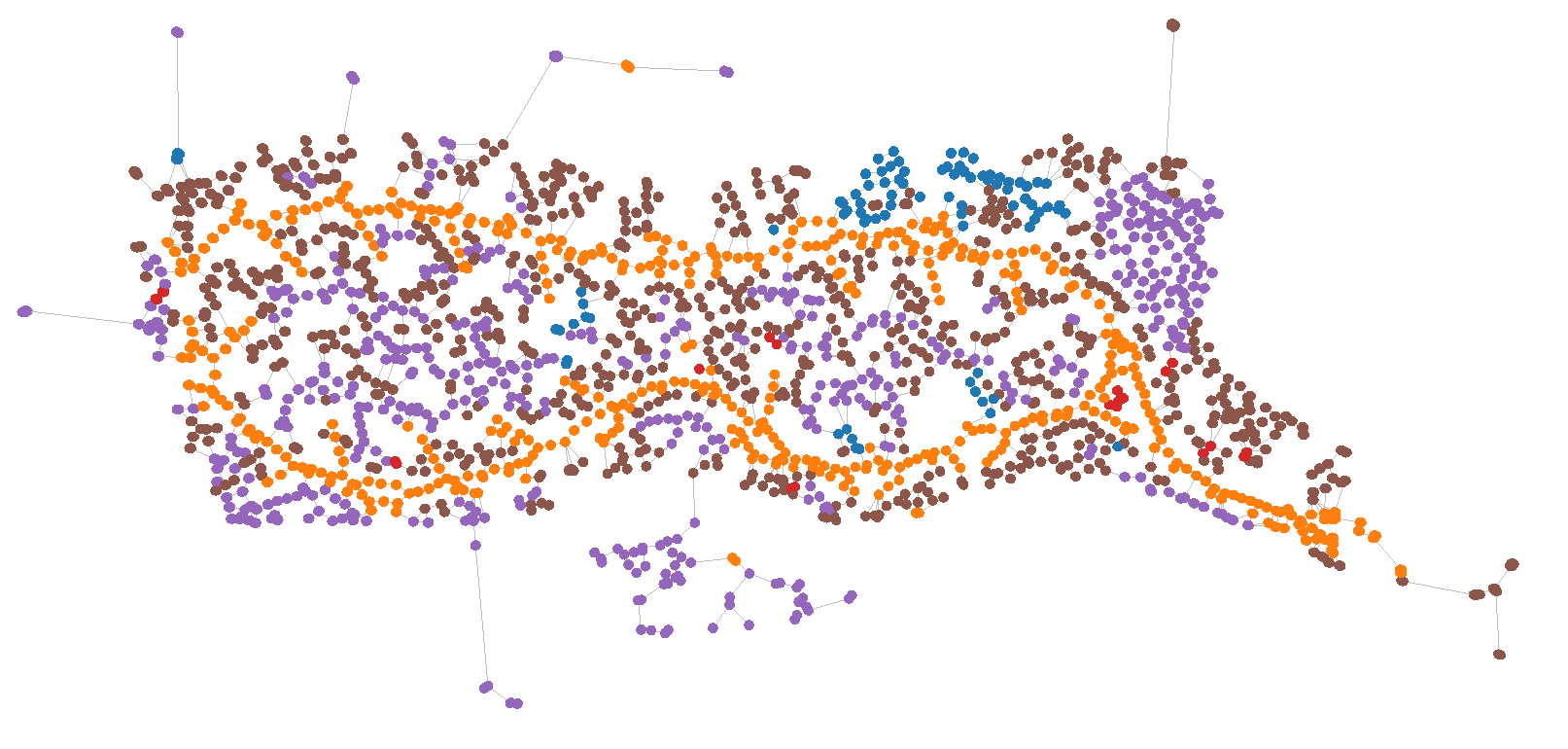}} \\

{\fontsize{8}{9.5}\selectfont\rmfamily(a) RV Park V1} &
{\fontsize{8}{9.5}\selectfont\rmfamily(b) RV Park V2} &
{\fontsize{8}{9.5}\selectfont\rmfamily(c) RV Park V3} &
{\fontsize{8}{9.5}\selectfont\rmfamily(d) RV Park V4}
\end{tabular}

\caption{
Four datasets collected across the same geographical area.
Top row: LIO-SAM metric map.
Bottom row: Place node graphs colored by terrain.
}
\label{fig:terra_consistency_maps}
\end{figure*}

There is an unclear relationship between the number of Terra 3DSG nodes and the area covered. 
This ambiguity is likely caused by the structure of the scene itself affecting the density of place nodes.
For example, let us compare an RV Park dataset with the Marina Part 2 dataset which are similar in area.
The RV Park is mostly open whereas Marina Part 2 consists of long, narrow breakwater segments predominantly surrounded by water (where point cloud points don't exist) creating mostly a skeleton topographical place node graph of the breakwater segments (compare the density of RV Park place nodes in Figure~\ref{fig:terra_consistency_maps} with the visual skeleton structure of Marina Part 2's satellite image in Figure~\ref{fig:dataset_visual_overview}(d)). 

These experiments demonstrate an outdoor 3DSG's capability to maintain a small memory footprint across large-scale varied outdoor environments allowing for the potential of building 3DSGs on-board a mobile robot in real-time.

\subsection{CONSISTENCY}
\label{subsec:consistency_analysis}

To evaluate the consistency of a 3DSG structure graphically, geometrically and semantically, we collected four datasets over the same geographical area.
The location was a mostly empty RV Park which was in part of the Marina Part 1 location.
We purposefully varied our robot trajectories, start, and end locations for each dataset to analyze Terra's graph consistency across differing views and locations.
Each dataset is referred to as RV Park V1, RV Park V2, RV Park V3, RV Park V4 (or V1, V2, V3, V4 respectively). 

\begin{figure*}[t]
\centering

\begin{tabular}{ccc}
\includegraphics[width=0.32\textwidth]{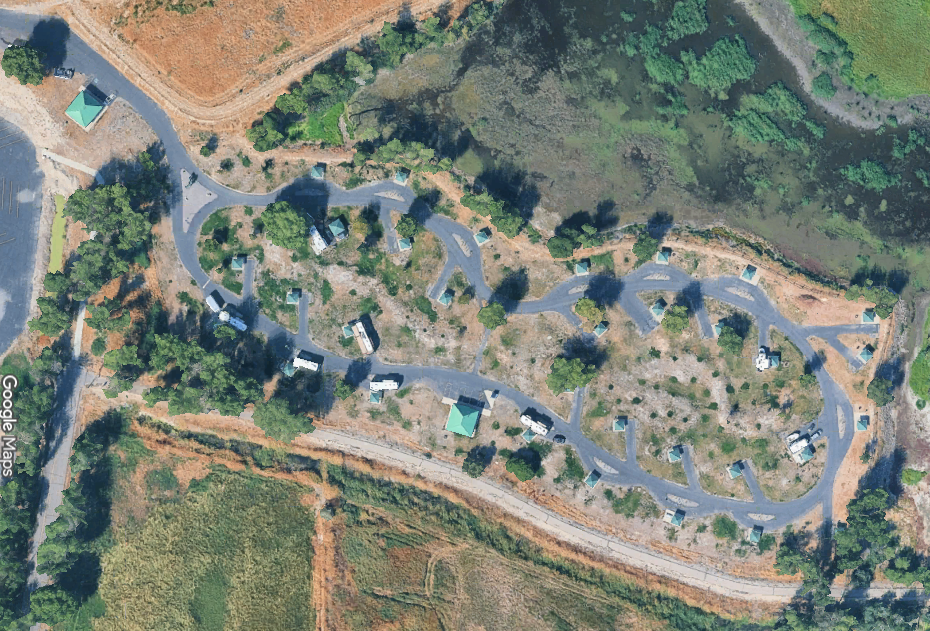} &
\includegraphics[width=0.32\textwidth]{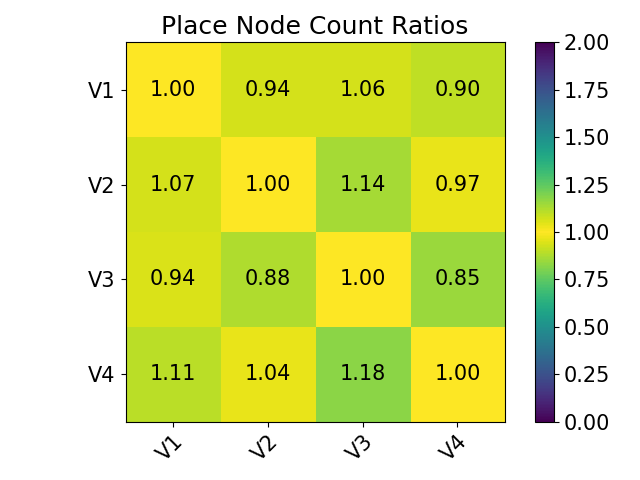} &
\includegraphics[width=0.32\textwidth]{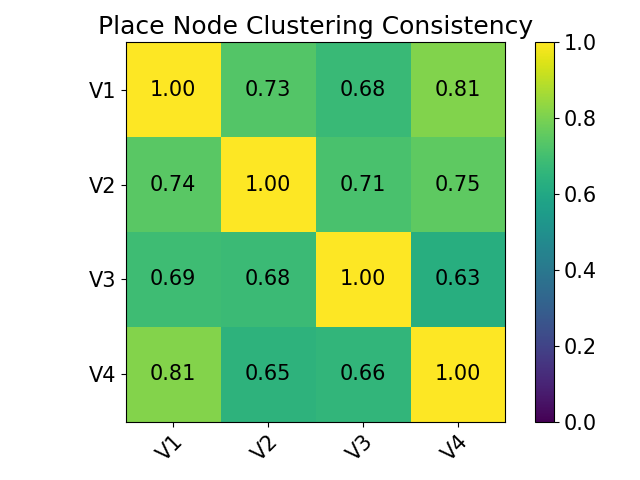} \\

{\fontsize{8}{9.5}\selectfont\rmfamily(a) BEV Satelite View} &%
{\fontsize{8}{9.5}\selectfont\rmfamily(b) NCR} &%
{\fontsize{8}{9.5}\selectfont\rmfamily(c) NCC} \\ %

\includegraphics[width=0.32\textwidth]{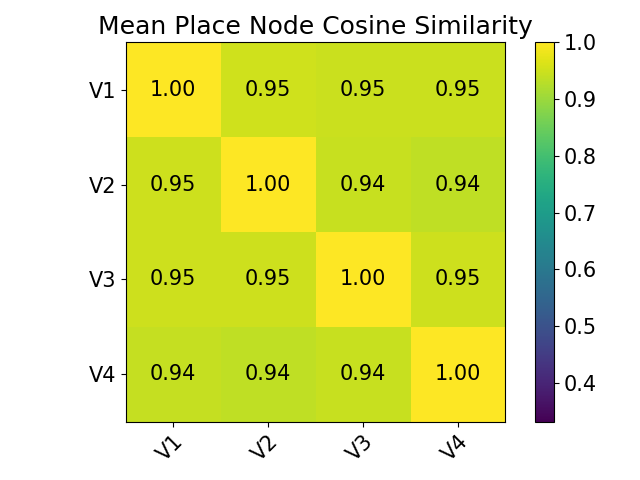} &
\includegraphics[width=0.32\textwidth]{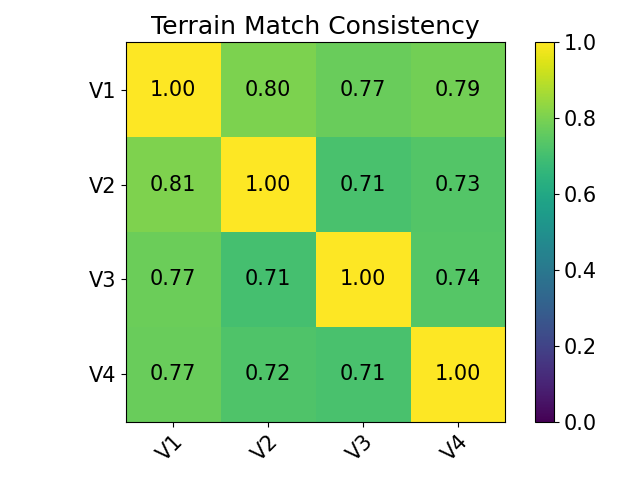} &
\includegraphics[width=0.32\textwidth]{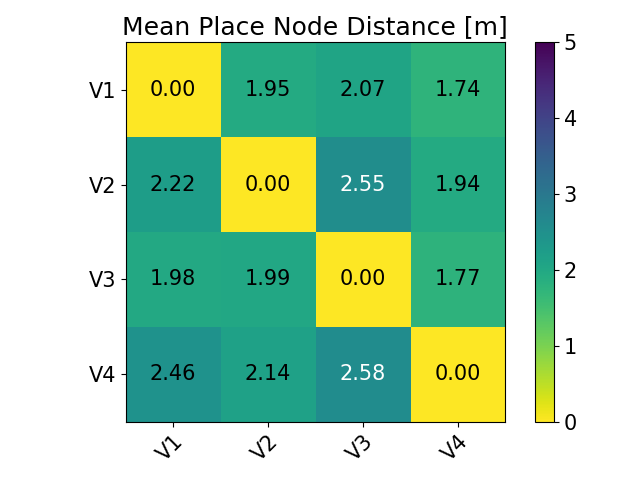} \\

{\fontsize{8}{9.5}\selectfont\rmfamily(d) NCS} & %
{\fontsize{8}{9.5}\selectfont\rmfamily(e) TMC} & %
{\fontsize{8}{9.5}\selectfont\rmfamily(f) Dist} \\ %

\end{tabular} \\

\caption{
Consistency results across RV Park datasets. 
(a) Shows a birds-eye-view (BEV) satelite image of the area mapped.
(b) Evaluates graph structure consistency with the Node Count Ratio (NCR).
(c) Evaluates graph structure consistency with the Node Clustering Consistency (NCC).
(d) Evaluates the semantic consistency with the Node Cosine Similarities (NCS).
(e) Evaluates the semantic consistency with the Terrain Matching Consistency (TMC).
(f) Evaluates the geometric consistency with the Mean distance between nodes (Dist).
All metrics are calculated from the row graph to the column graph.
The results indicate the Terra 3DSG structure is reliably consistent across graph structure, semantics, and geometry.
}
\label{fig:terra_consistency_results}
\end{figure*}

\subsubsection{Experimental Setup}
The built metric maps from LIO-SAM and the place node layer of the generated Terra 3DSGs are shown in Figure~\ref{fig:terra_consistency_maps}.
Prior to consistency evaluations, we compute the transformation to place each dataset into the frame of RV Park V1.
This transformation was extracted using a coarse-to-fine point cloud registration technique built on both iterative closest point (ICP~\cite{beslICP}) and general iterative closest point (GICP~\cite{Segal2009GeneralizedICP}) implementations in the Open3D library \cite{Zhou2018Open3DAM}.
We then transform all place and region node positions, using these transformations, into the RV Park V1 point cloud frame.
For all consistency evaluations, we used agglomerative hierarchical regions. 

\subsubsection{Notation} 
We introduce the following notation to define our consistency metrics.
Let $G_i = (V_i, E_i)$ denote Terra 3DSG $i$, where $V_i$ and $E_i$ are the sets of all nodes and edges respectively in graph $i$.
Let $P_i \subset V_i$ be the set of place nodes (level $=1$),
and $R_i \subset V_i$ the set of region nodes (levels $>1$).
The set of semantic embeddings for the place-node set $P_i$ is represented as $S_i$ where an element of $S_i$ is defined as $s_i \in S_i$.
The terrain ID for the place-node set $P_i$ is represented as $T_i$ where an element of $T_i$ is defined as $t_i \in T_i$.
Unless otherwise specified, distances between place nodes are based on the euclidean distance between the 2D real-world position of the place node. 
The map $\phi_{AB}$ maps a place node $p_A$ in $G_A$ to its nearest-neighbor place node in $G_B$ as follows: 
\begin{equation}
    \phi_{AB} : P_A \to P_B,
    \quad
    \phi_{AB}(p_A) =
    \argmin_{p_B \in P_B}
    \lVert p_A - p_B \rVert.
\end{equation}
The corresponding place association set is
\begin{equation}
    P_{AB} =
    \{ (p_A, \phi_{AB}(p_A)) \mid p_A \in P_A \}.
\end{equation}
We define $S_{AB}$ as the corresponding pairs of semantic embeddings for each place-node pair in $P_{AB}$. 
We further define $T_{AB}$ as a binary indicator vector where each element, $t_{AB} \in T_{AB}$, is either a $1$ if the terrain classification is matching between place-node pair $p_{AB}$ or 0 otherwise, where $p_{AB} \in P_{AB}$ (i.e., $\sum_{T_{AB}} t_{AB} \leq |P_{AB}|$).

Let $R_A^{l}$ and $R_B^{l}$ denote the sets of region nodes
at hierarchical level $l$ in graphs $G_A$ and $G_B$ respectively.
For a region node $r_{A}^{l} \in R_A^{l}$, its set of descendant place nodes is defined as
\begin{equation}
    P_{A}^{l}(r^l_A) =
    \{ p_A \in P_A \mid p_A \text{ is a descendant of } r_{A}^{l} \}.
\end{equation}
$P_{B}^{l}(r^l_B) \subset P_B$ is similarly defined for $r_{B}^{l} \in R_B^{l}$.
We define the nearest-neighbor region mapping at level $l$ as
\begin{equation}
    \psi_{AB}^{l} : R_A^{l} \to R_B^{l},
    \quad
    \psi_{AB}^{l}(r_{A}^{l}) =
    \argmin_{r_{B}^{l} \in R_B^{l}}
    \lVert r_{A}^{l} - r_{B}^{l} \rVert.
\end{equation}
The region association set at level $l$ is
\begin{equation}
    R_{AB}^{l} =
    \{ (r_{A}^{l}, \psi_{AB}^{l}(r_{A}^{l}))
    \mid
    r_{A}^{l} \in R_A^{l} \}.
\end{equation}
For each associated region pair $(r_{A}^{l}, \psi_{AB}^{l}(r_{A}^{l}))$,
we define the clustered place-node association set as
\begin{equation}
\begin{aligned}
    P_{AB}^{l}(r_{A}^{l}) =
    & \{ (p_A, \phi_{AB}(p_A)) \mid p_A \in P_{A}^{l}(r^l_A); \\
    & \phi_{AB}(p_A) \in P_{B}^{l}(\psi_{AB}^l (r^l_A))
    \}.
\end{aligned}
\end{equation}
This set represents the place-node association pairs for place-node children of the region node $r_{A}^{l}$ whose associated nearest neighbor place nodes in $G_B$ are also children of $\psi_{AB}^l (r^l_A) \in R^l_B$. We will use this notation in defining our metrics in the next section.

\subsubsection{Metrics} 
We define five consistency metrics that focus on evaluating the consistency of the graph structure, semantic embeddings, and geometric place-node locations.

To evaluate graph structure consistency, we evaluate place-node count ratios and place-node clusterings.
The place-node count ratio (NCR) metric from $G_A$ to $G_B$ is defined as
\begin{equation}
    NCR_{AB} = \frac{|P_A|}{|P_B|}.
\end{equation} 
NCR computes the ratio of the number of place nodes in $G_A$ compared to $G_B$ to evaluate the consistency in the quantity of place nodes a 3DSG generates.

The metric for evaluating place-node clustering consistency (NCC) from $G_A$ to $G_B$ is defined as
\begin{equation}
    NCC_{AB} = \frac{  
    \sum_{l=1}^L \sum^{R^l_A}_{r^l_A} \left| P_{AB}^l (r^l_A) \right| 
    }{
    \sum_{l=1}^L \sum^{R^l_A}_{r^l_A} \left| P_A^l (r^l_A) \right|
    }.
\end{equation}
NCC measures the ratio of how many children place nodes of a region in $G_A$ are associated to the children place nodes of the corresponding region in $G_B$. 
In other words, a NCC of $1$ means the children place nodes of all regions in $G_A$ perfectly match the same children place nodes of all corresponding regions in $G_B$ demonstrating a consistent hierarchical clustering structure. 

Semantic embedding consistency metrics are the mean place-node cosine similarities (NCS) and terrain matching consistency (TMC). 
Let NCS from $G_A$ to $G_B$ be defined as
\begin{equation}
    NCS_{AB} = \frac{ \sum_{S_{AB}} cos\_sim (s_A, s_B)}{|P_{AB}|},
\end{equation}
where $s_B$ is the semantic embedding of place node $\phi_{AB}(p_A) \in P_B$.
NCS evaluates how consistent the place-node semantic embeddings are in $G_A$ compared to their associated place nodes in $G_B$.
Cosine similarity ranges in values between $-1$ and $1$, meaning an NCS closer to $1$ represents semantically consistent embeddings between associated place nodes.

Let TMC from $G_A$ to $G_B$ be defined as
\begin{equation}
    TMC_{AB} = \frac{\sum_{T_{AB}} t_{AB} }{ |P_{AB}| }.
\end{equation}
TMC is the ratio of matching terrain classified place nodes in $G_A$ and their associated place nodes in $G_B$.
If all terrains match, then $\sum_{T_{AB}} t_{AB} = | P_{AB} |$ resulting in a $TMC_{AB} = 1$.

Lastly, the geometric consistency metric simply computes the distance between each associated place node in $P_{AB}$ (with respect to a global reference frame) and is defined as
\begin{equation}
    Dist_{AB} = \frac{ \sum_{P_{AB}} \lVert p_A - p_B \rVert }{|P_{AB}|},
\end{equation}
where values closer to $0$ represent a more geometrically consistent place-node location between $P_A$ and $P_B$.

These metrics combine to form a thorough summary of the overall consistency of a 3DSG.

\subsubsection{Results} 
The consistency results are shown in Figure~\ref{fig:terra_consistency_results}.
The $NCR$ and $NCS$ results demonstrate robust consistent place node quantities and semantics across all datasets with results close to $1.0$. 
Although cosine similarity ranges from -1 to 1, we scale the colormap from the minimum observed similarity between place node embeddings (approximately 0.3) to the maximum value of 1 to better highlight the high semantic similarity between corresponding place nodes as shown in Figure~\ref{fig:terra_consistency_results}(d).
$TMC$ and $NCC$ demonstrate moderate consistency (i.e. values are above $0.5$) for terrain classification and similar hierarchical clustering structures.
$Dist$ shows that the worst average distance between place nodes for any two datasets is $2.58~m$ demonstrating reliable place node placement which inherits the average $RMSE$ error from imperfect point cloud alignment of about $0.34~m$.

The worst case consistency for terrain ($TMC$), clustering ($NCC$), and distance ($Dist$) is generally between any dataset of RV Park V4 as well as between RV Park V2 and RV Park V3.
These inconsistencies can be largely attributed to the inconsistent grouping of purple nodes (i.e., dirt) at the top of the place node graphs as well as the group of blue classified nodes (i.e., leaves) at the bottom left of V4 that are not classified as that terrain type in any other dataset.
There is also inconsistent terrain classifications between classes ``dirt" [purple] and ``grass" [brown].
This is likely due to the time of year, fall and winter, that these datasets were collected in where much of the grass is dead with patches of dirt around it or a shade of yellow or tan similar to that of dirt.
Much of these weaker consistencies can be improved upon with a more robust terrain classification model. 
Overall we see that the Terra 3DSG is robustly consistent in a natural and complex outdoor dataset.

\subsection{GENERAL TAKEAWAYS}
\label{subsec:full_3dsg_takeaways}

Our results demonstrate that outdoor 3DSGs can maintain a sufficiently small memory footprint for deployment in large-scale field robotics applications.
Additionally, through the introduction of novel 3DSG consistency metrics, we show that outdoor 3DSGs can preserve geometric, semantic, and structural consistency across repeated traversals of the same environment.
These properties highlight the potential of 3DSGs for applications such as long-term mapping, robot relocalization, and multi-session outdoor autonomy.

We further observe that the majority of the memory usage in Terra, and likely other 3DSG methods, arises from the underlying MS-Map representation, while the place-node and region-node graph structures require comparatively little memory and exhibit strong consistency across sessions.
This suggests an interesting direction for future work: lightweight 3DSG representations built primarily from place and region node hierarchies.
Such representations could enable compact, consistent map models suitable for downstream tasks including collaborative multi-robot mapping, localization, and long-term environment understanding.

\section{LESSONS LEARNED AND FUTURE DIRECTIONS}
\label{sec:lessons_learned}

This field report provides several broader insights into the challenges and opportunities of deploying 3DSGs in complex outdoor environments.
While the previous sections analyzed individual components of outdoor 3DSGs in detail, we summarize here the primary lessons learned across all experiments and datasets.

\subsection{OUTDOOR VLM EMBEDDINGS CONTAIN OUTLIERS AND MULTIPLE SEMANTIC MODES}

A major observation throughout this work is that outdoor VLM embeddings associated with a 3D location or object are often neither clean nor unimodal.
Semantic embeddings associated with a single geometric map point frequently contain outliers as well as multiple distinct semantic modes caused by viewpoint variation, segmentation inconsistencies, occlusions, lighting changes, and partial observations.
These factors make multimodal embedding distributions an important consideration for outdoor 3DSGs that associate image-derived VLM embeddings with persistent 3D locations, particularly when SAM-like class-agnostic segmentation produces masks with varying semantic content across frames.

Naively averaging embeddings into a single representation may discard important semantic information and reduce downstream reasoning performance.
Instead, future work should investigate methods for robustly identifying, filtering, and leveraging multiple semantic modes within outdoor semantic maps.

\subsection{OUTDOOR 3DSG NAVIGATION REQUIRES TRAVERSABILITY-AWARE GRAPHS}

Our experiments demonstrate that outdoor 3DSGs can successfully support navigation to semantically queried objects.
However, navigation performance depends heavily on the structure of the underlying place-node graph.

While existing graph construction approaches provide safe navigational structures, they do not inherently encode traversability constraints or path efficiency.
As a result, planned paths may be inefficient or may fail in challenging outdoor settings.
These results suggest that future outdoor 3DSGs should more tightly integrate traversability reasoning, terrain understanding, and navigation-aware graph generation directly into the place-node layer.

\subsection{OUTDOOR REGION UNDERSTANDING IS HIGHLY AMBIGUOUS}

Region-level reasoning remains one of the most challenging aspects of outdoor 3DSGs.
Natural outdoor environments often contain regions with ambiguous boundaries, overlapping semantic descriptions, and homogeneous appearances.
This ambiguity makes both region clustering and region querying difficult and highly sensitive to prompt wording and semantic representation choices.

We additionally observe evidence that averaging embeddings, as done in Terra, to summarize region semantics may suppress important semantic information needed to distinguish regions effectively.
Future work should therefore investigate richer methods for representing and summarizing region semantics that preserve broader contextual and structural information.

\subsection{HIERARCHICAL 3DSGS ENABLE COMPACT LARGE-SCALE MAPPING}

Despite the challenges identified throughout this study, Terra demonstrated strong scalability and memory efficiency in large-scale environments.
Most memory usage originates from dense low-level semantic map representations (i.e., MS-Map), while the higher-level place-node and region-node graph layers remain comparatively lightweight. 
Thus, the memory use of the underlying low-level semantic map representation has the largest impact on memory efficiency for any outdoor 3DSG method.

This suggests an interesting future direction in which outdoor 3DSGs are represented purely through compact place-node and region-node hierarchies.
Such lightweight graph-centric representations could enable highly compressed maps suitable for long-term autonomy, collaborative multi-robot mapping, and large-scale deployment.

\subsection{OUTDOOR 3DSGS MAINTAIN STRONG MULTI-SESSION CONSISTENCY}

Across repeated traversals of the same environment, the Terra 3DSGs evaluated in this work exhibit relatively strong geometric, semantic, and structural consistency demonstrating the potential for consistent mapping.
Although some inconsistencies arise from terrain ambiguity, segmentation variability, and environmental appearance changes, the overall graph structures remain stable across sessions.

These results demonstrate the potential of outdoor 3DSGs for long-term mapping, relocalization, and persistent environment understanding.
The extent to which these results generalize to other 3DSG methods should be evaluated on a method-specific basis.
The metrics presented in this paper may serve as a framework for these evaluations.

\section{CONCLUSION}
\label{sec:conclusion}

This field report investigated the behavior and utility of outdoor 3D scene graphs (3DSGs) across five diverse outdoor robotic datasets using the Terra 3DSG as a case study. 
Through analyses spanning VLM point embeddings, place-node navigation, region-level reasoning, memory efficiency, and multi-session consistency, we identified several key challenges and opportunities that arise when deploying 3DSGs in complex outdoor environments.

Our experiments revealed that outdoor semantic embeddings exhibit outlier behavior, with outlier ratios above $0.1$ for around $30\%$ of points, and multiple distinct semantic modes potentially caused by viewpoint variation, segmentation inconsistencies, occlusions, and environmental ambiguity. 
We additionally demonstrated that outdoor Terra 3DSGs can support contextual object navigation with success rates around $70\%$ and path efficiency around $66\%$, though more robust navigation requires traversability-aware and path efficient graph structures. 
Region-level reasoning remains  particularly challenging due to the ambiguous and subjective nature of outdoor regions shown by a low average F1 score around $0.359$.
Additionally, simple embedding averaging for region semantics often fails to obtain higher-level semantics necessary for reliable region understanding.

Despite these challenges, our results show that outdoor 3DSGs can maintain compact memory footprints (less than $600$MB for multi-kilometer trajectories for Terra 3DSGs) and relatively consistent geometric, semantic, and structural representations across repeated traversals of the same environment. 
These findings highlight the potential of outdoor 3DSGs for long-term autonomy, relocalization, and large-scale, multi-session mapping as well as the potential for cooperative multi-agent mapping and navigation. 
Overall, this work demonstrates the potential of outdoor 3DSGs as semantic mapping frameworks for field robotics while highlighting several challenges that must be addressed for robust deployment, including (1) handling multiple semantic modes, (2) incorporating traversability directly into graph structures, and (3) providing richer high-level outdoor scene understanding.

\section*{APPENDIX}
\label{sec:appendix}

The parameters used to construct all Terra 3DSGs are explicitly listed and described in Table~\ref{tab:terra_params}.

\begin{table*}[t]
\centering
\caption{Terra System Parameters}
\setlength{\tabcolsep}{3pt}
\begin{tabular}{p{120pt}p{200pt}p{50pt}p{65pt}}
\hline
\textbf{Parameter} & \textbf{Description} & \textbf{Symbol} & \textbf{Value} \\
\hline

\hline
\multicolumn{4}{c}{\textbf{Metric Map}} \\ 
\hline
Surface Voxel Size & Surface voxel resolution for LIO-SAM & $v_s$ & $0.4$ m \\
Corner Voxel Size & Corner voxel resolution for LIO-SAM & $v_c$ & $0.2$ m \\

\hline
\multicolumn{4}{c}{\textbf{Phase 1: Metric-Semantic Map}} \\ 
\hline
Time Alignment Threshold & Max LiDAR-image timestamp difference & $\tau_{align}$ & $0.05$ s \\
Image Resolution & Input RGB resolution & $H \times W$ & $480 \times 768$ \\
Terrain Confidence Threshold & YOLO terrain confidence cutoff & $\tau_{terrain}$ & $0.05$ \\
Embedding Match Threshold & Cosine similarity for embedding match & $\tau_{match}$ & $0.90$ \\
DBSCAN Radius & Neighborhood radius for clustering & $\epsilon_{ms}$ & $2.0$ m \\
DBSCAN Min Samples & Minimum points per cluster & $p_{min,ms}$ & $5$ \\
Camera-Point Distance & Max frame–point association distance & $d_{cp}$ & $20.0$ m \\ 

\hline
\multicolumn{4}{c}{\textbf{Phase 2: Task-Agnostic 3DSG}} \\ 
\hline
Terrain Similarity Threshold & Cosine similarity for terrain assignment & $\tau_{terrain,sim}$ & $0.90$ \\
GVD Resolution & Grid cell size & $c_{res}$ & $0.4$ m \\
Floodfill Iterations & Max GVD floodfill iterations & $\tau_{ff}$ & $10$ \\
Max Deviation & Split deviation threshold & $\tau_{dev}$ & $2.0$ m \\
Max Node Distance & Split distance threshold & $\tau_{dist}$ & $5.0$ m \\
LiDAR Height Offset & Ground height from LiDAR frame & $z_L$ & $-2.5$ m \\
Semantic Edge Weight & Weight for semantic similarity edges & $\gamma_{edge}$ & $100.0$ \\

\hline
\multicolumn{4}{c}{\textbf{Phase 3: Task-Driven 3DSG Querying}} \\ 
\hline
DBSCAN Radius & Query clustering radius & $\epsilon_{q}$ & $1.0$ m \\
DBSCAN Min Samples & Min points for query clusters & $p_{min,q}$ & $4$ \\
Search Radius & Neighborhood of semantic points prior to clustering & $r$ & $2.5$ m \\
Similarity Threshold & Task relevance similarity cutoff & $\alpha$ & $0.26$ \\
Agglomerative Distances & Distances for agglomerative region splitting & $d_{agg}$ & $\begin{bmatrix} 50, 100, 200, 400 \end{bmatrix}$ \\
Min Region Area & Minimum region area threshold for spectral clustering & $\tau_{spec,area}$ & $10{,}000$ m$^2$ \\
Max Semantic Difference & Max similarity difference for splitting in spectral clustering & $\tau_{spec,diff}$ & $0.45$ \\

\hline
\multicolumn{4}{p{435pt}}{Terra System Parameters.}
\end{tabular}
\label{tab:terra_params}
\end{table*}

We also provide a brief ablation to justify the choice of $\alpha=0.26$ used throughout this article's experiments.
We perform object detection experiments on the datasets used in this article, following an evaluation procedure similar to~\cite{samuelsonTerra2025}, except that we use only the MS-Max method and compute F1-scores. 
We first perform a coarse sweep of $\alpha$ values from $0.20$--$0.33$ on the River Park dataset, chosen for its richness and diversity in object classes, and identify $0.25$--$0.27$ as the range yielding the highest F1-scores.
We then perform a refined sweep across all datasets using $\alpha \in[0.25,0.26,0.27]$ to select the value that provides the most consistent performance across datasets.
The resulting F1-scores are reported in Table~\ref{tab:alpha_ablation}.

\begin{table}[t]
\centering
\caption{$\alpha$ Ablation Study}
\begin{tabular}{cccc}
\hline
    Dataset & $\alpha=0.25$ & $\alpha=0.26$ & $\alpha=0.27$ \\
    \hline
    River Park & 0.308 & \textbf{0.478} & 0.333 \\ 
    Nunns Park & 0.195 & 0.420 & \textbf{0.499} \\
    Marina P1 & 0.162 & 0.188 & \textbf{0.213} \\
    Marina P2 & 0.260 & \textbf{0.368} & 0.197 \\
    Rock Canyon & \textbf{0.143} & \textbf{0.143} & 0.000 \\
    \hline
    Average F1-Score & 0.214 & \textbf{0.319} & 0.248 \\
\hline
\multicolumn{4}{p{220pt}}{F1-score sensitivity to the task-relevance threshold $\alpha$ using the MS-Max method for object detection.
The refined sweep shows that $\alpha=0.26$ achieves the highest average F1-score across the five datasets.}
\end{tabular}
\label{tab:alpha_ablation}
\end{table}

The object prompts and context for each dataset as used in \ref{sec:place_node_nav_analysis}.\ref{subsec:place_node_nav_exp}
are listed in Table~\ref{tab:object_prompts}.

\begin{table}[t]
\centering
\caption{Object Prompt Descriptions}
\begin{tabular}{p{50pt}p{175pt}}
\hline
Object & Context \\
\hline
\multicolumn{2}{l}{\textbf{River Park}} \\ 
\hline
``bridge" & ``large brown over the river" \\ 
``bridge" & ``small white in the woods" \\
``playground" & ``in sand" \\
``garbage" & ``in the parking lot" \\
``garbage" & ``by the red building" \\

\hline
\multicolumn{2}{l}{\textbf{Nunns Park}} \\ 
\hline
``underpass" & ``go in" \\
``honey bucket" & ``blue" \\
``trailer" & ``behind the truck" \\
``dumpster" & ``by the brick building" \\
``playground" & ``in the sand" \\

\hline
\multicolumn{2}{l}{\textbf{Marina Part 1}} \\ \hline
``dumpster" & ``near the end of a road by rocks and water" \\
``stop sign" & ``by road" \\ 
``fire hydrant" & ``on grass in RV park" \\
``dumpster" & ``in white fence by tree" \\

\hline
\multicolumn{2}{l}{\textbf{Marina Part 2}} \\ \hline
``building" & ``brick with a green roof" \\
``dumpsters" & ``by building" \\
``buoy" & ``white round on road that says SLOW NO WAKE" \\

\hline
\multicolumn{2}{p{235pt}}{Object and context prompts used for 3DSG navigation for each dataset.}
\end{tabular}
\label{tab:object_prompts}
\end{table}

A brief ablation on the selection of object $w_o$, and context, $w_c$, weights for the \textit{Object+Context} method, as described in Sections \ref{sec:place_node_nav_analysis}.\ref{subsec:nav_context_object_method} are listed in Table~\ref{tab:context_object_ablation}.

\begin{table}[t]
\centering
\caption{Contextualized Object Querying Ablation Study}
\begin{tabular}{ccc}
\hline
    Object Weight & Context Weight & SS-1 \\
    \hline
    $w_o=0.0$ & $w_c=1.0$ & 0.412 \\ 
    $w_o=0.1$ & $w_c=0.9$ & 0.471 \\ 
    $w_o=0.2$ & $w_c=0.8$ & 0.471 \\ 
    $w_o=0.3$ & $w_c=0.7$ & 0.647 \\ 
    $w_o=0.4$ & $w_c=0.6$ & 0.588 \\ 
    $w_o=0.5$ & $w_c=0.5$ & 0.588 \\ 
    $w_o=0.6$ & $w_c=0.4$ & 0.706 \\ 
    $w_o=0.7$ & $w_c=0.3$ & \textbf{0.765} \\ 
    $w_o=0.8$ & $w_c=0.2$ & \textbf{0.765} \\ 
    $w_o=0.9$ & $w_c=0.1$ & \textbf{0.765} \\
    $w_o=1.0$ & $w_c=0.0$ & 0.706 \\ 
\hline
\multicolumn{3}{p{190pt}}{Ablation study to select the best weight ratio to balance object and context information for object prediction.}
\end{tabular}
\label{tab:context_object_ablation}
\end{table}

The region prompts for each dataset as used in \ref{sec:region_analysis}.\ref{subsec:region_querying} are listed in Table~\ref{tab:region_prompts}.

\begin{table}[t]
\centering
\caption{Region Prompt Descriptions}
\begin{tabular}{p{170pt}p{55pt}}
\hline
Region Prompt & GT Node Count \\
\hline

\textbf{River Park} - 4 regions \\ \hline
``Playground area with slides and swings'' & 30 \\
``Dense trees with picnic tables'' & 214 \\
``Duck Pond'' & 51 \\
``Open field with red pavilions'' & 49 \\

\hline
\textbf{Nunns Park} - 4 regions \\ \hline
``Paved trail with the white posts'' & 83 \\
``Trailhead parking lot with cars and porta potties'' & 69 \\
``Large empty parking lot with blue dumpsters'' & 108 \\
``Paved campsites'' & 259 \\

\hline
\textbf{Marina Part 1} - 4 regions\\ \hline
``Pavilions and picnic tables with parking stalls'' & 1257 \\
``Long road with water on both sides'' & 470 \\
``Boat ramps and docks with parking lot'' & 179 \\
``Buildings and sidewalk path'' & 266 \\

\hline
\textbf{Marina Part 2} - 1 region \\ \hline
``Long dirt road with water and large rocks'' & 371 \\

\hline
\textbf{Rock Canyon Campground} - 2 regions \\ \hline
``Large open dirt clearing with fire pit'' & 33 \\
``Dead grass and short brush field'' & 113 \\

\hline
\multicolumn{2}{p{235pt}}{Region prompts used and the ground truth number of associated place nodes used for each dataset.}
\end{tabular}
\label{tab:region_prompts}
\end{table}

\section*{ACKNOWLEDGMENT}
ChatGPT-4 and 5 and GitHub Copilot AI models were used as collaborative tools for code development. All generated code was thoroughly edited and tested by the authors and the authors accept full responsibility for the results.

\bibliographystyle{IEEEtran}
\bibliography{references}

\begin{IEEEbiography}[{\includegraphics[width=1in,height=1.25in,clip,keepaspectratio]{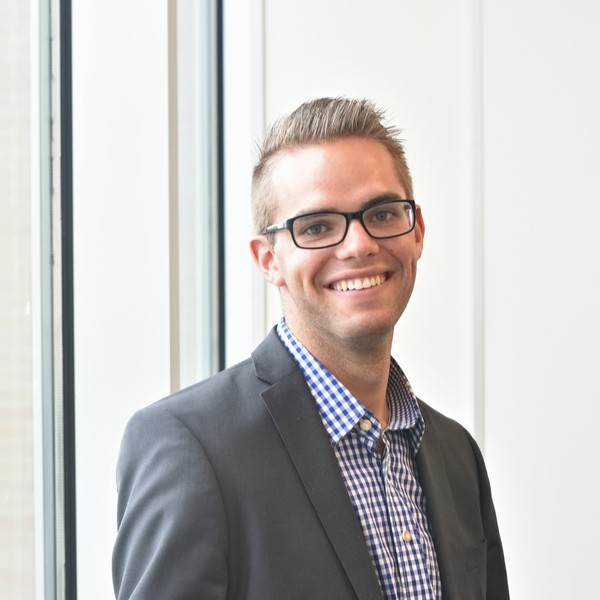}}]
{CHAD R. SAMUELSON}~received the B.Sc. degree in mechanical engineering with a minor in computer science from Brigham Young University (BYU), Provo, Utah, USA, in 2021.
He is currently pursuing a Ph.D. degree in electrical and computer engineering at BYU with the Field Robotic Systems Lab (FROST Lab).

His research interests include robotic perception, semantic mapping, 3D scene graphs, and semantic uncertainty. 
\end{IEEEbiography}

\begin{IEEEbiography}[{\includegraphics[width=1in,height=1.25in,clip,keepaspectratio]{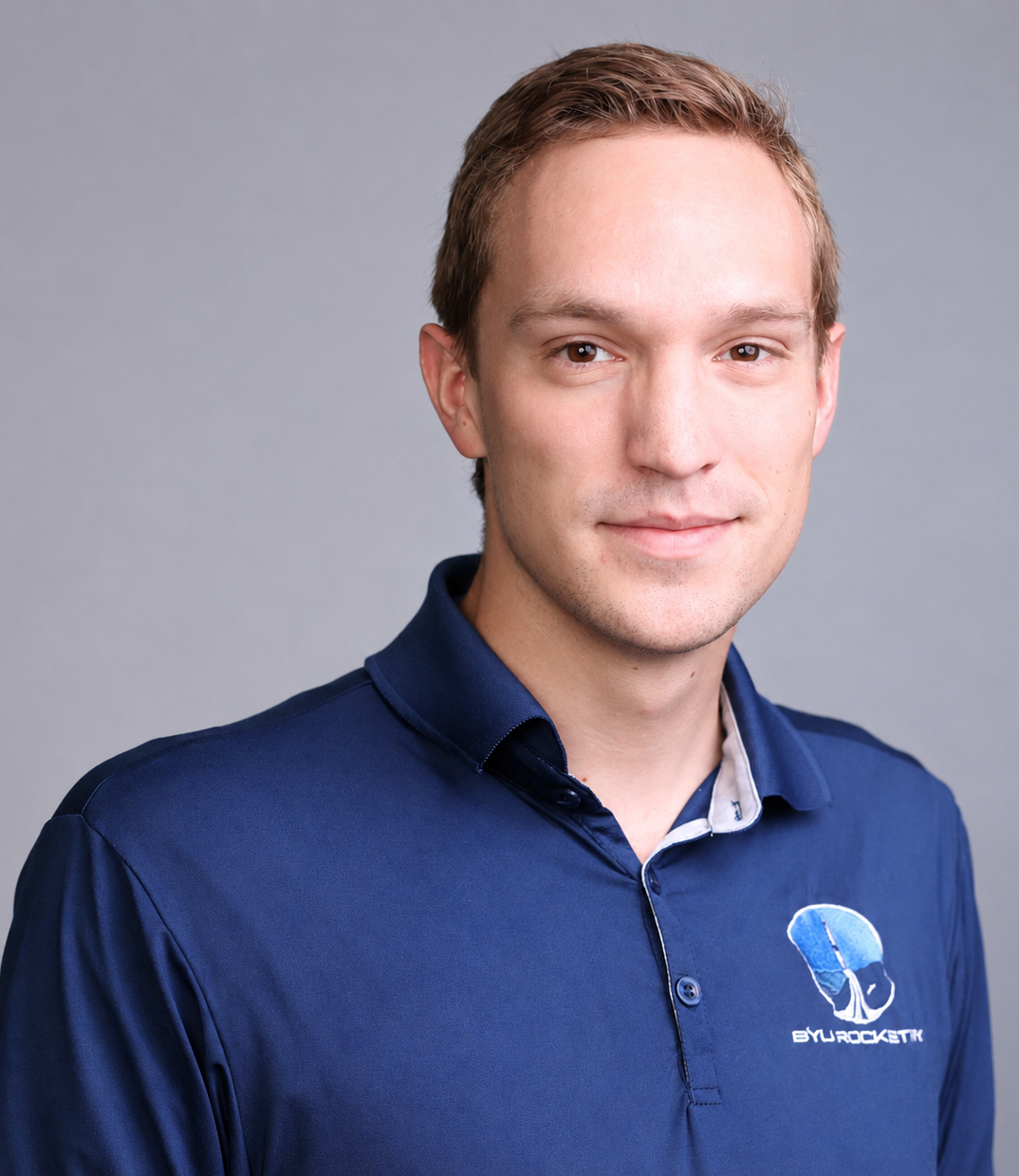}}]
{Gabriel R. Slade}~received the B.Sc. degree in mechanical engineering with a minor in computer science from Brigham Young University (BYU), Provo, Utah, USA, in 2025.
He is currently pursuing a M.S. degree in electrical and computer engineering at BYU with the Field Robotic Systems Lab (FROST Lab).

His research interests include robotic vision, semantic mapping, 3D scene graphs, and marine robotics.
\end{IEEEbiography}

\begin{IEEEbiography}
[{\includegraphics[width=1in,height=1.25in,clip,keepaspectratio]{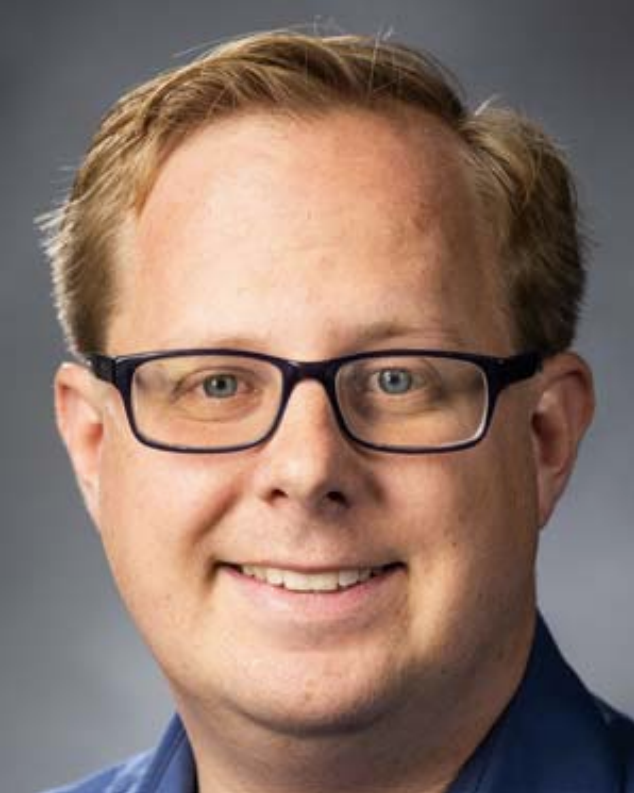}}]
{Joshua G. Mangelson}~(Member, IEEE) received the B.S. degree in electrical engineering from Brigham Young University, Provo, UT, USA, in 2014, and the M.S. and Ph.D. degrees in robotics from the University of Michigan, Ann Arbor, MI, USA, in 2016 and 2019, respectively. Following completion of his doctorate, he was a Postdoctoral Research Fellow with Carnegie Mellon University. He subsequently joined the Department of Electrical and Computer Engineering at Brigham Young University, where he is currently an Associate Professor.

His research interests include robotic perception, mapping, and localization, with an emphasis on marine robotics. He has conducted field trials in locations worldwide, including Hawaii, San Diego, Boston, northern Michigan, and Utah. His work has received multiple best paper and poster awards, and he is a recipient of the Office of Naval Research Young Investigator Award (2024). Dr. Mangelson currently serves as an Associate Editor for \textit{The International Journal of Robotics Research}, \textit{IEEE Robotics and Automation Letters}, and \textit{IEEE Transactions on Field Robotics.}
\end{IEEEbiography}

\vfill\pagebreak

\end{document}